\documentclass[sigconf,screen]{acmart}
\AtBeginDocument{%
  }

\copyrightyear{2025}
\acmYear{2025}
\setcopyright{acmlicensed}\acmConference[KDD '25]{Proceedings of the 31st ACM SIGKDD Conference on Knowledge Discovery and Data Mining V.1}{August 3--7, 2025}{Toronto, ON, Canada}
\acmBooktitle{Proceedings of the 31st ACM SIGKDD Conference on Knowledge Discovery and Data Mining V.1 (KDD '25), August 3--7, 2025, Toronto, ON, Canada}
\acmDOI{10.1145/3690624.3709237}
\acmISBN{979-8-4007-1245-6/25/08}

\usepackage{microtype}
\usepackage{graphicx}
\usepackage{subfigure}
\usepackage{balance}
\usepackage{booktabs} % for professional tables
\usepackage{multicol,multirow}
\usepackage{bbding}
\usepackage{amsmath}
\usepackage{mathtools}
\usepackage{amsthm}
\usepackage{tcolorbox}
\definecolor{dkblue}{rgb}{0,0.08,0.45}
\newcommand{\red}[1]{\textcolor{red}{#1}}

\definecolor{tickgreen}{rgb}{0,0.6,0}
\newcommand{\tick}{\textcolor{tickgreen}{\CheckmarkBold}}
\newcommand{\fail}{\textcolor{red}{\XSolidBrush}}
\usepackage{wrapfig}

\theoremstyle{plain}
\newtheorem{theorem}{Theorem}[section]
\newtheorem{proposition}[theorem]{Proposition}

\theoremstyle{definition}

\newtheorem{assumption}[theorem]{Condition}

\theoremstyle{remark}

\DeclareMathOperator*{\argmin}{arg\,min}
\DeclareMathOperator*{\argmax}{arg\,max}
\DeclareMathOperator*{\bias}{\mathrm{Bias}}
\DeclareMathOperator*{\var}{\mathrm{Var}}

\DeclareMathOperator*{\mse}{\mathrm{MSE}}

\DeclareMathOperator*{\pair}{\text{pair}}

\DeclareMathOperator*{\onehottimefeature}{\mathrm{one\_hot_{\phi(t)}}}
\DeclareMathOperator*{\onehottimefeaturecontext}{\mathrm{one\_hot_{\phi_x(t)}}}
\DeclareMathOperator*{\onehottimefeaturetranspose}{\mathrm{one\_hot_{\phi(t)}^\top}}
\DeclareMathOperator*{\onehottimefeaturefiner}{\mathrm{one\_hot_{\phi_f(t)}}}
\DeclareMathOperator*{\onehottimefeaturephi}{\mathrm{one\_hot_{\phi_p}}}
\DeclareMathOperator*{\onehottimefeaturefinertranspose}{\mathrm{one\_hot_{\phi_f(t)}^\top}}
\DeclareMathOperator*{\onehotaction}{\mathrm{one\_hot_{a}}}
\DeclareMathOperator*{\onehotactiontimefeature}{\mathrm{one\_hot_{\phi(t), a}}}
\DeclareMathOperator*{\onehotactiontimefeaturefiner}{\mathrm{one\_hot_{\phi_f(t), a}}}
\DeclareMathOperator*{\oldesttime}{\mathrm{t\_oldest}}
\DeclareMathOperator*{\futuretime}{\mathrm{t\_future}}
\newcommand{\meanN}{\frac{1}{n} \sum_{i=1}^n}

\newcommand{\sumA}{\sum_{a \in \mathcal{A}}}

\newcommand{\mE}{\mathbb{E}}
\newcommand{\mV}{\mathbb{V}}

\newcommand{\mR}{\mathbb{R}}

\newcommand{\ind}{\mathbb{I}}
\newcommand{\calD}{\mathcal{D}}
\newcommand{\calX}{\mathcal{X}}
\newcommand{\calA}{\mathcal{A}}

\newcommand{\calC}{\mathcal{C}}
\newcommand{\calN}{\mathcal{N}}

\newcommand{\trueV}{V(\pi_e)}
\newcommand{\ips}{\hat{V}_{\mathrm{IPS}} (\pi_e; \calD)}
\newcommand{\dr}{\hat{V}_{\mathrm{DR}} (\pi_e; \calD, \hat{q})}

\newcommand{\drtarget}{\hat{V}_{t'}^{\mathrm{DR}} (\pi_e; \calD, \hat{q})}

\newcommand{\opfv}{\hat{V}_{t'}^{\mathrm{OPFV}} (\pi_e; \calD)}
\newcommand{\opfvphi}{\hat{V}_{t'}^{\mathrm{OPFV}} (\pi_e; \calD, \phi)}

\newcommand{\opfvphiinf}{\hat{V}_{t'}^{\mathrm{OPFV}} (\pi_e; \calD, \phi_{\infty})}

\newcommand{\opfvgrad}{ \nabla_{\zeta} \hat{V}_{t'}^{\mathrm{OPFV}} (\pi_{\zeta}; \calD)}

\begin{document}

\title{Off-Policy Evaluation and Learning for the Future\\under Non-Stationarity}

\author{Tatsuhiro Shimizu} 
\affiliation{
\institution{Yale University}
\state{Connecticut}
\country{USA}}
\email{tatsuhiro.shimizu@yale.edu}

\author{Kazuki Kawamura} 
\affiliation{
\institution{Sony Group Corporation}
\state{Tokyo}
\country{Japan}}
\email{Kazuki.Kawamura@sony.com}

\author{Takanori Muroi} 
\affiliation{
\institution{Sony Group Corporation}
\state{Tokyo}
\country{Japan}}
\email{Takanori.Muroi@sony.com}

\author{Yusuke Narita} 
\affiliation{
\institution{Yale University}
\state{Connecticut}
\country{USA}}
\email{yusuke.narita@yale.edu}

\author{Kei Tateno} 
\affiliation{
\institution{Sony Group Corporation}
\state{Tokyo}
\country{Japan}}
\email{Kei.Tateno@sony.com}

\author{Takuma Udagawa} 
\affiliation{
\institution{Sony Group Corporation}
\state{Tokyo}
\country{Japan}}
\email{Takuma.Udagawa@sony.com}

\author{Yuta Saito}
\affiliation{
\institution{Cornell University}
\state{New York}
\country{USA}}
\email{ys552@cornell.edu}

\renewcommand{\shortauthors}{Tatsuhiro Shimizu, et al.}

\begin{abstract}
    We study the novel problem of future off-policy evaluation (F-OPE) and learning (F-OPL) for estimating and optimizing the future value of policies in non-stationary environments, where distributions vary over time. In e-commerce recommendations, for instance, our goal is often to estimate and optimize the policy value for the upcoming month using data collected by an old policy in the previous month. A critical challenge is that data related to the future environment is not observed in the historical data. Existing methods assume stationarity or depend on restrictive reward-modeling assumptions, leading to significant bias. To address these limitations, we propose a novel estimator named \textit{\textbf{O}ff-\textbf{P}olicy Estimator for the \textbf{F}uture \textbf{V}alue (\textbf{\textit{OPFV}})}, designed for accurately estimating policy values at any future time point. The key feature of OPFV is its ability to leverage the useful structure within time-series data. While future data might not be present in the historical log, we can leverage, for example, seasonal, weekly, or holiday effects that are consistent in both the historical and future data. Our estimator is the first to exploit these time-related structures via a new type of importance weighting, enabling effective F-OPE. Theoretical analysis identifies the conditions under which OPFV becomes low-bias. In addition, we extend our estimator to develop a new policy-gradient method to proactively learn a good future policy using only historical data. Empirical results show that our methods substantially outperform existing methods in estimating and optimizing the future policy value under non-stationarity for various experimental setups.
\end{abstract}

\begin{CCSXML}
<ccs2012>
   <concept>
       <concept_id>10010147.10010257.10010282.10010292</concept_id>
       <concept_desc>Computing methodologies~Learning from implicit feedback</concept_desc>
       <concept_significance>500</concept_significance>
       </concept>
   <concept>
       <concept_id>10010147.10010257.10010282.10010283</concept_id>
       <concept_desc>Computing methodologies~Batch learning</concept_desc>
       <concept_significance>500</concept_significance>
       </concept>
   <concept>
       <concept_id>10010147.10010257.10010258.10010259.10003343</concept_id>
       <concept_desc>Computing methodologies~Learning to rank</concept_desc>
       <concept_significance>300</concept_significance>
       </concept>
   <concept>
       <concept_id>10010147.10010257.10010258.10010259.10003268</concept_id>
       <concept_desc>Computing methodologies~Ranking</concept_desc>
       <concept_significance>300</concept_significance>
       </concept>
 </ccs2012>
\end{CCSXML}
\ccsdesc[500]{Computing methodologies~Batch learning}

\keywords{Off-Policy Evaluation and Learning, Non-stationarity.}

\maketitle

\section{Introduction}
Various decision-making problems, ranging from recommendation systems to precision medicine, increasingly use logged interaction data to implement optimal decisions. Most of these systems operate under a \textit{contextual bandit protocol}, where they observe the context, take actions based on a bandit policy, and observe the resulting rewards. In such an automated decision-making problem, a crucial counterfactual question is how effective a new policy would have been compared to the one actually implemented in the past. This problem is known as \textit{Off-Policy Evaluation (OPE)}~\citep{dudik2011doubly,saito2021counterfactual,kiyohara2024towards}. Answering this question using historical logged data is important, as conducting an online experiment may entail ethical issues and significant expense~\citep{agarwal2020reinforcement,saito2021counterfactual}.

While there has been considerable progress in OPE within stationary environments, where the distributions and the value of policies remain constant over time, most practical applications of OPE are concerned with the future effectiveness of new policies under varying distributions~\citep{chandak2020optimizing,jagerman2019people,thomas2017predictive}. For instance, when developing a new recommendation policy, we are interested in its expected reward during the upcoming week or month. However, the historical data available for this estimation was collected by an older (logging) policy in previous weeks or months. A challenge here is the absence of observed data from future distributions. Most existing methods, such as Inverse Propensity Scoring (IPS)~\citep{precup2000eligibility} and Doubly Robust (DR)~\citep{dudik2011doubly}, estimate the value of a new policy based on past data (when the logs were collected) and naively apply their estimates to predict the future policy value, assuming a stationary distribution. This assumption of stationarity, however, is often unrealistic and violated in practical situations~\citep{chandak2020optimizing,jagerman2019people,liu2023asymptotically}. For instance, in content recommendations, conversion rates (rewards) tend to vary over time. A typical example can be observed in music streaming services, where users' preferences for songs or podcasts change between weekdays and weekends, as well as across different seasons. This variability renders the standard formulation ineffective~\citep{chandak2020optimizing}. Some previous methods have attempted to address this limitation of the typical formulation. In particular, Chandak et al.~\citep{chandak2020optimizing} tackle abrupt non-stationarity, where expected rewards can incur non-smooth jumps over time. Under such a formulation, they propose a method named Prognosticator to address both OPE and OPL in terms of future policy performance. Prognosticator first estimates the value of a new policy at each past time step using typical OPE estimators, and then predicts the future value by learning its time-series trend via a parametric regression. While this approach often outperforms typical OPE, it still introduces significant bias in its regression stage due to the difficulty of learning a regressor that generalizes well to the future using only past data.

To overcome the limitations of previous methods, we first formally formulate the \textit{Future OPE (F-OPE)} problem, which aims to estimate the future value of new policies using only historical data provided by a logging policy in the past, under non-stationary (context and reward) distributions. In particular, we formulate F-OPE, considering the timestamp as a new random variable. This approach allows us to address both smooth and abrupt non-stationarity, previously studied somewhat independently, under a unified framework. Based on this new formulation, we introduce a new estimator for F-OPE, named \textit{\textbf{O}ff-\textbf{P}olicy Estimator for the \textbf{F}uture \textbf{V}alue (\textbf{\textit{OPFV}})}. A key feature of OPFV is its ability to leverage the useful structure in time-series data such as seasonal, weekly, or holiday effects, that are applicable both in the past and in the future. By doing so, OPFV can more accurately estimate the future value of policies with only historical data. Technically, our estimator applies a new type of importance weighting to unbiasedly estimate the effects characterized by the structure in the time series (which we call \textit{time-series features}). It also deals with the residual effect that cannot be fully explained by time-series features alone, using a regression model to further reduce the bias and variance depending on the model's accuracy. In our theoretical analysis, we identify conditions under which our estimator can provide a low-bias estimation of the future values of new policies. We also demonstrate that the choice of time-series features affects the bias-variance tradeoff of OPFV and introduce a simple data-driven procedure for selecting time-series features to optimize the accuracy of OPFV. Additionally, we propose a novel policy-gradient method for learning good future policies using only historical data. This involves extending the OPFV estimator to estimate policy gradients regarding future policy values. Finally, we conducted experiments with both synthetic and real-world recommendation data under non-stationary conditions, demonstrating that OPFV is more substantially effective in evaluating and learning future policies than existing methods.

\section{Typical Formulation}
We first formulate the problem of OPE within the contextual bandit protocol under the typical assumption of stationarity. Let $x \in \calX$ denote a context vector and $a \in \calA$ denote a (discrete) action, such as content recommendation in a movie streaming service. Let $r \in [0, r_{\mathrm{max}}]$ represent a reward variable, which is independently sampled from an unknown conditional distribution $p(r \,|\, x, a)$. A decision-making policy is modeled as a distribution over the action space, i.e., $\pi: \calX \rightarrow \Delta(\calA)$.

In conventional OPE, we are given logged data $\calD := \{(x_i, a_i, r_i)\}_{i=1}^n$, consisting of $n$ independent draws from the \textit{logging policy} $\pi_0$. OPE aims to estimate the expected reward as the \textit{value} of a given \textit{evaluation policy} $\pi_e$ (, which is different from $\pi_0$):
\begin{align}
    \trueV := \mE_{(x, a, r) \sim p(x) \pi_e(a | x) p(r | x, a)} [r]. \label{eq:policy_value}
\end{align}

The value represents the ground-truth performance of the evaluation policy if deployed in an environment of interest. OPE aims to develop an estimator $\hat{V}(\pi_e;\calD)$ to approximate $\trueV$ based only on logged data $\calD$. A typical choice for $\hat{V}$ is IPS:
\begin{align*}
    \ips := \frac{1}{n} \sum_{i=1}^n \frac{\pi_e(a_i \,|\, x_i)}{\pi_0(a_i \,|\, x_i)} r_i,
\end{align*}
Beyond IPS, significant efforts have been made to enable more accurate OPE~\cite{dudik2011doubly,wang2017optimal,su2020doubly,saito2023off,kiyohara2022doubly,metelli2021subgaussian,kallus2021optimal}.

These conventional estimators evaluate the value of an evaluation policy if it were deployed in the past and apply the estimated value as a prediction of the future policy values. This procedure is valid only in a stationary environment, where distributions (i.e., $p(x), p(r | x, a)$) and policy value ($V(\pi_e)$) remain constant over time. Under such conditions, IPS provides an unbiased estimate of the policy value.
However, in most practical situations where the distributions are non-stationary~\citep{chandak2020optimizing,hong2021non,jagerman2019people}, these conventional methods fail to accommodate varying policy values over different time steps, resulting in substantial bias~\citep{chandak2020optimizing}.

\begin{figure*}[t]
\centering
\vspace{1mm}
\includegraphics[clip, width=13cm]{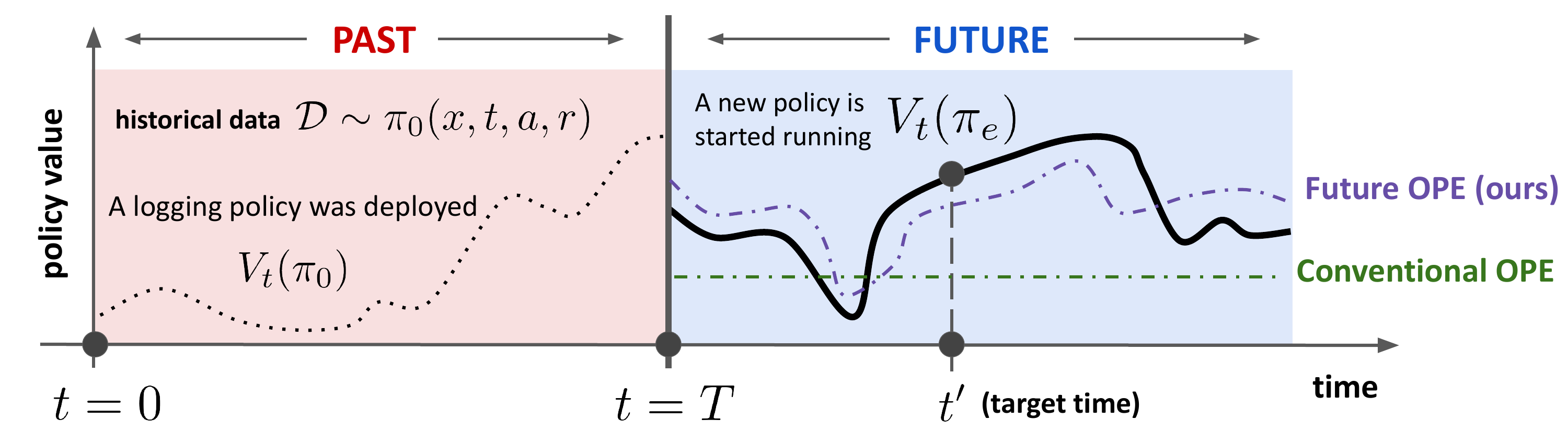}
\vspace{-2mm}
\caption{The key concept of Future OPE (F-OPE) and its comparison with conventional OPE, which operates under the assumption of stationarity. In F-OPE, the policy value $V_t(\pi)$ changes over time due to non-stationary distributions, and our goal is to develop an estimator that adapts to this change in the future ($t > T$) yet using only historical logged data $\calD$ collected by the logging policy $\pi_0$ in the past ($0 \le t \le T$).
} 
\label{fig:f-ope} 
\raggedright
\end{figure*}

\subsection{Existing Methods for OPE in Nonstationary Environments}\label{sec:prognosticator}
Conventional literature on non-stationary OPE differentiates between (1) smooth non-stationarity~\cite{jagerman2019people} and (2) abrupt non-stationarity~\cite{thomas2017predictive,chandak2020optimizing,liu2023asymptotically}. The former refers to situations where the distributions shift gradually over time, whereas the latter involves scenarios characterized by sudden shifts in the distribution. This section focuses on the most relevant work done by~\citet{chandak2020optimizing}, and Appendix~\ref{app:related} discusses other relevant work extensively.

Particularly, in abrupt non-stationary environments, the logged data $\calD$ can be divided into $K$ distinct subsets $\calD = \{\calD_k\}_{k=1}^K$ where the distribution remains constant within each period. Each subset $\calD_k$ is considered generated from a corresponding distribution as
\begin{align*}
    \calD_k
    &:= \{(x_{i, k}, a_{i, k}, r_{i, k})\}_{i=1}^{n_k} \\
    &\sim \prod_{i=1}^{n_k} p_{\red{k}}(x_{i, k}) \pi_0(a_{i, k}\,|\,x_{i, k}) p_{\red{k}}(r_{i, k}\,|\,x_{i, k}, a_{i, k}),
\end{align*}
where $n_k$ represents the number of samples observed in period $k$, and $n = \sum_{k \in [K]} n_k$. For this setting,~\citet{chandak2020optimizing} proposed the Prognosticator algorithm for OPL, which includes an OPE estimator we regard as a baseline method to evaluate the future performance of an evaluation policy. Prognosticator first performs an estimator $\hat{V}_k(\pi_e; \calD)$ such as IPS to estimate the past policy value of $\pi_e$ within each period $k \in [K]$. To estimate the policy value in the future, Prognosticator employs linear regression $\hat{V}_k(\pi_e, \calD) \approx \psi(k)^\top w$, where $\psi(k) \in \mR^d$ is a basis function encoding the period index, $w$ is a regression coefficient, and $\hat{V}_k(\pi_e, \calD)$ is an estimated past policy value. The regression is performed using training data $(\Psi, Y)$ where $\Psi := (\psi(1), \cdots, \psi(K))^\top \in \mR^{K \times d}$ and $Y := (\hat{V}_1(\pi_e; \calD), \cdots, \hat{V}_K(\pi_e; \calD))^\top \in \mR^K$. The future value estimator for period $K+\delta$ ($\delta = 1,2,\ldots$) is given by:
\begin{align*}
    \hat{V}_{K+\delta}(\pi_e; \calD) = \psi(K+\delta)^\top (\Psi^\top \Psi)^{-1} \Psi^\top Y,
\end{align*}
where $w^* := (\Psi^\top \Psi)^{-1} \Psi^\top Y$ is derived from standard ordinary least squares regression. Despite its improved adaptability to non-stationarity compared to conventional OPE, this method faces limitations in both the initial estimation phase and the extrapolation (regression) phase. In the former, Prognosticator experiences significant variance when frequent distribution shifts occur (large $K$) due to the reduced sample size $n_k$ in each subset $\calD_k$. For the extrapolation phase, accurately learning future trends from only the period index $k$ in the past as inputs is extremely challenging with the simple parametric regression, resulting in substantial bias.

\section{Our Formulation and Approach} \label{sec:opfv}
To address the limitations of existing estimators, we first formally formulate the problem of \textit{future OPE}, encompassing both smooth and abrupt changes in distributions. The crux of our new formulation involves considering an unknown probability density $p(t)$ for continuous time $t \in [0, T]$. $[0, T]$ represents the observation window of our historical logged data, where $T$ is the time at which we perform F-OPE of $\pi_e$ (see Figure~\ref{fig:f-ope}). In this new formulation, we observe the context $x$, time $t$, and action $a$ from the joint distribution $\pi(x,t,a) = p(x,t)\pi(a\,|\,x,t)$. Reward $r$ is then sampled from an unknown conditional distribution $p(r\,|\,x, t, a)$. Note that, in our formulation, the reward distribution can be different for every individual $t$, and thus \textbf{we formulate non-stationarity not just seasonality}. Note also that the conventional formulation of OPE is a special case of our F-OPE formulation with $p(x,t)\pi(a|x,t)p(r|x,t,a) = p(x)\pi(a|x)p(r|x,a),\; \forall t \ge 0$.

The goal in our future OPE problem is to estimate the value of evaluation policy $\pi_e$ at a given time point $t'$ in the future (which we call the \textbf{\textit{target time}}), defined as follows.
\begin{align}
    V_{t'}(\pi_e) :=  \mE_{p(x|t')\pi_e(a|x, t')} [q(x, t', a)],  \label{eq:future-policy-value}
\end{align}
where $q(x, t, a) := \mE[r\,|\, x, t, a]$ is the expected reward function at time $t$. Note that we do not impose any specific requirement about the structure of the expected reward $q(x, t, a)$, so this formulation encompasses both gradual and abrupt non-stationarity. 

To estimate the value of $\pi_e$ for a given target time $t' \,(> T)$ , we can use historical logged data $\calD$ collected by a logging policy $\pi_0(a\,|\,x, t)$ in the past in the following process:
\begin{align*}
    \calD &
    := \{ (x_i, t_i, a_i, r_i) \}_{i=1}^{n} \sim \prod_{i=1}^{n} p(x_i, t_i) \pi_0(a_i\,|\,x_i, t_i) p(r_i \,|\, x_i, t_i, a_i),
\end{align*}
where $p(x,t) = 0, \forall (x,t)$ for $t > T$ because \textbf{we cannot observe any data from future distributions}. This inability to observe any data from the target time $t' (> T)$ makes the F-OPE problem highly non-trivial, and thus the goal of this work is to provide a simple method that is substantially better than existing ideas, rather than to develop a perfect or optimal method in this setup.

It should be noted here that we quantify the accuracy of an estimator by the mean-squared-error (MSE), which can be decomposed into the squared bias and variance as follows.
\begin{align*}
    \mse (\hat{V}_{t'}(\pi_e)) 
    &:= \mE_{\calD}\left[ \left( \hat{V}_{t'}(\pi_e; \calD) - V_{t'}(\pi_e) \right)^2 \right] \\
    &= \bias \left(\hat{V}_{t'}(\pi_e; \calD)\right)^2 + \var\left[ \hat{V}_{t'}(\pi_e; \calD) \right].
\end{align*} 
where $\mE_{\calD}[\cdot]$ takes the expectation over historical logged data $\calD$, and the bias and variance terms are defined as follows.
\begin{align*}
    \bias \left[ \hat{V}(\pi; \calD) \right] &:= \mE_{\calD} \left[ \hat{V}(\pi; \calD) \right] - V(\pi) \\
    \var \left[ \hat{V}(\pi; \calD) \right] &:= \mE_{\calD} \left[ \left( \hat{V}(\pi; \calD) - \mE_{\calD} \left[ \hat{V}(\pi; \calD) \right] \right)^2 \right].
\end{align*} 

\subsection{The OPFV Estimator}
A core challenge in our estimation problem is that we cannot observe any data $(x,t,a,r)$ from the target time $t' (> T)$ because the historical dataset $\calD$ was collected in the past, specifically within the time range $t \in [0,T]$.
Despite this seemingly intractable situation, we propose the OPFV estimator, which enables a more accurate estimation of the future value of evaluation policies using only $\calD$. The key behind our estimator lies in the utilization of time series features denoted by the function $\phi(t)$. Essentially, $\phi$ can be any function that forms a clustering of time $t$ with typical choices including the day of the week, the month, the season, or a combination of these.\footnote{For example, to formulate a weekly effect, we can use the function $\phi: [0, t'] \rightarrow \{\text{Monday}, \text{Tuesday}, \ldots, \text{Sunday}\}$. Note that $\phi$ may not necessarily represent some type of seasonal effects; we can use any function as $\phi$, and we will later discuss how to optimize it based only on historical data.}
Even if the target time $t' (> T)$ does not appear in the historical data $\calD$, we may still find many timestamps observed in $\calD$ that share the same time features as the target time, i.e., $i: \phi(t_i) = \phi(t')$. 
Leveraging this structure in the time series, we can estimate the future value of new policies at least substantially more accurately than conventional methods such as IPS, DR, and Prognosticator of Chandak et al.~\citep{chandak2020optimizing}, if not perfectly.

To develop a new estimator, for a given time feature function $\phi$, we consider the following decomposition of the reward function:
\begin{align}
    q(x, t, a) = \underbrace{g(x, \phi(t), a)}_{\textit{time feature effect}} + \underbrace{h(x, t, a)}_{\textit{residual effect}}, \label{eq:decomposition}
\end{align}
where we decompose $q(x, t, a)$ into the \textit{time feature effect} and \textit{residual effect}. 
Note that we do not impose any restrictions on the functional forms of $g$ and $h$, and thus Eq.~\eqref{eq:decomposition} is not an assumption.

Based on Eq.~\eqref{eq:decomposition}, even when considering a target time in the future, we can unbiasedly estimate the time feature effect, by applying importance weighting to the datapoints that share the same feature with the target time ($i: \phi(t_i) = \phi(t')$). This idea leads us to propose the OPFV estimator:
\begin{align}
    \opfv := \meanN \Bigg\{ & \frac{\ind_{\phi}(t_i, t')}{p(\phi(t'))} \frac{\pi_e(a_i\,|\,x_i, t')}{\pi_0(a_i\,|\,x_i, t_i)} \Big( r_i  - \hat{f}(x_i, t_i, a_i) \Big) \notag  \\
    & \qquad\qquad + \mE_{\pi_e(a|x_i, t')} \left[ \hat{f}(x_i, t', a) \right]\Bigg\}, \label{eq:opfv}
\end{align}
where $\ind_{\phi}(t, t') := \ind \{ \phi(t) = \phi(t') \}$ denotes the indicator that outputs $1$ if $t$ and $t'$ have an identical time feature and $0$ otherwise. $p(\phi(t)) := \int_{s \in [0,T]} p(s) \ind_{\phi}(s, t) ds$ is the marginal probability density of $\phi$. OPFV consists of two terms where the first term applies the new weight $\ind_{\phi}(t_i, t') / p(\phi(t'))$ in addition to the original importance weight $\pi_e(a\,|\,x,t')/\pi_0(a\,|\,x,t_i)$ to estimate the time feature effect non-parametrically by using only the samples in $\calD$ whose time feature is identical to that of the target time $t'$. The second term aims to deal with the residual effect in Eq.~\eqref{eq:decomposition} based on a reward regressor $\hat{f}$ obtained via solving $\min_f \sum_{(x,t,a,r) \in \calD} (r - f(x,t,a))^2$. We do not expect this regressor to be highly accurate, but it often reduces bias compared to completely ignoring the residual effect.

Intuitively, OPFV can estimate the future policy value at arbitrary target time $t'$, even in the distant future, with much greater precision than existing methods, which do not leverage any structure within time-series. Another advantage of OPFV is its simplicity; it can readily be implemented on top of DR~\citep{dudik2011doubly} by applying the additional importance weighting factor regarding the time-series feature, yet leading to substantial improvements in the F-OPE problem as we will demonstrate. In addition, our estimator not only enables much more accurate estimations, but also enables a novel and more insightful analysis of the F-OPE problem as done below.

\subsection{Theoretical Analysis} \label{sec:analysis}
To analyze the bias and variance of OPFV, we are based on the following regularity conditions.

\begin{assumption}[Common Support]
    \label{ass.common_support}
    The logging policy $\pi_0$ is said to satisfy the common support for a given $\pi_e$ and $t'$ if $\pi_e(a \,|\, x, t') > 0 \implies \pi_0(a \,|\, x, t) > 0$ for any $x \in \calX$, $t \in [0, T]$, and $a \in \calA$.
\end{assumption}
Condition~\ref{ass.common_support} requires that $\pi_0$ assigns non-zero probabilities to the actions whose probabilities are positive under $\pi_e$, which is a standard condition in most OPE estimators~\citep{dudik2011doubly,precup2000eligibility,sachdeva2020off}.

\begin{assumption}[Common Time Feature Support]
    \label{ass.common_time_feature_support}
    The time feature function $\phi$ is said to satisfy the common time feature support for a given target time $t'$ if $p(\phi(t')) > 0$.
\end{assumption}

Condition~\ref{ass.common_time_feature_support} requires that we can find a datapoint $i$ that shares the same time feature with the target time with some positive probability. Given that we can define $\phi$ by ourselves, this condition is within our control.

We can now derive the bias and variance of OPFV as follows.
\begin{theorem}[Bias Analysis]
    \label{thm:bias-opfv}
    Under Conditions~\ref{ass.common_support} and~\ref{ass.common_time_feature_support}, OPFV has the following bias.
    \begin{align}
        \bias \left( \opfv \right) = \mE \Bigg[ \frac{\ind_{\phi}(t, t')}{p(\phi(t'))} \left( \Delta_q(x, t, t', a) - \Delta_{\hat{f}}(x, t, t', a) \right) \Bigg], \label{eq:bias-of-opfv}
    \end{align}
    where the expectation is taken over $p(x,t) \pi_e(a\,|\,x, t')$. $\Delta_q(x, t, t', a) := q(x, t, a) - q(x, t', a)$ is the relative difference in the expected rewards between timestamps $t, t'$ given $x$ and $a$, and $\Delta_{\hat{f}}(x, t, t', a) := \hat{f}(x, t, a) - \hat{f}(x, t', a)$ is its estimate based on $\hat{f}(x,t,a)$. See Appendix~\ref{proof_bias_OPFV} for the proof.
\end{theorem}

\begin{proposition}[Variance of OPFV]
    \label{prop.variance_OPFV_true_p_phi}
    Under Conditions~\ref{ass.common_support} and~\ref{ass.common_time_feature_support}, the variance of OPFV is
    \begin{align}
        & n \var \left[ \opfv \right]  \notag\\
        & =  \mE_{p(x,t) \pi_0(a|x, t)} \left[ \left( \frac{\ind_{\phi}(t, t')}{p(\phi(t'))} \frac{\pi_e(a|x, t')}{\pi_0(a|x, t)} \right)^2 \sigma^2(x, t, a) \right] \notag\\
        &\quad + \mE_{p(x,t)} \Bigg[ \left( \frac{\ind_{\phi}(t, t')}{p(\phi(t'))} \right)^2 \!\!\! \mV_{\pi_0(a|x, t)} \bigg[ \frac{\pi_e(a|x, t')}{\pi_0(a|x, t)} \Delta_{q, \hat{f}}(x, t', a) \bigg] \Bigg] \notag\\
        &\quad+ \mV_{p(t)} \left[ \frac{ \ind_{\phi}(t, t')}{p(\phi(t'))} \right]\mE_{p(x)}\left[ \mE_{\pi_e(a|x, t')} \left[ \Delta_{q, \hat{f}}(x, t', a) \right]^2 \right] \notag\\
        &\quad+ \mV_{p(x)} \left[ \mE_{\pi_e(a|x, t')} \left[q(x, t', a) \right] \right] \label{eq:var-of-opfv}
    \end{align}
    where $\Delta_{q, \hat{f}}(x, t, a) := q(x, t, a) - \hat{f}(x, t, a)$ is the estimation error of $\hat{f}$, and $\sigma^2(x, t, a) := \mV[r\,|\, x, t, a]$ is the conditional variance of the reward. See Appendix~\ref{proof_variance_OPFV_true_p_phi} for the proof.
\end{proposition}

Theorem~\ref{thm:bias-opfv} shows that there are two key factors that characterize the bias of OPFV.
The first factor is the coarseness of the time feature function.
We can see in Eq.~\eqref{eq:bias-of-opfv} that the bias occurs only when time $t \in [0,T]$ shares the same time feature with the target time, i.e., $i: \ind_{\phi}(t, t')=1$. 
Therefore, when we use a fine-grained time feature $\phi$ and $\ind_{\phi}(t, t')=1$ is less likely, the bias of OPFV becomes closer to zero.
The second factor is the accuracy of the regressor $\hat{f}$ used in OPFV regarding the estimation of $\Delta_q(x, t, t', a)$ for pair of timestamps $t$ and $t'$ that share the same time feature. 
When $t$ and $t'$ have the same time feature, i.e., $\phi(t)=\phi(t')$, the difference in their expected rewards is attributed to the difference in their residual effects, i.e., $\Delta_q(x, t, t', a) = h(x,t,a) - h(x,t',a)$.
Therefore, the second factor $\Delta_q(x, t, t', a) - \Delta_{\hat{f}}(x, t, t', a)$ characterizes how accurately the regressor $\hat{f}$ estimates the variation of the residual effect.
Overall, Theorem~\ref{thm:bias-opfv} suggests that the bias of OPFV arises only from the estimation error of the residual effect, because the importance weighting term already unbiasedly estimates the time feature effect. 
As the time feature becomes finer, the residual effect becomes smaller, so does the bias of OPFV. Appendix~\ref{appendix.two_stage_regression} takes advantage of this bias analysis and discusses how to optimize the regression model $\hat{f}$ to directly minimize the bias of OPFV.

Next, Proposition~\ref{prop.variance_OPFV_true_p_phi} suggests that the variance of the weight $\ind_{\phi}(t, t')/p(\phi(t'))$ contributes to the variance of OPFV. This observation implies that the granularity of the time feature function $\phi$ influences the variance of OPFV via the variance of $\ind_{\phi}(t, t')/p(\phi(t'))$. Specifically, using a coarser time feature leads to less variation in the weight $\ind_{\phi}(t, t')/p(\phi(t'))$, resulting in a smaller variance for OPFV. Conversely, employing a finer-grained time feature in OPFV tends to increase its variance.

\begin{figure}[t]
\centering
\vspace{1mm}
\includegraphics[clip, width=8cm]{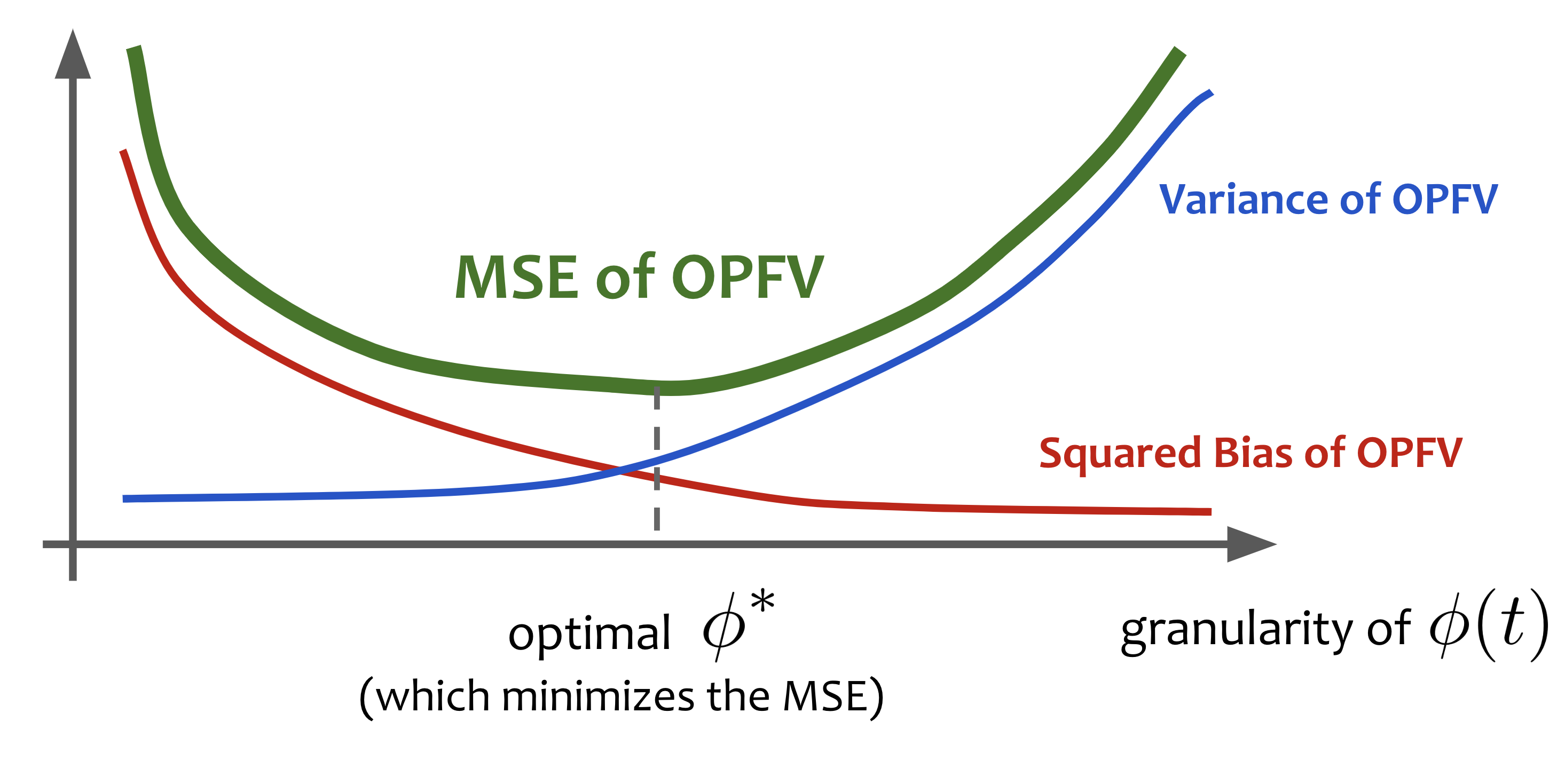}
\vspace{-2mm}
\caption{The bias-variance tradeoff of OPFV is controlled by the granularity or the cardinality of the time feature, $|\phi|$. A coarser time feature reduces variance at the cost of introducing some bias. On the other hand, a finer-grained time feature decreases bias but increases variance. An optimal time feature function $\phi^*$ that minimizes the MSE is often the one that balances the squared bias and variance.} 
\label{fig:bias-variance} 
\raggedright
\end{figure}

Coupled with the bias analysis described earlier, we now know that the granularity of the time feature function dictates the bias-variance tradeoff in OPFV. A coarser time feature reduces variance at the cost of introducing some bias. On the other hand, a finer-grained time feature decreases bias but increases variance. Therefore, optimizing the function $\phi$, is crucial, so OPFV achieves minimal MSE (as depicted in Figure~\ref{fig:bias-variance}).

\subsection{Data-Driven Optimization of Time Feature} \label{sec:tune}
The theoretical analysis tells us that the accuracy of the OPFV estimator depends on what time feature function we use. To be able to leverage this fact to further improve OPFV, we provide a data-driven method to optimize the time feature function $\phi$ using only the historical data. Ideally, we aim to optimize it via $\phi^* \in \argmin_{\phi \in \Phi} \; \mse (\opfvphi )$ where $\Phi$ is a candidate set of time feature functions that we set. However, the true MSE is unknown as it depends on the true future value ($V_{t'}(\pi_e)$), so we are based on its estimate to perform a data-driven optimization of $\phi$:
\begin{align}
   \hat{\phi} \in \argmin_{\phi \in \Phi}\; \widehat{\bias} \left( \phi \right)^2 + \widehat{\var} \left( \phi \right),
   \label{eq:tune}
\end{align}
where $\widehat{\bias} \left( \phi \right) := \opfvphi - \opfvphiinf$ is an estimate of the bias of OPFV and $\phi_{\infty}$ is the finest time feature function in $\Phi$, which leads to the lowest bias. To estimate the bias, we can also use other methods with theoretical guarantees such as the ones proposed by~\citep{su2020doubly,su2020adaptive,udagawa2023policy,felicioni2024autoope,cief2024cross,saito2024potec,sachdeva2024off}.
The variance of OPFV can be estimated by its sample variance~\citep{wang2017optimal}. Note that it is straightforward to satisfy Condition~\ref{ass.common_time_feature_support} even after Eq.~\eqref{eq:tune} by including only the candidate functions that satisfy this condition in $\Phi$. In addition, Eq.~\eqref{eq:tune} is akin to hyperparameter optimization in supervised learning, and its computational cost is within our control provided that we can control the cardinality $|\Phi|$.

\subsection{Extension to Future Off-Policy Learning}\label{sec:f-opl}
We have thus far focused on evaluating the future value of a new policy in a non-stationary environment. In this section, we extend the problem of F-OPE to F-OPL, aimed at optimizing the expected reward in the future. Specifically, for the F-OPL problem, we aim to optimize the expected reward at some future target time $t' \, (> T)$. In particular, we use a policy-based approach to learn the parameter $\zeta$ of a parameterized policy $\pi_{\zeta}(a\,|\,x)$ to maximize the future value, i.e., $\zeta^* = \argmax_{\zeta}\, V_{t'} (\pi_{\zeta})$. The policy-based approach updates the policy parameter via iterative gradient ascent as $\zeta_{\tau + 1} \gets \zeta_{\tau} + \eta \nabla_{\zeta} V_{t'}(\pi_{\zeta_{\tau}})$, where $\eta$ is the learning rate. Since we have no access to the true policy gradient $\nabla_{\zeta} V_{t'}(\pi_{\zeta}) = \mE_{p(x|t')\pi_{\zeta}(a|x)} \left[ q(x,t',a) \nabla_{\zeta} \log \pi_{\zeta}(a\,|\,x) \right]$, we need to estimate it using historical data while accounting for non-stationarity. We can achieve this as
\begin{align}
    & \opfvgrad  \notag \\
    &:= \meanN \Bigg\{ \frac{\ind_{\phi}(t_i, t')}{p(\phi(t'))} \frac{\pi_{\zeta}(a_i\,|\,x_i)}{\pi_0(a_i\,|\,x_i, t_i)} \Big( r_i - \hat{f}(x_i, t_i, a_i) \Big) \nabla_{\zeta} \log \pi_{\zeta}(a_i\,|\,x_i) \notag \\
    & \hspace{2.5cm} +  \mE_{\pi_{\zeta}(a|x_i)} \left[ \hat{f}(x_i, t', a) \nabla_{\zeta} \log \pi_{\zeta}(a\,|\,x_i) \right]\Bigg\}. \label{eq:opfv-pg}
\end{align}
Appendix~\ref{appendix.non_stationary_OPL} derives the bias and variance of this policy-gradient estimator where we obtain similar theoretical observations with those for the analysis regarding F-OPE. Note that we call the policy gradient (PG) method for F-OPL based on the estimator in Eq.~\eqref{eq:opfv-pg} as OPFV-PG. It is also straightforward to extend OPFV-PG to incorporate pessimism by combining it with relevant methods~\citep{jeunen2021pessimistic,kumar2020conservative,liang2022local,ma2019imitation}.

\subsection{Discussion on Relevant Techniques}
This section discusses some relevant techniques around OPE that might look similar to the contributions of this work.

First, we differentiate our contributions from those of \citet{uehara2020off}, which studied OPE and OPL under covariate shift, where the context distributions differ between the logged data ($p^{\text{hist}}(x)$) and in the evaluation environment ($p^{\text{eval}}(x)$), while the reward distribution $p(r|x,a)$ does not change. Our setup might appear similar to that of \citet{uehara2020off} when we formulate the combined feature $\tilde{x} = (x,t)$. However, the most critical difference is that our focus is on evaluating and optimizing the policy value for \textit{future} times. This implies that our target time $t' (> T)$ has zero density in the logged data $\calD$, making the density ratio $p^{\text{eval}}(\tilde{x}')/p^{\text{hist}}(\tilde{x}')=p^{\text{eval}}(x,t')/p^{\text{hist}}(x,t')$ nonexistent. In contrast, \citet{uehara2020off} assume the existence of the density ratio of $p(\tilde{x})$ (i.e., the assumption of strong overlap, $p^{\text{eval}}(\tilde{x}') >0 \implies p^{\text{hist}}(\tilde{x}') > 0$ as described in their Assumptions 1 and 2), a condition our setup inherently does not satisfy as $p^{\text{hist}}(x,t') = 0$ for all $x \in \calX$ due to the inability to observe any data from future distributions. We have managed to address and analyze this seemingly intractable situation for the first time through the concepts of reward function decomposition and importance weighting regarding the time feature function $\phi$. Thus, our problem setup, approach, and associated analysis (particularly the bias analysis given in Theorem 3.3) are conceptually and technically different from \citet{uehara2020off}. Another notable difference is that \citet{uehara2020off} assumes access to the realizations of the context from the distribution in the evaluation period ($p^{\text{eval}}(\tilde{x}')$) to estimate the density ratio, but this is not possible in our setup where we cannot observe any data from the future distributions. Therefore, the method from \citet{uehara2020off} (and its variants) is simply infeasible in our problem of F-OPE.

Second, we discuss critical differences between our contributions and those of \citet{saito2023off}, which develops the OffCEM estimator to deal with OPE in large action spaces in a stationary setup. Even though our main idea of reward function decomposition is inspired by OffCEM, its application to deal with the non-stationarity is non-trivial and was never mentioned in \citet{saito2023off}. Moreover, we discuss how to identify an appropriate time-future function in a data-driven fashion as in Section~\ref{sec:tune} and perform associated empirical evaluations, which are missing in \citet{saito2023off}. We also propose an extension of our estimator to an OPL method while \citet{saito2023off} focused solely on the OPE problem. Therefore, our work is the first to formulate the F-OPE/L problem and offer multiple unique contributions methodologically and empirically.

\section{Empirical Evaluation} \label{sec:experiment}
This section empirically compares our proposed methods with existing methods on both synthetic and real-world data.\footnote{The experiment code is available at https://github.com/sony/ds-research-code/tree/master/kdd2025-opfv .}

\begin{figure*}[t]
\centering
\includegraphics[clip, width=13cm]{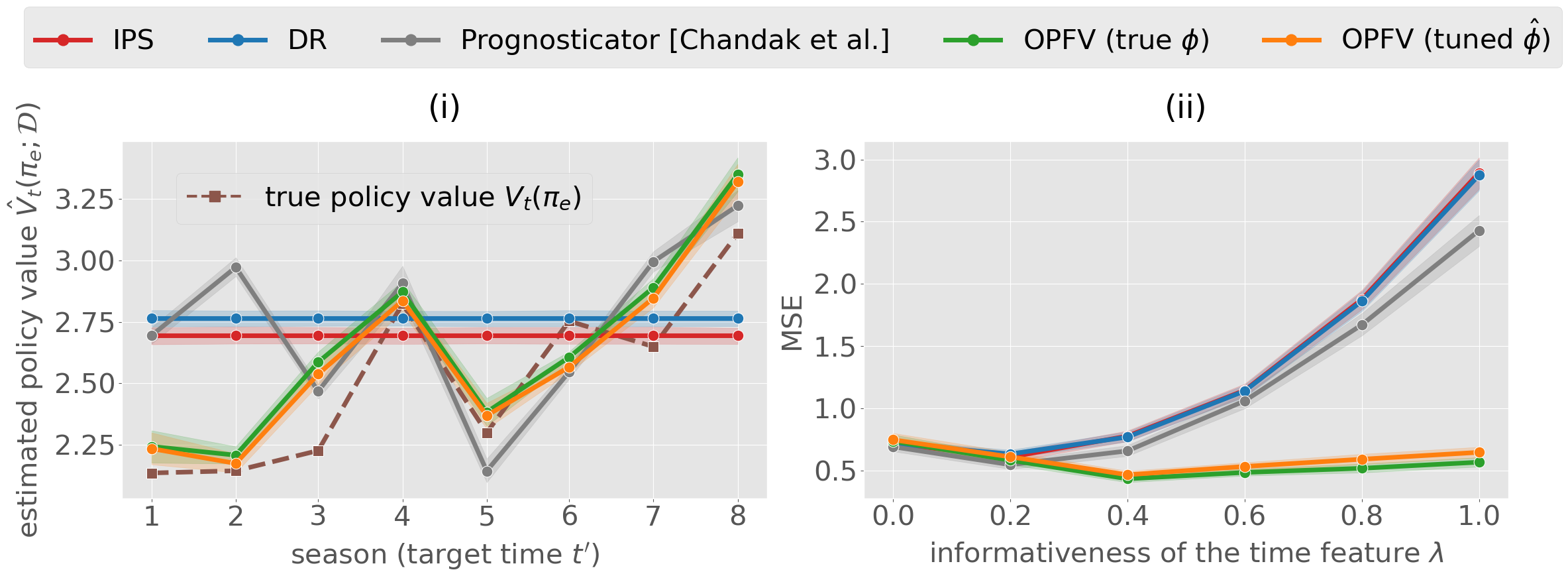}
\vspace{-2mm}
\caption{Comparing (i) the true future policy value $V_t(\pi_e)$ and the estimated future policy values $\hat{V}_t(\pi_e; \calD)$ with varying target times $t'$ in the future and (ii) the MSE of estimators with varying $\lambda$ values that control the informativeness of the time feature effect as in Eq.~\eqref{eq:synthetic_reward_function}.} \label{fig:f-ope-target-time-lambda} \vspace{-2mm}
\raggedright
\end{figure*}

\begin{figure*}[t]
\centering
\includegraphics[clip, width=13cm]{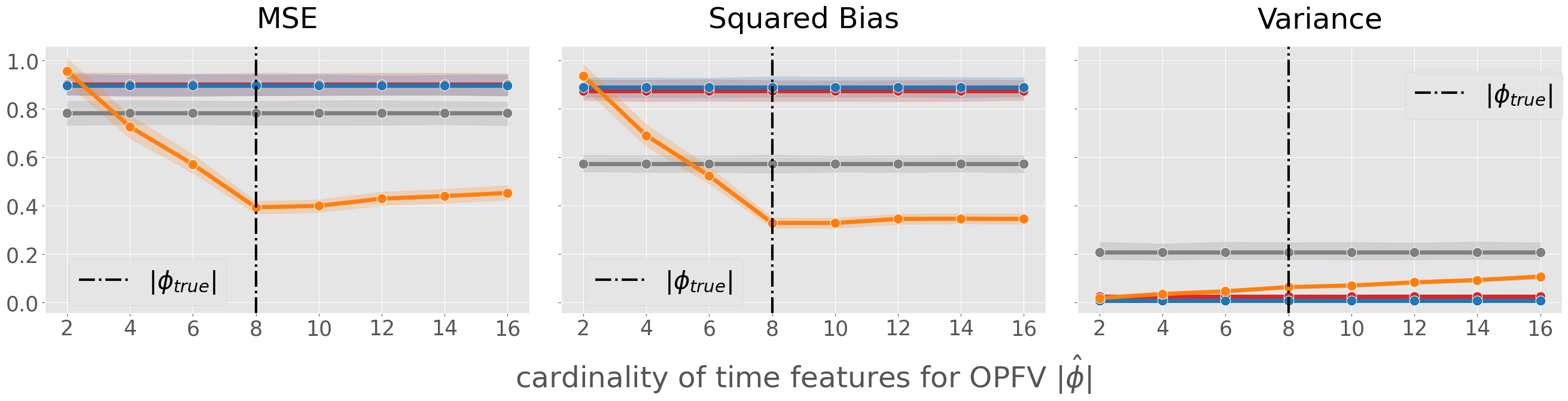}
\vspace{-2mm}
\caption{Comparison of the MSE, squared bias, and variance of estimators with varying cardinalities of time features $|\hat{\phi}|$ used in OPFV (which can be different from the true time feature $\phi_{\text{true}}$ that defines the reward function) in Eq.~\eqref{eq:synthetic_reward_function} where the cardinality of the true time features is fixed at eight.} \label{fig:f-ope-num-time-feature} \vspace{-2mm}
\raggedright
\end{figure*}

\begin{figure*}[t]
\centering
\includegraphics[clip, width=13cm]{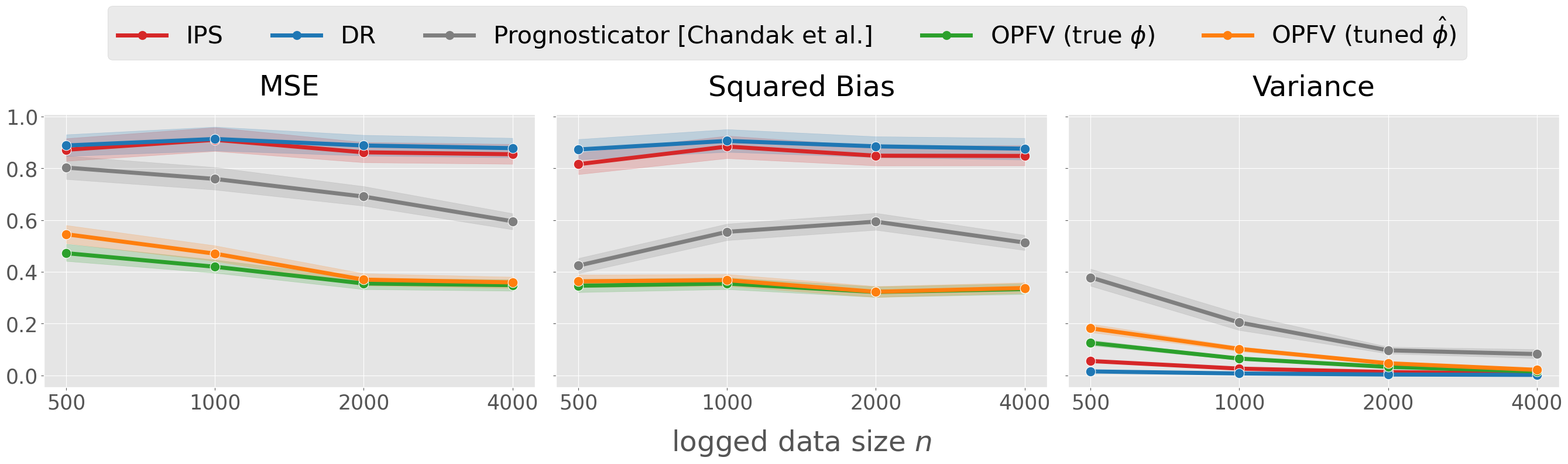}
\vspace{-2mm}
\caption{Comparison of the MSE, squared bias, and variance of estimators in F-OPE with varying logged data sizes $n$.} \vspace{-5mm} \label{fig:f-ope-n-trains}
\raggedright
\end{figure*}

\paragraph{Synthetic Data.}
\label{sec:synthetic-data}
In our synthetic experiment, we consider F-OPE and F-OPL problems where the logged data are collected during \underline{\textbf{Year 1}}, and we are interested in estimating the future values of a new policy at some target times in \underline{\textbf{Year 2}}. To generate historical datasets synthetically, we first sample Unix timestamps uniformly within the range from January 1st ($t=0$) to December 31st ($t=T$) in \underline{\textbf{Year 1}}. We then sample a 10-dimensional context vector $x$ from the standard normal distribution and synthesize the expected reward function as
\begin{align}
    q(x, t, a; \lambda) = \lambda \cdot g(x, \phi_{\text{true}}(t), a) + (1 - \lambda) \cdot h(x, t, a), \label{eq:synthetic_reward_function}
\end{align}
where $\lambda$ is an experimental parameter that controls the informativeness of the true time feature $\phi_{\text{true}}$. The true time feature considers seasonality, evenly dividing a year into eight different seasons (i.e., $\phi_{\text{true}}: t \rightarrow \{\text{season 1},\ldots,\text{season 8}\}$).\footnote{Specifically, ``season 1'' corresponds to Jan 1st to Feb 14th, and ``season 8'' corresponds to Nov 16th to Dec 31st. These eight seasons are repeated in Year 1 and Year 2.} Note that the $h$ function in Eq.~\eqref{eq:synthetic_reward_function} produces the effect of individual timestamp $t$, so \textbf{timestamps that have the same time feature do not have the same expected reward, producing non-stationarity}. The concrete definitions of the $g$ and $h$ functions are provided in Appendix~\ref{app:additional_setup}.

We then synthesize the logging policy $\pi_0$, which produced the historical dataset $\calD$ during Year 1, by applying the softmax function to the expected reward $q(x, t, a)$ over the range $t \in [0, T]$ as
\begin{align*}
    \pi_0(a \,|\, x, t; \beta) = \frac{\exp\left(\beta \cdot q(x, t, a)\right)}{\sum_{a' \in \calA} \exp\left(\beta \cdot q(x, t, a')\right)},
\end{align*}
where $\beta$ is an experimental parameter to define $\pi_0$. We use $\beta = 0.1$ and set the cardinality of the action space to $|\calA| = 10$ throughout the experiment. Lastly, we sample the reward $r$ from a normal distribution with mean $q(x, t, a)$ and a standard deviation of $\sigma = 1.0$. The logged data $\calD = \{ (x_i, t_i, a_i, r_i) \}_{i=1}^n$ is generated by repeating the above sampling procedure $n$ times independently.

We employ an epsilon-greedy policy to define $\pi_e$ that is to be deployed in the future ($t' > T$):
\begin{align*}
    \pi_e(a \,|\, x, t';\epsilon) = (1 - \epsilon) \cdot \Big\{ a = \argmax_{a' \in \calA} q(x, t', a') \Big\} + \epsilon/|\calA|,
\end{align*}
where $\epsilon \in [0, 1]$ is a noise parameter that determines the quality of $\pi_e$. We default to using $\epsilon = 0.2$.

For F-OPE, we compare OPFV with both true ($\phi_{\text{true}}$) and tuned ($\hat{\phi}$) time features against OPE estimators for stationary environments (IPS and DR) and Prognosticator~\citep{chandak2020optimizing} as baselines. For F-OPL, we compare OPFV-PG with the regression-based (RegBased) and policy-based (IPS-PG and DR-PG) methods as baselines for stationary environments, and Prognosticator as a baseline that considers non-stationarity. The baseline methods in the F-OPL experiment are defined rigorously in Appendix~\ref{app:baselines-methods-for-F-OPL}. We use a Random Forest model to estimate $q(x, a)$ for DR and $q(x, t, a)$ for OPFV. Note that we use $n=1000$ for F-OPE, $n=8000$ for F-OPL, $\lambda=0.5$ for both F-OPE and F-OPL experiments as default configurations.

Figure~\ref{fig:f-ope-target-time-lambda} (i) illustrates the true future values of $\pi_e$ and the respective estimated values for varying target times $t'$, spanning from season 1 to season 8 in Year 2, which is after the end of the collection of the logged data.\footnote{Note that here we plot the true future policy values and estimated policy values averaged within each season to ease the discussion, but there also exists non-stationarity within each season in our experiments, as we can see in Eq.~\eqref{eq:synthetic_reward_function}.} At each target time $t'$, OPFV with both true ($\phi_{\text{true}}$) and tuned ($\hat{\phi}$) time features consistently tracks the fluctuations of the true future value $V_{t'}(\pi_e)$ by selectively leveraging the samples whose timestamps share the identical time feature with $t'$. In contrast, Prognosticator occasionally fails substantially, such as seasons 1 and 2, due to its unreliable regression phase. Figure~\ref{fig:f-ope-target-time-lambda} (ii) compares the estimators' accuracy under varying levels of informativeness of the time features ($\lambda$ in Eq.~\eqref{eq:synthetic_reward_function}). The figure clearly shows that, with $\lambda$ increasing, OPFV becomes increasingly more accurate than other estimators that do not leverage the time features. In contrast, when the time features are less useful ($\lambda \le 0.2$), all estimators perform similarly, which could be seen as a limitation of our method; it performs similarly to the conventional methods if any useful structure within time-series does not exist. However, this also suggests that the worst case performance of OPFV with no useful structure in the time-series reduces to that of existing methods, and therefore there is no reason to rely on existing methods under non-stationarity. We can also see that OPFV with the tuned time feature ($\hat{\phi}$) is almost as effective as OPFV with the true time feature ($\phi_{\text{true}}$), suggesting our methods' applicability even without the availability of the true function $\phi_{\text{true}}$.

Figure~\ref{fig:f-ope-num-time-feature} shows the effect of the cardinality of the time features $|\hat{\phi}|$ used for OPFV on its bias-variance tradeoff. Note that we change only the cardinality of the time features used in OPFV, $|\hat{\phi}|$, while the true time feature $\phi_{\text{true}}$ that defines the reward function in Eq.~\eqref{eq:synthetic_reward_function} is fixed in this experiment. We can see from the figure that, as the cardinality of the used time features increases, the bias of OPFV decreases until the time feature becomes finer than the true one (i.e., $|\hat{\phi}| \ge |\phi_{\text{true}}| = 8$), which is consistent with our bias analysis in Theorem~\ref{thm:bias-opfv}. Conversely, the variance of OPFV slightly increases with the cardinality of the time features, which is also consistent with our theoretical observations. It also implies that the choice of $\phi$ may greatly affect the MSE of OPFV, which highlights the importance of the data-driven optimization procedure, which we described in Section~\ref{sec:tune}.

\begin{figure}
    \centering
    \includegraphics[clip, width=8.5cm]{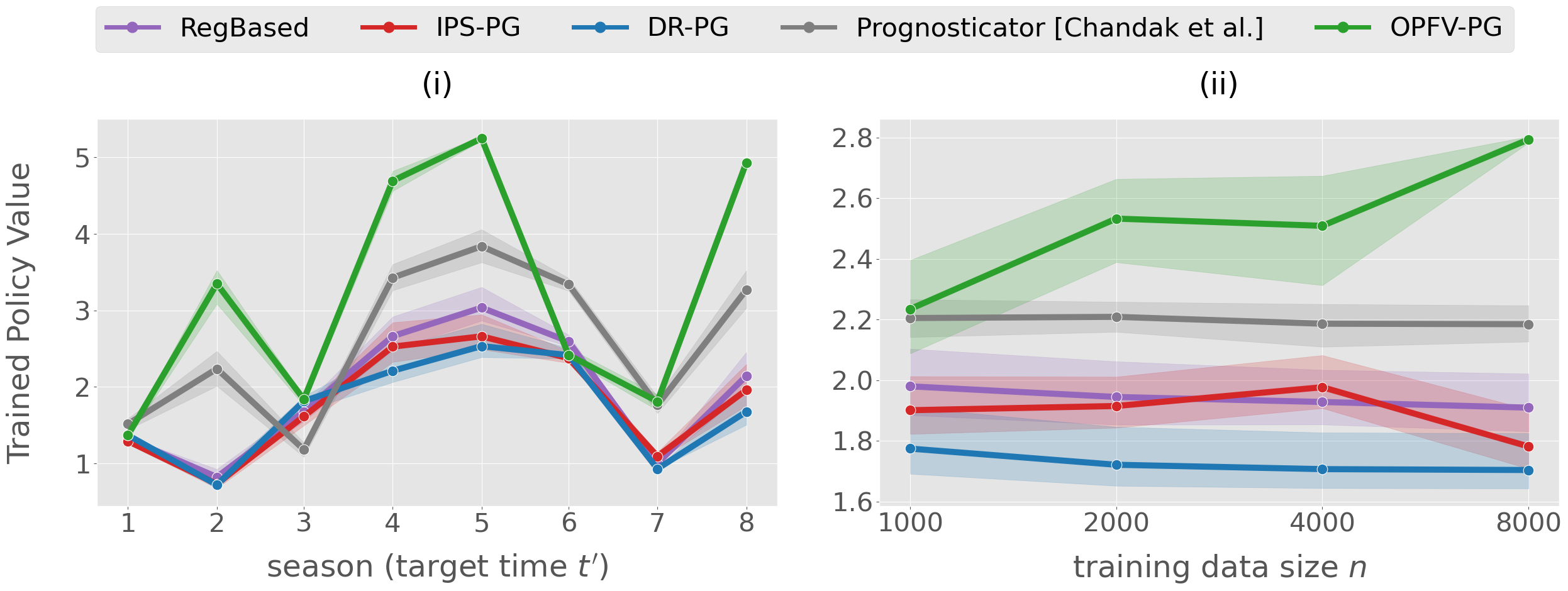}
    \vspace{-3mm}
    \caption{Comparison of the policy values (higher the better) achieved by F-OPL methods with (left) varying target times $t'$ and (right) training data sizes $n$.} \label{fig:f-opl}
\end{figure}

\begin{table*}[t]
    \caption{Comparison of the future policy values learned by F-OPL methods in the test set of the Kuairec dataset. Mean and standard deviation of the future policy values over 10 runs are reported.}
    \vspace{-2mm}
    \centering
    \scalebox{1}{
        \begin{tabular}{l|cccccc}
        \toprule
         & RegBased & IPS-PG & DR-PG & Prognosticator & OPFV-PG (\textbf{w/o} tuned $\phi$) & OPFV-PG (\textbf{w/} tuned $\phi$) \\
        \midrule
        Mean Value($\pm$ SD) & 1.48($\pm$ 0.31) & 1.02($\pm$ 0.70) & 1.15($\pm$ 0.57) & 1.39($\pm$ 0.59) & 1.60($\pm$ 0.56) & \textbf{1.67($\pm$ 0.63)} \\
        \bottomrule
        \end{tabular}}
    \label{tab:real_data}
\end{table*}

Figure~\ref{fig:f-ope-n-trains} reports the accuracy of the estimators when we vary the logged data sizes from $500$ to $4000$. Overall, OPFV performs more accurately than the baseline methods irrespective of the data size, because it can at least unbiasedly estimate the time feature effect, leading to the lowest bias as we can see in the middle figure. Note that OPFV has higher variance than IPS and DR due to the additional weighting factor to estimate the time feature effect, but its bias reduction advantage is much larger, making OPFV the most effective. IPS and DR produce the largest bias due to their unrealistic stationarity assumption. Prognosticator fails particularly when the data size is limited due to the significant variance produced when performing OPE in each period in the historical data.

Figure~\ref{fig:f-opl} (i) presents the future policy values achieved by five OPL methods with varying target times $t'$. We observe that OPFV-PG mostly achieves the highest value adaptive to different target times $t'$ due to its effective usage of the time features in estimating the policy gradient for future values. Prognosticator rarely produces a better policy than OPFV-PG and it ends up with a policy even worse than the conventional OPL methods in season 3. In contrast, OPFV-PG always performs better than the conventional methods. Figure~\ref{fig:f-opl} (ii) reports the earned future policy values with varying training data sizes. As the training data size $n$ increases, OPFV-PG is able to learn increasingly better policies because of its low-bias estimation of the policy gradient $\nabla_{\zeta} V_{t'}(\pi_{\zeta})$ for any target time $t'$. In contrast, the baseline methods are based on biased policy gradients due to their inability to deal with non-staitonarity, so they do not necessarily improve the future policy values even with increased training data. Appendix~\ref{app:additional} provides additional results about F-OPL showing that OPFV-PG outperforms existing methods particularly when the $g$ function in Eq.~\eqref{eq:synthetic_reward_function} has larger effect (larger $\lambda$) and the cardinality of the used time feature $|\phi_{\text{true}}|$ is not too small.

\paragraph{Real-World Data.}\label{sec:real-experimnet}
This section assesses the real-world applicability of our method using the KuaiRec dataset~\citep{gao2022kuairec}, which consists of recommendation logs of the video-sharing app, Kuaishou, with timestamps from July 5th to September 5th, 2020. Each record in the dataset includes a user ID, timestamp $t$, recommended video ID as action $a$, and the watch ratio of the recommended video as reward $r \in [0, \infty)$, which represents the play duration divided by the video duration. Each user and video is associated with user and video features, which we consider as context $x$ and features of action $a$, respectively. We omit variables in the user and video features that are too sparse or have missing values. To construct an action space, we uniformly subsample 100 actions from the original set of 3327 unique videos due to computational efficiency reasons, but non-stationarity of the dataset remains after this pre-processing as we validate in Appendix~\ref{app:non-stationarity-in-kuairec}. The range of the timestamps in the logged data spans from July 5th (corresponding to $t=0$) to August 4th (corresponding to $t=T$), while the test data covers the period from August 5th to September 5th (considered as the future) to perform an F-OPL experiment under non-stationarity.

We compare the future policy values achieved by OPFV-PG (\textbf{w/} tuned $\phi$), OPFV-PG (\textbf{w/o} tuned $\phi$), Reg-based, IPS-PG, DR-PG, and Prognosticator. All methods train a policy parameterized by a neural network with three hidden layers. We used the day of the week as the time feature for OPFV-PG (\textbf{w/o} tuned $\phi$). We optimized the time feature function used for OPFV-PG (\textbf{w/} tuned $\phi$) based only on the training data following the procedure described in Section~\ref{sec:tune}, and we refer the reader to Appendix~\ref{app:real-time-feature-opt} for more details about the tuning procedure employed in the real-world experiment.

Table~\ref{tab:real_data} presents the mean and standard deviation of the values of the learned policies across 10 experimental runs with varying random seeds. It is important to note that the values in Table~\ref{tab:real_data} were estimated by averaging three distinct OPE estimators on the test set to produce reliable results. The results with the three individual estimators (prior to averaging) are detailed in Appendix~\ref{app:additional-results-by-DM-SNIPS-SNDR}, and they lead to a similar conclusion. The table illustrates that the two versions of OPFV-PG consistently achieve much higher future policy values than existing methods even under unknown and possibly complicated non-stationarity of the real-world data. In particular, the promising result of OPFV-PG (\textbf{w/} tuned $\phi$) demonstrates the real-world applicability of the optimization procedure of the time feature function presented in Section~\ref{sec:tune}. We also observe that OPFV-PG (\textbf{w/o} tuned $\phi$) generally outperforms the baseline OPL methods, which implies that our method is at least substantially more effective than existing methods that do not deal with non-stationarity or leverage the structure in the time-series, even without carefully tuning $\phi$.

\section{Conclusion}
This paper formulated the problem of \textit{future} OPE and OPL for estimating and optimizing the future policy values in non-stationary environments. We then proposed the OPFV estimator, which leverages time series features to unbiasedly estimate the time feature effect via importance weighting. We have demonstrated that OPFV is low-bias when the time feature is fine-grained or the regression component accurately estimates the pairwise differences of the residual effect, resulting in much more effective F-OPE and F-OPL compared to existing ideas.  We also showed how our estimator can readily be extended to a policy-gradient estimator to solve the future OPL problem. Experiments demonstrated that our methods outperform existing ones in a variety of environments on both synthetic data with simplified settings and real-world data with non-trivial non-stationarity.

\bibliographystyle{ACM-Reference-Format}
\balance
\bibliography{main.bbl}

\newpage
\appendix
\onecolumn

\begin{table}
  \caption{Comparing Our Proposed Method (OPFV) with Key Existing Methods}
  \vspace{-1mm}
  \centering
  \scalebox{1}{
  \begin{tabular}{c|ccc}
    \toprule
     & 
     \begin{tabular}{c}
        Thomas et al.~\citep{thomas2017predictive}, Chandak et al.~\citep{chandak2020optimizing} 
     \end{tabular}
     & Jagerman et al.~\citep{jagerman2019people} 
     & \begin{tabular}{c}
        OPFV\\ (ours) 
     \end{tabular} \\
    \midrule
    Is it applicable to smooth non-stationarity?  & \fail & \tick & \tick  \\
    Is it applicable to abrupt non-stationarity? & \tick & \fail & \tick  \\
    Does it aim to estimate future policy values? & \tick & \fail & \tick  \\
    Can it leverage time-series features? & \fail & \fail & \tick \\
    Is its bias-variance tradeoff analyzable? & \fail & \tick & \tick  \\
    \bottomrule
  \end{tabular}
  }
  \vskip 0.1in
  \raggedright
  \fontsize{8.5pt}{8.5pt}\selectfont 
  \textit{Note}: Chandak et al.~\citep{chandak2020optimizing} and Thomas et al.~\citep{thomas2017predictive} proposed the Prognosticator method for abrupt non-stationarity, while Jagerman et al.~\citep{jagerman2019people} developed sliding-window and exponential IPS for smooth non-stationarity. These existing methods cannot handle abrupt and smooth non-stationarity simultaneously, whereas our method and its associated F-OPE formulation can address arbitrary types of non-stationarity. Both Chandak et al.~\citep{chandak2020optimizing} and our work focus on estimating and optimizing the future value of policies, in contrast to Jagerman et al.~\citep{jagerman2019people}, which did not aim to handle future policy values. None of the existing works leverage time-series features as our proposed method does, resulting in them suffering from substantial bias, similar to conventional OPE under stationarity. Moreover, our method is the first to present an analyzable bias-variance tradeoff among the methods that aim to estimate future policy values.
\end{table}

\section{Related work} \label{app:related}

\paragraph{Off-policy evaluation and learning under stationarity.} 
Backgrounds of our work lie in conventional literature of off-policy evaluation (OPE) in contextual bandits~\citep{dudik2014doubly, farajtabar2018more,  kallus2021optimal, kiyohara2024off, kiyohara2023off, liu2018breaking,
 metelli2021subgaussian, saito2024long, saito2022off, saito2023off, su2020doubly, su2019cab,  wang2017optimal}. In this line of work, there exist three primary approaches. The first one, known as the Direct Method (DM)~\citep{beygelzimer2009offset}, employs an estimator of the expected reward function to estimate the policy value. Although its variance is typically much lower than other approaches, it often incurs significant bias due to challenges in accurately estimating the expected reward of every action in the action space using only partial feedback (logged bandit data). Inverse Propensity Scoring (IPS)~\citep{horvitz1952generalization}, in comparison to DM, achieves unbiasedness under the condition of common support by re-weighting the observed rewards. However, its main drawback is its high variance, attributable to the use of importance weights. Doubly Robust (DR)~\citep{dudik2011doubly} combines DM and IPS, aiming to mitigate the variance issue of IPS while retaining unbiasedness under the same assumptions as IPS. 

Note that the naive application of the DR estimator to our F-OPE problem would look like the following, where we consider the combined context $\tilde{x} := (x, t)$:
\begin{align}
    \hat{V}_{\text{DR}}(\pi_e; \mathcal{D}, \hat{q}) \notag
    &= \frac{1}{n} \sum_{i=1}^n \left\{ \frac{\pi_e(a_i\,|\,\tilde{x}_i)}{\pi_0(a_i\,|\,\tilde{x}_i)} (r_i - \hat{q}(\tilde{x}_i,a_i)) + \mathbb{E}_{\pi_e(a|\tilde{x}_i)}[\hat{q}(\tilde{x}_i,a)] \right\} \notag \\ 
    &= \frac{1}{n} \sum_{i=1}^n \left\{  \frac{\pi_e(a_i\,|\,x_i,t_i)}{\pi_0(a_i\,|\,x_i,t_i)} (r_i - \hat{q}(x_i,t_i,a_i)) + \mathbb{E}_{\pi_e(a|x_i,t_i)}[\hat{q}(x_i,t_i,a)] \right\}, \label{eq:dr-naive-application}
\end{align}
where $\hat{q}$ is the estimator of the expected reward function. The most significant difference between our OPFV estimator (Eq.~\eqref{eq:opfv}) and a version of the DR estimator introduced above is whether we consider additional importance weight $\ind_{\phi}(t,t') / p(\phi(t))$ in terms of the time feature function $\phi$. Since DR overlooks this weighting factor regarding the time-series structure, it incurs non-vanishing bias. In contrast, the bias of our OPFV estimator is presented in Theorem~\ref{thm:bias-opfv}, which can be small when $\phi$ is fine-grained and the reward model $\hat{f}$ estimates the relative difference in the expected rewards $\Delta_{q}(x,t,t',a)$ for pairs of timestamps that share the same time feature ($\phi(t)=\phi(t')$). This is because the new weighting factor ($\mathbb{I}_{\phi}(t,t')/p(\phi(t))$) unbiasedly estimates the time feature effect $g$, and the bias only arises due to the estimation error of the regression model $\hat{f}$ against the residual effect. This analysis is specific to our proposed method and is novel because our problem formulation is novel.

It is also worth differentiating our contributions from those of Uehara et al.~\citep{uehara2020off}, which studied OPE and OPL under covariate shift, where the context distributions differ between the logged data ($p^{\text{hist}}(x)$) and in the evaluation environment ($p^{\text{eval}}(x)$), while the reward distribution $p(r|x,a)$ does not change. Our setup might appear similar to that of Uehara et al.~\citep{uehara2020off} when we formulate the combined feature $\tilde{x} = (x,t)$. However, the most critical difference is that our focus is on evaluating and optimizing the policy value for \textit{future} times. This implies that our target time $t' (> T)$ has zero density in the logged data $\calD$, making the density ratio $p^{\text{eval}}(\tilde{x}')/p^{\text{hist}}(\tilde{x}')=p^{\text{eval}}(x,t')/p^{\text{hist}}(x,t')$ nonexistent. In contrast, Uehara et al.~\citep{uehara2020off} assume the existence of the density ratio of $p(\tilde{x})$ (i.e., the assumption of strong overlap, $p^{\text{eval}}(\tilde{x}') >0 \implies p^{\text{hist}}(\tilde{x}') > 0$ as described in their Assumptions 1 and 2), a condition our setup inherently does not satisfy as $p^{\text{hist}}(x,t') = 0$ for all $x \in \calX$ due to the inability to observe any data from future distributions. We have managed to address and analyze this seemingly intractable situation for the first time through the novel concepts of reward function decomposition and importance weighting regarding the time feature function $\phi$. Thus, our problem setup, approach, and associated analysis (particularly the bias analysis given in Theorem 3.3) are conceptually and technically different from Uehara et al.~\citep{uehara2020off}. Another notable difference is that Uehara et al.~\citep{uehara2020off} assumes access to the realizations of the context from the distribution in the evaluation period ($p^{\text{eval}}(\tilde{x}')$) to estimate the density ratio, but this is not possible in our setup where we cannot observe any data from the future distributions. Therefore, the method from Uehara et al.~\citep{uehara2020off} (and its variants) is simply infeasible in our problem of F-OPE.

For off-policy learning (OPL), there exist two typical approaches: regression-based and policy-based. The regression-based approach~\citep{jeunen2021pessimistic, sachdeva2020off} estimates the expected reward using off-the-shelf supervised learning methods and outputs a policy based on certain rules that take into account the estimated expected reward, such as the softmax function. Similar to OPE, this approach is vulnerable to high bias due to difficulties in accurately estimating the expected reward from only the logged data. On the other hand, the policy-based approach optimizes a parameterized policy via iterative gradient ascent based on an estimated gradient of the value with respect to the policy parameters. While we can estimate the policy gradient in an unbiased fashion under common support by applying OPE estimators such as IPS and DR, it often results in considerable variance, akin to the issue of importance weighting in OPE. Another issue that often arises in OPL is the uncertainty in learning a policy that is substantially different from the logging policy~\citep{jeunen2021pessimistic,levine2020offline}. Some recent work~\citep{liang2022local,ma2019imitation} addressed this issue by employing pessimistic approaches, such as imitating the logging policy to some extent during policy learning. Even though our primary focus is on evaluating and learning the future value of policies, it is straightforward to extend OPFV-PG to incorporate pessimism by combining it with methods proposed by~\citep{jeunen2021pessimistic, kumar2020conservative, liang2022local, ma2019imitation}.

\paragraph{Off-policy evaluation and learning under non-stationarity.}\label{sec:prognosticator}
There exist previous efforts related to non-stationary environments, and we first describe the work of Chandak et al.~\citep{chandak2020optimizing} in detail, which is the most relevant to ours.

Chandak et al.~\citep{chandak2020optimizing} is particularly interested in abrupt non-stationarity where the logged data $\calD$ can be divided into $K$ distinct subsets $\calD = \{\calD_k\}_{k=1}^K$ where the distribution remains constant within each period. Each subset $\calD_k$ is considered generated from a corresponding distribution as follows:
\begin{align*}
    \calD_k
    &:= \{(x_{i, k}, a_{i, k}, r_{i, k})\}_{i=1}^{n_k} \sim \prod_{i=1}^{n_k} p_{\red{k}}(x_{i, k}) \pi_0(a_{i, k}\,|\,x_{i, k}) p_{\red{k}}(r_{i, k}\,|\,x_{i, k}, a_{i, k}),
\end{align*}
where $n_k$ represents the logged data size in period $k$, and $n = \sum_{k \in [K]} n_k$. For this setting, Chandak et al.~\citep{chandak2020optimizing} proposed the Prognosticator algorithm for OPL, which includes an OPE estimator we regard as a baseline method to evaluate the future performance of an evaluation policy. Prognosticator first performs an estimator $\hat{V}_k(\pi_e; \calD)$ (e.g., IPS) to estimate the past policy value of $\pi_e$ within each period $k \in [K]$. For a future performance evaluation, Prognosticator employs linear regression $\hat{V}_k(\pi_e, \calD) \approx \psi(k)^\top w$, where $\psi(k) \in \mR^d$ is a basis function encoding the period index, $w$ is a regression coefficient, and $\hat{V}_k(\pi_e, \calD)$ is an estimated past policy value. The regression is performed using training data $(\Psi, Y)$ where $\Psi := (\psi(1), \cdots, \psi(K))^\top \in \mR^{K \times d}$ and $Y := (\hat{V}_1(\pi_e; \calD), \cdots, \hat{V}_K(\pi_e; \calD))^\top \in \mR^K$. The future value estimator for period $K+\delta$ ($\delta = 1,2,\ldots$) is given by:
\begin{align*}
    \hat{V}_{K+\delta}(\pi_e; \calD) = \psi(K+\delta)^\top (\Psi^\top \Psi)^{-1} \Psi^\top Y,
\end{align*}
where $w^* := (\Psi^\top \Psi)^{-1} \Psi^\top Y$ is derived from standard ordinary least squares regression. Despite its improved adaptability to non-stationarity compared to conventional OPE, this method faces limitations in both the initial estimation phase and the extrapolation (regression) phase. In the former, Prognosticator experiences significant variance when frequent distribution shifts occur due to the reduced sample size $n_k$ in each subset $\calD_k$. For the regression phase, accurately learning future trends from only the period index $k$ in the past as inputs is extremely challenging with the simple parametric regression, resulting in substantial bias as we empirically demonstrated.

Other than Chandak et al.~\citep{chandak2020optimizing}, Thomas et al.~\citep{thomas2017predictive} proposed essentially the same estimator as Prognosticator~\citep{chandak2020optimizing} and empirically demonstrated their OPE procedure on a real digital marketing dataset. Liu et al.~\citep{liu2023asymptotically} recently proposed the regression-assisted DR estimator to more accurately estimate the current value of a new policy by leveraging historical data from varying distributions under abrupt non-stationarity. Their particular focus is to leverage past data to reduce variance without introducing much bias and to construct valid confidence intervals, rather than estimating and optimizing future policy values. Chandak et al.~\citep{chandak2022off} introduced the concept of \textit{active non-stationarity}, where distributional shifts occur due to policy's interactions with the system in abrupt non-stationary environments. However, their method still relies on a two-stage estimation process similar to Prognosticator for non-stationary OPE, encountering limitations in both stages. Hong et al.~\citep{hong2021non} focused on policy learning under abrupt non-stationarity. Their algorithm learns a new policy for each cluster of distributions, where clustering is achieved through change point detection (also studied in Padakandla et al.~\citep{padakandla2020reinforcement} for the online reinforcement learning setup) or a Hidden Markov Model applied to the logged data. However, their algorithm requires online updates during deployment and, therefore, does not fall under offline learning. Moreover, it is constrained to only existing distributions in the past and is incapable of handling unknown distribution shifts that may arise in the future. Liu et al.~\citep{li2022testing} proposed a procedure for hypothesis testing of stationarity given offline data in reinforcement learning settings and a change point detection method based on the proposed testing. However, their focus was not on OPE and OPL for the future and thus is crucially different from our formulation. 

In smooth non-stationary environments, Jagerman et al.~\citep{jagerman2019people} introduced the \textit{sliding window IPS} and \textit{exponential IPS} estimators. Sliding window IPS re-weights observed rewards using data from times close to the target time, at which the value of the evaluation policy is defined. In contrast, exponential IPS uses all data, giving more weight to recent observations. Domingues~\citep{domingues2021kernel} generalized sliding window and exponential estimators mostly in online settings in reinforcement learning. However, when applied to F-OPE and OPL, the estimators and algorithms based on sliding windows and exponential discounting incur significant bias when estimating the value of an evaluation policy in the distant future. This is because they predominantly weigh recent data and thus fail to capture potential distributional shifts in the future, which might significantly differ from the distributions present when the logged data were collected. Another key formulation is distributionally robust OPL, where the training and testing data are assumed to be sampled from different distributions~\citep{kallus2022doubly, mu2022factored, si2020distributionally, xu2022towards}. This approach aims to evaluate the value of a policy in the worst-case scenarios among a set of potential distribution shifts and learns the optimal policy that maximizes this worst-case value. Consequently, the resulting policy is often excessively pessimistic in real-world applications. Additionally, this formulation typically involves only two distributions (training and testing), which is a much simpler scenario than our formulation, which considers continual distribution changes.

\section{Additional theoretical analysis of the OPFV estimator}\label{app:additional-theoretical-analysis-opfv-f-ope}
Among the estimators for OPE under stationary context and reward distributions described in the previous section, one might observe that the form of DR in Eq.~\eqref{eq:dr-naive-application} is similar to our proposed OPFV estimator in Eq.~\eqref{eq:opfv}. However, there exists a notable distinction between OPFV and DR since OPFV is particularly tailored for F-OPE. To distinguish the difference between DR and the OPFV estimator, we present the bias reduction of the OPFV estimator against DR in the following theorem.
\begin{theorem}[Bias Reduction of OPFV against DR]
    \label{thm.bias-reduction-OPFV-against-DR}
    Under Conditions \ref{ass.common_support} and \ref{ass.common_time_feature_support}, the difference in the bias between the DR and OPFV is given by:
    \begin{align*}
        &\bias (\dr) - \bias (\opfv)\\
        ={}& \underbrace{\mE_{p(x, t) \pi_e(a|x, t')} \Bigg[ \bigg( \Delta_{q}(x, t, t', a) - \Delta_{\hat{q}}(x, t, t', a) \bigg) - \frac{\ind_{\phi}(t, t')}{p(\phi(t'))} \bigg( \Delta_{q}(x, t, t', a) - \Delta_{\hat{f}}(x, t, t', a) \bigg)\Bigg]}_{(i)} \\
        +& \underbrace{\mE_{p(x, t)} \Bigg[ \sumA  (\pi_e(a|x, t) - \pi_e(a|x, t')) q(x, t, a) \Bigg]}_{(ii)} + \underbrace{\mE_{p(x, t) \pi_e(a|x, t')} [ \Delta_{\hat{q}}(x, t, t', a)  ]}_{(iii)} \\
    \end{align*}
    See Appendix \ref{proof-bias-reduction-opfv-dr} for the proof.
\end{theorem}
Theorem~\ref{thm.bias-reduction-OPFV-against-DR} reveals that three terms capture the bias reduction of OPFV against DR. Specifically, the first term (i) of the bias reduction implies that the OPFV estimator reduces the bias against DR by the expectation with respect to the evaluation policy of the estimation error of the relative difference $\Delta_{q}(x, t, t', a)$ in the expected rewards between timestamp $t$ and $t'$ when $t$ does not share the identical time feature to target time $t'$. We can attribute this bias reduction of the term (i) to the use of the new importance weight $\ind_{\phi}(t_i, t')/p(\phi(t'))$ in the OPFV estimator. The second term (ii) indicates that additional bias is incurred to DR when the evaluation policies at timestamp $t$ and target time $t'$ differ. This bias reduction arises from the difference between the parts of the DR estimator (i.e., $\pi_e(a_i|x_i, \boldsymbol{t_i})$ and $\pi_e(a|x_i, \boldsymbol{t_i})$) and the OPFV estimator (i.e., $\pi_e(a_i|x_i, \boldsymbol{t'})$ and $\pi_e(a|x_i, \boldsymbol{t'})$), which correctly account for the evaluation policy at target time $t'$. The third term (iii) demonstrates that the OPFV estimator reduces the bias against DR by the expectation with respect to an evaluation policy $\pi_e(a|x, t')$ of the relative difference $\Delta_{\hat{q}}(x, t, t', a)$ in the estimated expected rewards between timestamps $t$ and $t'$. This bias reduction is due to the fact that the OPFV correctly uses the regression of the reward at the target time $\boldsymbol{t'}$ in the second term of the estimator in Eq.~\eqref{eq:opfv}, whereas DR naively applies the estimator of the expected reward $\hat{q}(x_i, \boldsymbol{t_i}, a)$ in Eq.~\eqref{eq:dr-naive-application} without aiming to estimate the reward at the target time $t'$. Thus, the theoretical analysis of the bias reduction of OPFV against DR clearly demonstrates that OPFV evaluates the policy for the future under non-stationarity with considerably less bias than DR, notably by the use of the time feature $\phi$ and the corresponding new importance weight as well as by the correct use of the evaluation policy and regression of the reward at target time $t'$. Note that some literature on OPE includes the theoretical analyses of a sample complexity bound, but we do not include such analyses as the problem of F-OPE is distinct from the standard OPE. More specifically, we do not have access to the samples from the distributions $p(x|t')$ and $p(r|x, t', a)$ in the future $t'$. Thus, it is impossible to construct the estimator $V_{t'}(\pi_e; \calD)$ that converges to the true policy value $V_{t'}(\pi_e)$ in the future $t'$. Even in such a challenging situation of the F-OPE problem, our novel theoretical analysis rigorously demonstrates how to optimize the time feature function $\phi$ to minimize the MSE of the proposed estimator. Hence, our bias-variance analysis is distinct from the theoretical analysis of the standard OPE estimator, providing a unique contribution to the problem of F-OPE.

\section{OPFV under non-stationary context and reward distributions}\label{app:non-stationary-context-reward}
\subsection{The extended estimator for F-OPE}
\label{appendix.non_stationary_context}
In this section, we describe the extension of the OPFV estimator to the case where both context and reward are non-stationary. To leverage the logged data, we consider the following decomposition of the non-stationary distribution $p(x|t)$ of the context $x$: 
\begin{align*}
    p(x|t) = \alpha \cdot p_1(x|\phi_x(t)) + (1 - \alpha) \cdot p_2(x|t), 
\end{align*}
where $p_1(x|\phi_x(t))$ represents the part of the non-stationary distribution of the context that can be explained by the time feature $\phi_x$ for the context, $p_2(x|t)$ represents the portion of the non-stationary distribution that cannot be captured by the time feature for the context, and $\alpha$ controls how informative the time feature $\phi_x$ for the context is. Then, we can decompose the value $V_{t'}(\pi_e)$ of an evaluation policy $\pi_e$ at the target time $t'$ into the following two terms.
% \begin{align*}
%     V_{t'}(\pi_e) 
%     &= \mE_{p(x|t') \pi_e(a|x, t')} [q(x, t', a)] \\
%     &= \alpha \mE_{p_1(x|\phi_x(t')) \pi_e(a|x, t')} [q(x, t', a)] + (1 - \alpha) \mE_{p_2(x|t') \pi_e(a|x, t')} [q(x, t, a)] \\
%     &= \underbrace{G_{\phi_x(t')}(\pi_e)}_{\text{time feature effect for context}} + \underbrace{H_{t'}(\pi_e)}_{\text{residual effect for context}}, 
% \end{align*}
\begin{align*}
    V_{t'}(\pi_e) 
    &= \underbrace{G_{\phi_x(t')}(\pi_e)}_{\text{time feature effect for context}} + \underbrace{H_{t'}(\pi_e)}_{\text{residual effect for context}} \\
    G_{\phi_x(t')}(\pi_e) &:= \alpha \cdot \mE_{p_1(x|\phi_x(t')) \pi_e(a|x, t')} [q(x, t', a)]\\
    H_{t'}(\pi_e) &:= (1 - \alpha) \cdot \mE_{p_2(x|t') \pi_e(a|x, t')} [q(x, t', a)], 
\end{align*}
where the time feature effect for context $G_{\phi_x(t')}(\pi_e)$ captures the value for the context whose non-stationarity in the distribution can be explained by the time feature $\phi_x$ for context $x$ while the residual effect for context $H_{t'}(\pi_e)$ represents the value for context whose non-stationarity in its distribution cannot be explained by the time feature $\phi_x$ for context. Note that we have decomposed the expected reward function $q(x, t', a)$ into the time feature effect for reward $g(x, \phi_r(t'), a)$ and the residual effect for reward $h(x, t', a)$ in Eq.~\eqref{eq:decomposition}. Thus, under the non-stationarity in the context distribution, we can consider the reward decomposition in Eq.~\eqref{eq:decomposition} for both the time feature effect and residual effect for context. Then, leveraging some data that have the same time feature $\phi_x(t_i)$ for context with the target time $t'$, we can extend the definition of the OPFV estimator in Eq.~\eqref{eq:opfv} to be able to estimate the time feature effect for context $G_{\phi_x(t')}(\pi_e)$ as follows.

\begin{align}
    \opfv &= \meanN \Bigg\{ \frac{\ind_{\phi_{x, r}}(t_i, t')}{p(\phi_{x, r}(t'))} \frac{\pi_e(a_i|x_i, t')}{\pi_0(a_i|x_i, t_i)} \left( r_i - \hat{f}(x_i, t_i, a_i) \right) \notag \\
    & \hspace{5cm} +  \frac{\ind_{\phi_x}(t_i, t')}{p(\phi_x(t'))} \mE_{\pi_e(a|x_i, t')} [\hat{f}(x_i, t', a)] \Bigg\} \label{eq:opfv_ns_x}
\end{align}
where $\phi_{x, r}(t) := (\phi_x(t), \phi_r(t))^\top$ is a vector of time features with respect to the context and reward, $\ind_{\phi_{x, r}}(t_i, t) := \ind\{ \phi_{x, r}(t_i) = \phi_{x, r}(t) \}$ denotes the indicator that outputs $1$ if timestamps $t_i$ and $t$ share an identical time feature for context and reward, and $\ind_{\phi_x}(t_i, t) := \ind\{ \phi_x(t_i) = \phi_x(t) \}$ denotes the indicator that outputs $1$ if timestamps $t_i$ and $t$ share an identical time feature for context. The marginal probability densities of time features for context and reward $\phi_{x, r}$, and context $\phi_x$ are defined as $p(\phi_{x, r}(t)) := \int_{s \in [0, T]} p(s) \ind\{ \phi_{x, r}(s) = \phi_{x, r}(t) \} ds$ and $p(\phi_x(t)) := \int_{s \in [0, T]} p(s) \ind\{ \phi_x(s) = \phi_x(t) \} ds$, respectively. The extended OPFV estimator consists of two terms as the original OPFV estimator in Eq.~\eqref{eq:opfv}. For the first term, the extended OPFV employs the new weight $\ind_{\phi_{x, r}}(t_i, t')/p(\phi_{x, r}(t'))$ rather than $\ind_{\phi_{r}}(t_i, t')/p(\phi_{r}(t'))$ used in the original OPFV estimator to be able to estimate the time feature effect for reward $g(x, \phi_r(t'), a)$ in the time feature effect for context $G_{\phi_x(t')}(\pi_e)$. For the second term in Eq.~\eqref{eq:opfv_ns_x}, the extended OPFV multiplies the second term of the original OPFV estimator in Eq.~\eqref{eq:opfv} by the weight with respect to the context $\ind_{\phi_x}(t_i, t') / p(\phi_x(t'))$ to capture the residual effect for reward $h(x, t', a)$ in the time feature effect for context $G_{\phi_x(t')}(\pi_e)$. Thus, the extended OPFV generalizes the original OPFV to accommodate the distributional shifts in both context and reward.

\subsection{Theoretical analysis of the extended OPFV estimator}\label{subsec:analysis_opfv_ns_context}
To theoretically analyze the extended OPFV estimator, we introduce the following condition on the distributional shift with respect to the context.
\begin{assumption}[Conditional Stationarity for Context]
    \label{ass.conditional_stationarity_wrt_context}
    A non-stationary distribution for context $p(x|t) = \alpha \cdot p_1(x|\phi_x(t)) + (1- \alpha) \cdot p_2(x|t)$ is said to satisfy conditional stationarity for context if the following holds for an arbitrary target time $t'$:
    \begin{align*}
        p_2(x|t) = p_2(x|t')
    \end{align*}
    for any $x \in \calX$ and $t \in [0, T]$ such that $\ind_{\phi_x}(t, t') = 1$
\end{assumption}
Condition \ref{ass.conditional_stationarity_wrt_context} requires that the part of the non-stationary context that can not be explained by the time feature $\phi_x$ for context is identical across the time $t$ which shares the same time feature for the context as the target time. 
Under such a condition, the extended OPFV estimator has the following bias.

\begin{theorem}[Bias of the extended OPFV estimator]
    \label{thm.bias_extended_OPFV}
    Under Conditions \ref{ass.common_support}, \ref{ass.common_time_feature_support}, and \ref{ass.conditional_stationarity_wrt_context}, the extended OPFV estimator has the following bias. 
    \begin{align}
        \bias \left( \opfv \right) = \mE_{p(x, t)\pi_e(a|x, t')}\left[ \frac{\ind_{\phi_{x, r}}(t, t')}{p(\phi_{x, r}(t'))} \left(\Delta_q(x, t, t', a) - \Delta_{\hat{f}}(x, t, t', a)\right) \right] \label{eq:bias-extended-opfv}
    \end{align}
    See Appendix \ref{proof_bias_extended_OPFV} for the proof.
\end{theorem}
The formula for the bias of the extended OPFV is similar to the bias of the original OPFV in Section~\ref{sec:analysis}. The key difference from the original OPFV is that the bias of the extended OPFV depends not only on the coarseness of the time feature for reward $\phi_r$ but also on the coarseness of the time feature for reward $\phi_x$ as the bias in Eq.~\eqref{eq:bias-of-opfv} occurs only when timestamp $t \in [0, T]$ in the logged data shares the identical time feature for reward with target time $t'$ (i.e., $\phi_r(t) = \phi_r(t')$) whereas the bias in Eq.~\eqref{eq:bias-extended-opfv} occurs when timestamp $t$ shares the identical time feature for not only reward but also context (i.e., $\phi_{x, r}(t) = \phi_{x, r}(t')$).

Furthermore, we formally introduce the following condition required to prove the unbiasedness and variance of the original OPFV estimator for non-stationary reward and the extended OPFV estimator for non-stationary context and reward distributions.

\begin{assumption}[Conditional Pairwise Correctness]
    \label{ass.conditional_piecewise_correctness}
    A regression model $\hat{f}$ is said to satisfy conditional pairwise correctness if the following holds for a given target time $t'$:
    \begin{align*}
        \Delta_q(x, t, t', a) = \Delta_{\hat{f}}(x, t, t', a), 
    \end{align*}
     for all $x \in \calX$, $a \in \calA$, and $t \in [0, T]$ such that $\ind_{\phi_r}(t, t') = 1$.
\end{assumption}

Condition~\ref{ass.conditional_piecewise_correctness} requires only the regressor $\hat{f}$ to preserve the pairwise difference of the expected reward that is produced between different timestamps $t,t'$ conditional on $x$ and $a$. This condition is less restrictive than the correct estimation of the global expected reward given $(x,t,a)$, which is the requirement of the extrapolation (regression) phase of Prognosticator.

Under such an ideal situation, we have the following proposition on the unbiasedness of the extended OPFV estimator. 

\begin{proposition}[Unibasedness of the extended OPFV estimator]
    \label{prop.unbiasedness_extended_OPFV}
    Under Conditions \ref{ass.common_support}, \ref{ass.common_time_feature_support}, \ref{ass.conditional_stationarity_wrt_context}, and \ref{ass.conditional_piecewise_correctness}, the extended OPFV estimator is unbiased for arbitrary target time $t' (>T)$, i.e., $\mE_{\calD} [\opfv] = V_{t'}(\pi_e)$. See Appendix \ref{proof_unbiasedness_extended_OPFV} for the proof.
\end{proposition}
Proposition \ref{prop.unbiasedness_extended_OPFV} suggests that the extended OPFV becomes unbiased as the time feature for the context gets fine-grained.

\section{Optimization of the regression model}
\label{appendix.two_stage_regression}
In this section, we provide how to further minimize the MSE via minimizing the error of the estimator of the reward regressor $\hat{f}$ used in our estimator in terms of both bias and variance. The bias analysis of OPFV given in Theorem~\ref{thm:bias-opfv} and variance analysis demonstrated in Proposition~\ref{prop.variance_OPFV_true_p_phi} imply that we should optimize the regressor $\hat{f}$ to minimize its error against $\Delta_q(x, t, t', a)$ for bias minimization while we should optimize $\hat{f}$ to minimize its error towards $\Delta_{q, \hat{f}}(x, t, a)$ for variance minimization. 

Specifically, we start from decomposing $\hat{f}(x, t, a)$ in the following way:
\begin{align*}
    \hat{f}(x, t, a; \theta, \omega) := \underbrace{\hat{g}\left(x, \phi(t), a; \omega\right)}_{\text{est. of time feature effect}} + \underbrace{\hat{h}(x, t, a; \theta)}_{\text{est. of residual effect}}
\end{align*}
where $\omega \in \Omega$ and $\theta \in \Theta$ are the parameters for the estimator of the time feature effect $\hat{g}$ and that of the residual effect $\hat{h}$, respectively. We use the following two-stage regression to learn the optimal parameters $\omega^*$ and $\theta^*$ that minimize the bias in Eq.~\eqref{eq:bias-of-opfv} and variance in Eq.~\eqref{eq:var-of-opfv}.
\begin{enumerate}
    \item Bias Minimization Step
    \begin{align*}
        \omega^* &\in \argmin_{\omega \in \Omega} \sum_{(x, a, t_j, t_k, r_j, r_k) \in \calD_{\pair}} \ell_{h} \Big( r_j - r_k,  \hat{h}(x, t_j, a; \omega) - \hat{h}(x, t_k, a; \omega) \Big)
    \end{align*}
    where $\calD_{\pair}$ is the augmented logged data for the pairwise regression, defined as 

    % formal expression 
    \begin{align*}
        \calD_{\pair} := \Bigg\{ (x, a, t_j, t_k, r_j, r_k) \Big| \begin{array}{l} 
          (x_j, t_j, a_j, r_j), (x_k, t_k, a_k, r_k) \in \calD \\
        x = x_j = x_k, a = a_j = a_k, \ind_{\phi}(t_j, t_k) = 1 
        \end{array} 
        \Bigg\}.
    \end{align*}
    \item Variance Minimization Step
    \begin{align*}
        \theta^* & \in \argmin_{\theta \in \Theta} \sum_{(x, t, a, r) \in \calD} \ell_{g} \Big( r, \hat{g}_l(x, \phi(t), a; \theta) + \hat{h}(x, t, a; \omega^*) \Big)
    \end{align*}
\end{enumerate}
Note that, for the reward regressor $\hat{f}(x, t', a)$ at target time $t'$ in the future, we cannot accurately estimate the residual effect $\hat{h}(x, t', a; \omega)$ unless we have the data at time $t'$ or assume strong conditions, so we randomly learn it from a neural network. On the other hand, we can learn the estimated time feature effect $\hat{g}(x, \phi(t'), a; \theta)$ from the data at time $t$ whose time feature $\phi(t)$ is identical to the time feature $\phi(t')$ at target time $t'$.

\section{Details of the optimization of the time feature function}\label{app:details-opt-time-feature}
In Section~\ref{sec:tune}, we provided the optimization procedure to tune the time feature function $\hat{\phi}$ to select the most accurate OPFV estimator. This section further covers the details regarding the satisfaction of Condition~\ref{ass.common_time_feature_support} after the optimization of the time feature and introduces some examples. Although the satisfaction of Condition~\ref{ass.common_time_feature_support} is not necessary in practice as the ultimate goal of our problem is the minimization of the MSE if we want to make sure that the tuned time feature function $\hat{\phi}$ obtained by the optimization procedure in Section~\ref{sec:tune} satisfies Condition~\ref{ass.common_time_feature_support}, then we can achieve it by considering the set $\Phi$ of the time feature functions which meet Condition~\ref{ass.common_time_feature_support}. One of the sufficient conditions for the satisfaction of Condition~\ref{ass.common_time_feature_support} is the observation of the time features $\phi(t)$ in the logged data $\calD$, which is identical to the time feature $\phi(t')$ of the target time $t'$. For instance, suppose we want to evaluate or optimize a policy for the upcoming month in 2024, given logged data collected in 2023, where we observe at least one data point on an arbitrary date in 2023. In such a situation, we can tune the time feature function $\phi$ by considering the following set
\begin{align*}
    \Phi &:= \bigg\{ 
\phi_{\text{season}}, \phi_{\text{month}}, \phi_{\text{week}}, \phi_{\text{date}}, \phi_{\text{day\_of\_week}}, 
\phi_{\text{holiday}}, \\
&\phi_{\text{season}} \bigotimes \phi_{\text{week}}, 
\phi_{\text{season}} \bigotimes \phi_{\text{date}}, 
\phi_{\text{season}} \bigotimes \phi_{\text{day\_of\_week}}, 
\phi_{\text{season}} \bigotimes \phi_{\text{holiday}}, \\
&\phi_{\text{month}} \bigotimes \phi_{\text{week}}, 
\phi_{\text{month}} \bigotimes \phi_{\text{date}}, 
\phi_{\text{month}} \bigotimes \phi_{\text{day\_of\_week}}, 
\phi_{\text{month}} \bigotimes \phi_{\text{holiday}}, \\
&\phi_{\text{week}} \bigotimes \phi_{\text{holiday}}, 
\phi_{\text{day\_of\_week}} \bigotimes \phi_{\text{holiday}} \bigg\}
\end{align*}
of the time feature functions where each function $\phi \in \Phi$ is defined as follows.
\begin{align*}
    \phi_{\text{season}} &: [0, t'] \rightarrow \{ \text{spring}, \text{summer}, \text{fall}, \text{winter} \} \\
    \phi_{\text{month}} &: [0, t'] \rightarrow \{ \text{Jan}, \text{Feb}, \cdots, \text{Dec} \} \\
    \phi_{\text{week}} &: [0, t'] \rightarrow \{ \text{1st week}, \text{2nd week}, \cdots, \text{6th week} \} \\
    \phi_{\text{date}} &: [0, t'] \rightarrow \{ \text{1st}, \text{2nd}, \cdots, \text{31st} \} \\
    \phi_{\text{day\_of\_week}} &: [0, t'] \rightarrow \{ \text{Sun}, \text{Mon}, \text{Tue}, \text{Wed}, \text{Thu}, \text{Fri},  \text{Sat} \} \\
    \phi_{\text{holiday}} &: [0, t'] \rightarrow \{ \text{holiday}, \text{not holiday} \} \\
    % \phi_{\text{month\&weekday\_weekend}} &: [0, t'] \rightarrow \{ \text{weekdays in Jan}, \text{weekends in Jan}, \cdots, \text{weekdays in Dec}, \text{weekends in Dec} \} \\
    % \phi_{\text{season\&day\_of\_week}} &: [0, t'] \rightarrow \{ \text{Sun in spring}, \cdots, \text{Sat in spring}, \cdots, \text{Sun in winter}, \cdots, \text{Sat in winter} \} \\
    % \phi_{\text{month\&day\_of\_week}} &: [0, t'] \rightarrow \{ \text{Sun in Jan}, \cdots, \text{Sat in Jan}, \cdots, \text{Sun in Dec}, \cdots, \text{Sat in Dec} \} \\
    % \phi_{\text{season\&date}} &: [0, t'] \rightarrow \{ \text{1st in spring}, \cdots, \text{31st in spring}, \cdots, \text{1st in winter}, \cdots, \text{31st in winter} \} \\
    % \phi_{\text{season\&holiday}} &: [0, t'] \rightarrow \{ \text{holiday in spring}, \text{not holiday in spring}, \cdots, \text{holiday in winter}, \text{not holiday in winter} \} \\
    % \phi_{\text{month\&holiday}} &: [0, t'] \rightarrow \{ \text{holiday in Jan}, \text{not holiday in Jan}, \cdots, \text{holiday in Dec}, \text{not holiday in Dec} \} \\
    % \phi_{\text{day\_o\_week\&holiday}} &: [0, t'] \rightarrow \{ \text{holiday and Sun}, \text{not holiday and Sun}, \cdots, \text{holiday and Sat}, \text{not holiday and Sat} \} \\
\end{align*}
where, for any $\phi_i, \phi_j \in \{ \phi_{\text{season}}, \phi_{\text{month}}, \phi_{\text{week}}, \phi_{\text{date}}, \phi_{\text{day\_of\_week}}, 
\phi_{\text{holiday}} \}$, we define the time feature function which combine both $\phi_i$ and $\phi_j$ by $\phi_i \bigotimes \phi_j: [0, t'] \rightarrow C_i \times C_j$ if the codomains of $\phi_i$ and $\phi_j$ are denoted by $C_i$ and $C_j$, respectively, (i.e., $\phi_i: [0, t'] \rightarrow C_i$ and $\phi_j: [0, t'] \rightarrow C_j$). 
This example illustrates that we can consider various candidate time feature functions satisfying Condition~\ref{ass.common_time_feature_support}. Thus, it is not challenging to satisfy Condition~\ref{ass.common_time_feature_support} even after the optimization of the time feature functions using a set $\Phi$, which is reasonably large enough to unbiasedly estimate the time feature effect $g$. Note that the problem of hyperparameter tuning in OPE is a more general topic related not only to the F-OPE problem but also to any OPE problem. Hence the finite sample suboptimality gap between $\phi^*$ and $\hat{\phi}$ is considered an independent area of interest. If we want to keep track of the gap, then we can utilize SLOPE~\citep{su2020adaptive, tucker2021improved} for the hyperparameter tuning as we discussed in the main text where the suboptimality gap is formally derived in their literature.

\section{Theoretical analysis of OPFV-PG}\label{sec:analysis_opfv_f_opl}
\label{appendix.non_stationary_OPL}
In this section, we analyze the statistical properties of the gradient of the OPFV estimator used in F-OPL in Section~\ref{sec:f-opl}. Overall, we have bias and variance similar to the OPFV estimator tailored for F-OPE. Before providing the theoretical properties of the gradient of OPFV, we introduce a new condition for F-OPL.
\begin{assumption}[Full Support]
    \label{ass.full-support}
    The logging policy $\pi_0$ is said to satisfy the full support if $\pi_0(a \,|\, x, t) > 0$ for any $x \in \calX$, $t \in [0, T]$, and $a \in \calA$.
\end{assumption}
Then, under the newly introduced condition, we derive the bias of the gradient of the OPFV estimator as follows.
\begin{theorem}[Bias of the Gradient of OPFV]
    \label{thm.bias_grad_OPFV}
    If Conditions \ref{ass.full-support} and \ref{ass.common_time_feature_support} hold, the gradient of the OPFV estimator has the following bias.
    \begin{align*}
        \bias \left( \opfvgrad \right) 
        = \mE_{p(x, t) \pi_{\zeta}(a|x, t')} \Bigg[ \frac{\ind_{\phi}(t, t')}{p(\phi(t'))} \left( \Delta_q(x, t, t', a) - \Delta_{\hat{f}}(x, t, t', a) \right) s_{\zeta}(x, t', a) \Bigg]
    \end{align*}
    See Appendix \ref{proof_bias_grad_OPFV} for the proof.
\end{theorem}
Theorem \ref{thm.bias_grad_OPFV} shows that the bias of the gradient of the OPFV can be obtained by multiplying the bias of the OPFV for F-OPE in Eq.~\eqref{eq:bias-of-opfv} by the policy score function $s_{\zeta}(x, t', a)$. Thus, as the time feature function becomes fine-grained, the bias decreases, as seen in the arguments for the bias of the OPFV estimator in F-OPE in Section~\ref{sec:analysis}. Furthermore, we observe that the error $\Delta_q(x, t, t', a) - \Delta_{\hat{f}}(x, t, t', a)$ of the relative difference in the expected reward between timestamps $t$ and $t'$ contributes to the bias of the gradient of the OPFV estimator. Hence, in an ideal situation when Condition \ref{ass.conditional_piecewise_correctness} is satisfied, the gradient of the OPFV estimator becomes the unbiased estimator as described in the following proposition.
\begin{proposition}[Unbiasedness of the Gradient of OPFV]
    \label{prop.unbiasedness_grad_OPFV}
    Under Conditions \ref{ass.full-support}, \ref{ass.common_time_feature_support}, and \ref{ass.conditional_piecewise_correctness}, the gradient of OPFV is unbiased. i.e., $\mE_{\calD}[\opfvgrad] = \nabla_{\zeta} V_{t'}(\pi_{\zeta})$. See Appendix \ref{proof_unbiasedness_grad_OPFV} for the proof.
\end{proposition}
In addition to the bias, we derive the variance of the gradient of OPFV in the following proposition.
\begin{proposition}[Variance of the Gradient of OPFV]
    \label{prop.variance_grad_OPFV_true_p_phi}
    Under Conditions \ref{ass.full-support}, \ref{ass.common_time_feature_support}, and \ref{ass.conditional_piecewise_correctness}, the variance of the gradient of the OPFV estimator is as follows.
    \begin{align*}
        n \mV_{\calD} \left[ \opfvgrad \right] 
        &=  \mE_{p(x, t) \pi_0(a|x, t)} \left[ \left( \frac{\ind_{\phi}(t, t')}{p(\phi(t'))} \frac{\pi_{\zeta}(a|x, t')}{\pi_0(a|x, t)} s_{\zeta}(x, t', a) \right)^2 \sigma^2(x, t, a) \right] \\
        &+ \mE_{p(x, t)} \left[ \left( \frac{\ind_{\phi}(t, t')}{p(\phi(t'))} \right)^2 \mV_{\pi_0(a|x, t)} \left[ \frac{\pi_{\zeta}(a|x, t')}{\pi_0(a|x, t)} \Delta_{q, \hat{f}}(x, t', a) s_{\zeta}(x, t', a) \right] \right] \\
        &+ \mV_{p(t)} \left[ \frac{ \ind_{\phi}(t, t')}{p(\phi(t'))} \right] \mE_{p(x)}\left[ \mE_{\pi_{\zeta}(a|x, t')} \left[ \Delta_{q, \hat{f}}(x, t', a) s_{\zeta}(x, t', a) \right]^2 \right] \\
        &+ \mV_{p(x)} \left[ \mE_{\pi_{\zeta}(a|x, t')} \left[q(x, t', a) s_{\zeta}(x, t', a) \right] \right]
    \end{align*}
    where $\Delta_{q, \hat{f}}(x, t, a) := q(x, t, a) - \hat{f}(x, t, a)$ is the estimation error of $\hat{f}(x, t, a)$ against the expected reward funciton $q(x, t, a)$ and $\sigma^2(x, t, a) := \mV_{p(r|x, t, a)}[r]$ is the variance of the reward $r$ given the context $x$, timestamp $t$, and action $a$. See Appendix \ref{proof_variance_grad_OPFV_true_p_phi} for the proof.
\end{proposition}
Proposition \ref{prop.variance_grad_OPFV_true_p_phi} suggests that the term $\ind_{\phi}(t, t') / p(\phi(t'))$ contributes to the variance of the gradient of OPFV. Specifically, using the coarse time feature $\phi$ leads to a low variance, whereas the fine-grained time feature is susceptible to incurring the variance. Hence, when employed in F-OPL, the gradient of the OPFV needs to consider the bias-variance tradeoff by optimizing the granularity of the time feature to achieve low MSE. 

\section{Extension of OPFV-PG with pessimism}
In Appendix~\ref{app:related}, we discussed that the OPL algorithm based on the importance weight might be susceptible to uncertainty when learning a new policy, particularly if a policy is significantly different from the logging policy. This problem also exists in F-OPL as the optimal policy in the future can be considerably different from a logging policy due to the non-stationarity. Thus, in this section, we further extend the OPFV-PG method by incorporating the pessimistic approach to cope with such uncertainty. Specifically, we add a regularization term denoted by $\text{IML}(\pi_{\zeta})$~\cite {ma2019imitation} in the objective of the OPL so that the learned policy does not deviate considerably from the logging policy. Concretely, when we combine OPFV-PG with the pessimistic approach, the objective we want to maximize becomes as follows: 
\begin{align*}
    \hat{V}_{t'}^{\text{OPFV}}(\pi_{\zeta}; \calD) + \rho \cdot \text{IML}(\pi_{\zeta}), 
\end{align*}
where $\rho$ is a parameter which controls the exploration and exploitation tradeoff in OPL and $\text{IML}(\pi_{\zeta}) := - \meanN \log \frac{\pi_{\zeta}(a_i|x_i)}{\pi_0(a_i|x_i, t_i)}$ is the estimator of the Kullback-Leibler (KL) divergence $\text{KL}(\pi_0 || \pi_{\zeta}) := \mE_{p(x, t) \pi_0(a|x, t)}[\log (\frac{\pi_0(a|x, t)}{\pi_{\zeta}(a|x)})]$ between the logging $\pi_0$ and the policy $\pi_{\zeta}$. To incorporate this objective in the policy gradient approach, we use the following gradient of the objective with respect to the parameter $\zeta$:
\begin{align*}
    &\nabla_{\zeta} \left( \hat{V}_{t'}^{\text{OPFV}}(\pi_{\zeta}; \calD) + \rho \cdot \text{IML}(\pi_{\zeta}) \right)\\
    :=& \meanN \Bigg\{ \frac{\ind_{\phi}(t_i, t')}{p(\phi(t'))} \frac{\pi_{\zeta}(a_i\,|\,x_i)}{\pi_0(a_i\,|\,x_i, t_i)} \Big( r_i - \hat{f}(x_i, t_i, a_i) \Big) \nabla_{\zeta} \log \pi_{\zeta}(a_i\,|\,x_i) \notag \\
    & +  \mE_{\pi_{\zeta}(a|x_i)} \left[ \hat{f}(x_i, t', a) \nabla_{\zeta} \log \pi_{\zeta}(a\,|\,x_i) \right] - \rho \cdot \nabla_{\zeta} \log \frac{\pi_{\zeta}(a_i|x_i)}{\pi_0(a_i|x_i, t_i)} \Bigg\}.
\end{align*}
Then, by the following iterative gradient ascent yields the pessimistic policy:  $\zeta_{\tau + 1} \gets \zeta_{\tau} + \eta \cdot \nabla_{\zeta} \left( \hat{V}_{t'}^{\text{OPFV}}(\pi_{\zeta_{\tau}}; \calD) + \rho \cdot \text{IML}(\pi_{\zeta_{\tau}}) \right)$, we can cope with uncertainty even in F-OPL.

\section{Omitted proofs}\label{app:omitted-proof}
In this section, we provide the comprehensive proofs for the theorems and propositions we have introduced on the unbiasedness, bias, and variance of the original OPFV estimator for F-OPE in Section~\ref{sec:analysis}, the gradient of the OPFV estimator employed in OPFV-PG for F-OPL in Section~\ref{sec:analysis_opfv_f_opl}, and the extended OPFV estimator for F-OPE under non-stationary context and reward distributions in Section~\ref{subsec:analysis_opfv_ns_context}.
\subsection{Proof of the unbiasedness of the OPFV estimator}

\label{proof_unbiasedness_OPFV}
\begin{proof}
We prove the unbiasedness of the OPFV estimator in F-OPE under Conditions \ref{ass.common_support}, \ref{ass.common_time_feature_support}, and \ref{ass.conditional_piecewise_correctness} below.
\begin{align}
    &\mE_{\calD}[\opfv] \notag \\
    ={}& \mE_{p(x_i, t_i) \pi_0(a_i|x_i, t_i) p(r_i|x_i, t_i, a_i)  \forall i \in [n]}\Bigg[\meanN \Bigg\{ \frac{\ind_{\phi}(t_i, t')}{p(\phi(t'))} \frac{\pi_e(a_i|x_i, t')}{\pi_0(a_i|x_i, t_i)} \left(r_i - \hat{f}(x_i, t_i, a_i)\right) \notag\\
    &+ \mE_{\pi_e(a|x_i, t')}[\hat{f}(x_i, t', a)] \Bigg\}\Bigg] \notag \\
    ={}& \mE_{p(x, t) \pi_0(a|x, t) p(r|x, t, a)} \left[  \frac{\ind_{\phi}(t, t')}{p(\phi(t'))} \frac{\pi_e(a|x, t')}{\pi_0(a|x, t)} \left(r - \hat{f}(x, t, a)\right) + \mE_{\pi_e(a'|x, t')}[\hat{f}(x, t', a')] \right] \notag \quad\because \text{i.i.d} \\
    ={}& \mE_{p(x, t) \pi_0(a|x, t)} \left[  \frac{\ind_{\phi}(t, t')}{p(\phi(t'))} \frac{\pi_e(a|x, t')}{\pi_0(a|x, t)} \Delta_{q, \hat{f}}(x, t, a) + \mE_{\pi_e(a'|x, t')}[\hat{f}(x, t', a')] \right] \notag  \quad\because \text{def. of $q(x, t, a)$} \\
    ={}& \mE_{p(x, t) \pi_0(a|x, t)} \left[  \frac{\ind_{\phi}(t, t')}{p(\phi(t'))} \frac{\pi_e(a|x, t')}{\pi_0(a|x, t)}\Delta_{q, \hat{f}}(x, t', a) + \mE_{\pi_e(a'|x, t')}[\hat{f}(x, t', a')] \right] \notag \quad\because \text{Condition \ref{ass.conditional_piecewise_correctness}} \\
    ={}& \mE_{p(x, t) \pi_e(a|x, t')} \left[  \frac{\ind_{\phi}(t, t')}{p(\phi(t'))} \Delta_{q, \hat{f}}(x, t', a) \right] + \mE_{p(x, t) \pi_e(a'|x, t')}[\hat{f}(x, t', a')] \notag \quad\because \text{cancel out $\pi_0(a|x, t)$} \\
    ={}& \mE_{p(x) p(t) \pi_e(a|x, t')} \left[  \frac{\ind_{\phi}(t, t')}{p(\phi(t'))} \Delta_{q, \hat{f}}(x, t', a) \right] + \mE_{p(x) \pi_e(a'|x, t')}[\hat{f}(x, t', a')] \notag \quad\because \text{stationary context} \\
    ={}& \mE_{p(x) \pi_e(a|x, t')} \left[ \int_{t \in [0, T]} p(t) \frac{\ind_{\phi}(t, t')}{p(\phi(t'))} \Delta_{q, \hat{f}}(x, t', a) dt \right] + \mE_{p(x) \pi_e(a'|x, t')}[\hat{f}(x, t', a')] \notag \\
    ={}& \mE_{p(x) \pi_e(a|x, t')} \left[ \frac{\Delta_{q, \hat{f}}(x, t', a)}{p(\phi(t'))}  \int_{t \in [0, T]} p(t) \ind_{\phi}(t, t') dt \right] + \mE_{p(x) \pi_e(a'|x, t')}[\hat{f}(x, t', a')] \notag \\
    ={}& \mE_{p(x) \pi_e(a|x, t')} \left[ \frac{\Delta_{q, \hat{f}}(x, t', a)}{p(\phi(t'))} p(\phi(t')) \right] + \mE_{p(x) \pi_e(a'|x, t')}[\hat{f}(x, t', a')] \notag \quad \because \text{definition of $p(\phi(t'))$} \\
    ={}& \mE_{p(x) \pi_e(a|x, t')} \left[\Delta_{q, \hat{f}}(x, t', a) \right] + \mE_{p(x) \pi_e(a'|x, t')}[\hat{f}(x, t', a')] \notag \quad \because \text{cancel out $p(\phi(t'))$} \\
    ={}& \mE_{p(x) \pi_e(a|x, t')} \left[\Delta_{q, \hat{f}}(x, t', a) + \hat{f}(x, t', a) \right] \notag \\
    ={}& \mE_{p(x) \pi_e(a|x, t')} \left[q(x, t', a) \right] \notag \quad \because \text{Cancel out $\hat{f}(x, t', a)$} \\
    ={}& \mE_{p(x|t') \pi_e(a|x, t')} \left[q(x, t', a) \right] \notag \quad \because \text{stationary context i.e., $p(x) = p(x|t')$} \\
    ={}& V_{t'}(\pi_e) \notag \quad \because \text{definition of the value function} 
\end{align}
\end{proof}

\subsection{Proof of Theorem \ref{thm:bias-opfv}}
\label{proof_bias_OPFV}
\begin{proof}
If the Condition \ref{ass.conditional_piecewise_correctness} is violated, the bias of the OPFV estimator in F-OPE under Conditions \ref{ass.common_support} and \ref{ass.common_time_feature_support} is derived as follows.
\begin{align*}
    & \bias \left( \opfv \right) \\
    ={}& \mE_{\calD} \left[ \opfv \right] - V_{t'}(\pi_e) \quad \because \text{definition of the bias} \\
    ={}& \mE_{p(x, t) \pi_0(a|x, t) p(r|x, t, a)} \left[  \frac{\ind_{\phi}(t, t')}{p(\phi(t'))} \frac{\pi_e(a|x, t')}{\pi_0(a|x, t)} \left(r - \hat{f}(x, t, a)\right) + \mE_{\pi_e(a'|x, t')}[\hat{f}(x, t', a')] \right] \notag \\
    & - \mE_{p(x|t') \pi_e(a|x, t')}[q(x, t', a)] \\
    ={}& \mE_{p(x, t) \pi_0(a|x, t)} \left[  \frac{\ind_{\phi}(t, t')}{p(\phi(t'))} \frac{\pi_e(a|x, t')}{\pi_0(a|x, t)} \Delta_{q, \hat{f}}(x, t, a) + \mE_{\pi_e(a'|x, t')}[\hat{f}(x, t', a')] \right] \notag \\
    & - \mE_{p(x|t') \pi_e(a|x, t')}[q(x, t', a)] \\
    ={}& \mE_{p(x) p(t) \pi_0(a|x, t)} \left[  \frac{\ind_{\phi}(t, t')}{p(\phi(t'))} \frac{\pi_e(a|x, t')}{\pi_0(a|x, t)} \Delta_{q, \hat{f}}(x, t, a) + \mE_{\pi_e(a'|x, t')}[\hat{f}(x, t', a')] \right] \notag \\
    & - \mE_{p(x) \pi_e(a|x, t')}[q(x, t', a)] \quad \because \text{stationary context} \\
    ={}& \mE_{p(x) p(t) \pi_0(a|x, t)} \left[  \frac{\ind_{\phi}(t, t')}{p(\phi(t'))} \frac{\pi_e(a|x, t')}{\pi_0(a|x, t)} \Delta_{q, \hat{f}}(x, t, a) \right] - \mE_{p(x) \pi_e(a|x, t')}[\Delta_{q, \hat{f}}(x, t', a)] \\
    ={}& \mE_{p(x) p(t) \pi_e(a|x, t')} \left[  \frac{\ind_{\phi}(t, t')}{p(\phi(t'))} \Delta_{q, \hat{f}}(x, t, a) \right] - \mE_{p(x) \pi_e(a|x, t')}[\Delta_{q, \hat{f}}(x, t', a)] \quad \because \text{cancel out $\pi_0(a|x, t')$} \\
    ={}& \mE_{p(x) \pi_e(a|x, t')} \left[  \frac{1}{p(\phi(t'))} \mE_{p(t)} [\ind_{\phi}(t, t') \Delta_{q, \hat{f}}(x, t, a)] - \Delta_{q, \hat{f}}(x, t', a) \right]  \\
    ={}& \mE_{p(x) \pi_e(a|x, t')} \left[  \frac{1}{p(\phi(t'))} \left( \mE_{p(t)} [\ind_{\phi}(t, t') \Delta_{q, \hat{f}}(x, t, a)] - p(\phi(t')) \Delta_{q, \hat{f}}(x, t', a) \right) \right] \\
    ={}& \mE_{p(x) \pi_e(a|x, t')} \left[  \frac{1}{p(\phi(t'))} \left( \mE_{p(t)} [\ind_{\phi}(t, t') \Delta_{q, \hat{f}}(x, t, a)] - \mE_{p(t)}[\ind_{\phi}(t, t')] \Delta_{q, \hat{f}}(x, t', a) \right) \right] \quad \because \text{def. of $p(\phi(t))$} \\
    ={}& \mE_{p(x, t) \pi_e(a|x, t')} \left[  \frac{\ind_{\phi}(t, t')}{p(\phi(t'))} \left( \Delta_{q, \hat{f}}(x, t, a) - \Delta_{q, \hat{f}}(x, t', a) \right) \right]\\
    ={}& \mE_{p(x, t) \pi_e(a|x, t')} \left[  \frac{\ind_{\phi}(t, t')}{p(\phi(t'))} \left( \Delta_{q}(x, t, t', a) - \Delta_{\hat{f}}(x, t, t', a)  \right) \right]
\end{align*}
Note that this quantity is bounded by $2 r_{\text{max}}$ since $|\Delta_{q}(x, t, t', a) - \Delta_{\hat{f}}(x, t, t', a)| \le 2 r_{\text{max}}$.
\end{proof}

\subsection{Proof of Proposition \ref{prop.variance_OPFV_true_p_phi}}
\label{proof_variance_OPFV_true_p_phi}
\begin{proof}
We can derive the variance of the OPFV estimator for F-OPE under Conditions \ref{ass.common_support}, \ref{ass.common_time_feature_support}, and \ref{ass.conditional_piecewise_correctness} below.
\begin{align*}
    % & n \mV_{\calD} \left[ \opfv \right] \\
    & n \var \left[ \opfv \right] \\
    ={}& n \mV_{p(x_i, t_i) \pi_0(a_i|x_i, t_i) p(r_i|x_i, t_i, a_i)  \forall i \in [n]}\Bigg[\meanN \Bigg\{ \frac{\ind_{\phi}(t_i, t')}{p(\phi(t'))} \frac{\pi_e(a_i|x_i, t')}{\pi_0(a_i|x_i, t_i)} \left(r_i - \hat{f}(x_i, t_i, a_i)\right) \notag \\
    & + \mE_{\pi_e(a|x_i, t')}[\hat{f}(x_i, t', a)] \Bigg\}\Bigg]  \\
    ={}& \mV_{p(x, t) \pi_0(a|x, t) p(r|x, t, a)} \left[  \frac{\ind_{\phi}(t, t')}{p(\phi(t'))} \frac{\pi_e(a|x, t')}{\pi_0(a|x, t)} \left(r - \hat{f}(x, t, a)\right) + \mE_{\pi_e(a'|x, t')}[\hat{f}(x, t', a')] \right] \quad \because \text{i.i.d} \\
    ={}& \mE_{p(x, t) \pi_0(a|x, t)} \left[ \mV_{p(r|x, t, a)} \left[ \frac{\ind_{\phi}(t, t')}{p(\phi(t'))} \frac{\pi_e(a|x, t')}{\pi_0(a|x, t)} \left(r - \hat{f}(x, t, a)\right) + \mE_{\pi_e(a'|x, t')}[\hat{f}(x, t', a')] \right] \right] \\
    &+ \mV_{p(x, t) \pi_0(a|x, t)} \left[ \mE_{p(r|x, t, a)} \left[ \frac{\ind_{\phi}(t, t')}{p(\phi(t'))} \frac{\pi_e(a|x, t')}{\pi_0(a|x, t)} \left(r - \hat{f}(x, t, a)\right) + \mE_{\pi_e(a'|x, t')}[\hat{f}(x, t', a')] \right] \right] \quad \because \text{total var.} \\
    ={}& \mE_{p(x, t) \pi_0(a|x, t)} \left[ \left( \frac{\ind_{\phi}(t, t')}{p(\phi(t'))} \frac{\pi_e(a|x, t')}{\pi_0(a|x, t)} \right)^2 \sigma^2(x, t, a) \right] \\
    &+ \mV_{p(x, t) \pi_0(a|x, t)} \left[\frac{\ind_{\phi}(t, t')}{p(\phi(t'))} \frac{\pi_e(a|x, t')}{\pi_0(a|x, t)} \Delta_{q, \hat{f}}(x, t, a) + \mE_{\pi_e(a'|x, t')}[\hat{f}(x, t', a')]\right] \quad \because \text{$\sigma^2(x, t, a) := \mV_{p(r|x, t, a)}[r]$} \\
    ={}& \mE_{p(x, t) \pi_0(a|x, t)} \left[ \left( \frac{\ind_{\phi}(t, t')}{p(\phi(t'))} \frac{\pi_e(a|x, t')}{\pi_0(a|x, t)} \right)^2 \sigma^2(x, t, a) \right] \\
    &+ \mE_{p(x, t)} \left[ \mV_{\pi_0(a|x, t)} \left[ \frac{\ind_{\phi}(t, t')}{p(\phi(t'))} \frac{\pi_e(a|x, t')}{\pi_0(a|x, t)} \Delta_{q, \hat{f}}(x, t, a) + \mE_{\pi_e(a'|x, t')}[\hat{f}(x, t', a')] \right] \right] \\
    &+ \mV_{p(x, t)} \left[ \mE_{\pi_0(a|x, t)} \left[ \frac{\ind_{\phi}(t, t')}{p(\phi(t'))} \frac{\pi_e(a|x, t')}{\pi_0(a|x, t)} \Delta_{q, \hat{f}}(x, t, a) + \mE_{\pi_e(a'|x, t')}[\hat{f}(x, t', a')] \right] \right] \quad \because \text{law of total variance} \\
    ={}& \mE_{p(x, t) \pi_0(a|x, t)} \left[ \left( \frac{\ind_{\phi}(t, t')}{p(\phi(t'))} \frac{\pi_e(a|x, t')}{\pi_0(a|x, t)} \right)^2 \sigma^2(x, t, a) \right] \\
    &+ \mE_{p(x, t)} \left[ \left( \frac{\ind_{\phi}(t, t')}{p(\phi(t'))} \right)^2 \mV_{\pi_0(a|x, t)} \left[ \frac{\pi_e(a|x, t')}{\pi_0(a|x, t)} \Delta_{q, \hat{f}}(x, t, a) \right] \right] \\
    &+ \mV_{p(x, t)} \left[ \mE_{\pi_e(a|x, t')} \left[ \frac{\ind_{\phi}(t, t')}{p(\phi(t'))}  \Delta_{q, \hat{f}}(x, t, a) \right] + \mE_{\pi_e(a'|x, t')}[\hat{f}(x, t', a')] \right] \quad \because \text{cancel out $\pi_0(a|x, t)$}\\
    ={}& \mE_{p(x, t) \pi_0(a|x, t)} \left[ \left( \frac{\ind_{\phi}(t, t')}{p(\phi(t'))} \frac{\pi_e(a|x, t')}{\pi_0(a|x, t)} \right)^2 \sigma^2(x, t, a) \right] \\
    &+ \mE_{p(x, t)} \left[ \left( \frac{\ind_{\phi}(t, t')}{p(\phi(t'))} \right)^2 \mV_{\pi_0(a|x, t)} \left[ \frac{\pi_e(a|x, t')}{\pi_0(a|x, t)} \Delta_{q, \hat{f}}(x, t, a) \right] \right] \\
    &+ \mV_{p(x) p(t)} \left[ \mE_{\pi_e(a|x, t')} \left[ \frac{\ind_{\phi}(t, t')}{p(\phi(t'))}  \Delta_{q, \hat{f}}(x, t, a) \right] + \mE_{\pi_e(a'|x, t')}[\hat{f}(x, t', a')] \right] \quad \because \text{stationary context} 
\end{align*}
\begin{align*}
    % & n \mV_{\calD} \left[ \opfv \right] \\
    ={}& \mE_{p(x, t) \pi_0(a|x, t)} \left[ \left( \frac{\ind_{\phi}(t, t')}{p(\phi(t'))} \frac{\pi_e(a|x, t')}{\pi_0(a|x, t)} \right)^2 \sigma^2(x, t, a) \right] \\
    &+ \mE_{p(x, t)} \left[ \left( \frac{\ind_{\phi}(t, t')}{p(\phi(t'))} \right)^2 \mV_{\pi_0(a|x, t)} \left[ \frac{\pi_e(a|x, t')}{\pi_0(a|x, t)} \Delta_{q, \hat{f}}(x, t, a) \right] \right] \\
    &+ \mE_{p(x)} \left[ \mV_{p(t)} \left[ \mE_{\pi_e(a|x, t')} \left[ \frac{\ind_{\phi}(t, t')}{p(\phi(t'))}  \Delta_{q, \hat{f}}(x, t, a) \right] + \mE_{\pi_e(a'|x, t')}[\hat{f}(x, t', a')] \right] \right] \\
    &+ \mV_{p(x)} \left[ \mE_{p(t)} \left[ \mE_{\pi_e(a|x, t')} \left[ \frac{\ind_{\phi}(t, t')}{p(\phi(t'))}  \Delta_{q, \hat{f}}(x, t, a) \right] + \mE_{\pi_e(a'|x, t')}[\hat{f}(x, t', a')] \right] \right] \quad \because \text{law of total variance} \\
    ={}& \mE_{p(x, t) \pi_0(a|x, t)} \left[ \left( \frac{\ind_{\phi}(t, t')}{p(\phi(t'))} \frac{\pi_e(a|x, t')}{\pi_0(a|x, t)} \right)^2 \sigma^2(x, t, a) \right] \\
    &+ \mE_{p(x, t)} \left[ \left( \frac{\ind_{\phi}(t, t')}{p(\phi(t'))} \right)^2 \mV_{\pi_0(a|x, t)} \left[ \frac{\pi_e(a|x, t')}{\pi_0(a|x, t)} \Delta_{q, \hat{f}}(x, t', a) \right] \right] \\
    &+ \mE_{p(x)} \left[ \mV_{p(t)} \left[ \mE_{\pi_e(a|x, t')} \left[ \frac{\ind_{\phi}(t, t')}{p(\phi(t'))}  \Delta_{q, \hat{f}}(x, t', a) \right] \right] \right] \\
    &+ \mV_{p(x)} \left[ \mE_{p(t) \pi_e(a|x, t')} \left[ \frac{\ind_{\phi}(t, t')}{p(\phi(t'))}  \Delta_{q, \hat{f}}(x, t', a) \right] + \mE_{\pi_e(a'|x, t')}[\hat{f}(x, t', a')] \right] \quad \because \text{Condition \ref{ass.conditional_piecewise_correctness}} \\
    ={}& \mE_{p(x, t) \pi_0(a|x, t)} \left[ \left( \frac{\ind_{\phi}(t, t')}{p(\phi(t'))} \frac{\pi_e(a|x, t')}{\pi_0(a|x, t)} \right)^2 \sigma^2(x, t, a) \right] \\
    &+ \mE_{p(x, t)} \left[ \left( \frac{\ind_{\phi}(t, t')}{p(\phi(t'))} \right)^2 \mV_{\pi_0(a|x, t)} \left[ \frac{\pi_e(a|x, t')}{\pi_0(a|x, t)} \Delta_{q, \hat{f}}(x, t', a) \right] \right] \\
    &+ \mE_{p(x)} \left[ \mE_{\pi_e(a|x, t)} \left[ \Delta_{q, \hat{f}}(x, t', a) \right]^2 \right] \cdot \mV_{p(t)} \left[ \frac{\ind_{\phi}(t, t')}{p(\phi(t'))} \right] \\
    &+ \mV_{p(x)} \left[ \mE_{\pi_e(a|x, t')} \left[ \int_{t \in [0, T]} p(t) \frac{\ind_{\phi}(t, t')}{p(\phi(t'))}  \Delta_{q, \hat{f}}(x, t', a) dt \right] + \mE_{\pi_e(a'|x, t')}[\hat{f}(x, t', a')] \right] \\
    ={}& \mE_{p(x, t) \pi_0(a|x, t)} \left[ \left( \frac{\ind_{\phi}(t, t')}{p(\phi(t'))} \frac{\pi_e(a|x, t')}{\pi_0(a|x, t)} \right)^2 \sigma^2(x, t, a) \right] \\
    &+ \mE_{p(x, t)} \left[ \left( \frac{\ind_{\phi}(t, t')}{p(\phi(t'))} \right)^2 \mV_{\pi_0(a|x, t)} \left[ \frac{\pi_e(a|x, t')}{\pi_0(a|x, t)} \Delta_{q, \hat{f}}(x, t', a) \right] \right] \\
    &+ \mE_{p(x)} \left[ \mE_{\pi_e(a|x, t')} \left[ \Delta_{q, \hat{f}}(x, t', a) \right]^2 \right] \cdot \mV_{p(t)} \left[ \frac{\ind_{\phi}(t, t')}{p(\phi(t'))} \right] \\
    &+ \mV_{p(x)} \left[ \mE_{\pi_e(a|x, t')} \left[ \Delta_{q, \hat{f}}(x, t', a) \right] + \mE_{\pi_e(a'|x, t')}[\hat{f}(x, t', a')] \right] \quad \because \text{cancel out $p(\phi(t))$} \\
    ={}& \mE_{p(x, t) \pi_0(a|x, t)} \left[ \left( \frac{\ind_{\phi}(t, t')}{p(\phi(t'))} \frac{\pi_e(a|x, t')}{\pi_0(a|x, t)} \right)^2 \sigma^2(x, t, a) \right] \\
    &+ \mE_{p(x, t)} \left[ \left( \frac{\ind_{\phi}(t, t')}{p(\phi(t'))} \right)^2 \mV_{\pi_0(a|x, t)} \left[ \frac{\pi_e(a|x, t')}{\pi_0(a|x, t)} \Delta_{q, \hat{f}}(x, t', a) \right] \right] \\
    &+ \mE_{p(x)} \left[ \mE_{\pi_e(a|x, t)} \left[ \Delta_{q, \hat{f}}(x, t', a) \right]^2 \right] \cdot \mV_{p(t)} \left[ \frac{\ind_{\phi}(t, t')}{p(\phi(t'))} \right] \\
    &+ \mV_{p(x)} \left[ \mE_{\pi_e(a|x, t')} \left[ q(x, t', a) \right] \right] \quad \because \text{cancel out $\hat{f}(x, t', a)$} \\
\end{align*}
Note that this quantity is well-bounded under the bounded importance weight $(i.e., \pi_e(a|x, t') / \pi_0(a|x, t) < \infty)$, the bounded conditional variance of the reward $(i.e., \sigma^2(x, t, a) < \infty)$, and the bounded variance of the expected reward in terms of the context $(i.e., \mV_{p(x)}[\mE_{\pi_e(a|x, t')}[q(x, t', a)]] < \infty)$ for all $x \in \calX, t \in [0, T], a \in \calA$, and $t' > T$.
% In addition, the difference between the variances of the OPFV estimators with coarse and fine time features in Proposition~\ref{prop.var-OPFV-w-different-phi} can be derived immediately from this proof.
\end{proof}

\subsection{Proof of Theorem \ref{thm.bias-reduction-OPFV-against-DR}}
\label{proof-bias-reduction-opfv-dr}
\begin{proof}
the bias reduction of the OPFV estimator against DR in F-OPE under Conditions \ref{ass.common_support} and \ref{ass.common_time_feature_support} is derived as follows. Note that we use the following DR estimator applied to the estimation of a policy at target time $t'$ to distinguish which part of the OPFV estimator contributes to the bias reduction against the naive application of DR:
\begin{align*}
    \drtarget := \meanN \left\{ \frac{\ind_{\phi}(t_i, t')}{p(\phi(t'))} \frac{\pi_e(a_i|x_i, t')}{\pi_0(a_i|x_i, t_i)} (r_i - \hat{f}(x_i, t_i, a_i)) + \mE_{\pi_e(a|x_i, t')} [\hat{f}(x_i, t', a)] \right\}
\end{align*}
Then, adding and subtracting the bias of the newly introduced DR estimator, we decompose the bias reduction of OPFV compared to DR into three terms in the following way.
\begin{align}
    & \bias \left( \dr \right) - \bias \left( \opfv \right) \notag \\
    ={}& \bias \left( \dr \right) - \bias \left( \drtarget \right) 
    + \bias \left( \drtarget \right) - \bias \left( \opfv \right) \notag \\
    ={}& \mE_{\calD} \left[ \dr \right] - \mE_{\calD} \left[ \drtarget \right] \notag \\
    +& \mE_{p(x, t) \pi_e(a|x, t')} \left[ \Delta_{q}(x, t, t', a) - \Delta_{\hat{q}}(x, t, t', a) \right] - \mE_{p(x, t) \pi_e(a|x, t')} \left[ \frac{\ind_{\phi}(t, t')}{p(\phi(t'))} \left( \Delta_{q}(x, t, t', a) - \Delta_{\hat{q}}(x, t, t', a) \right) \right] \notag \\
    ={}& \mE_{p(x, t) \pi_e(a|x, t)} \left[ q(x, t, a) \right] - \mE_{p(x, t) \pi_e(a|x, t')} \left[ \Delta_{q, \hat{q}}(x, t, a) + \hat{q}(x, t', a) \right] \notag \\
    +& \mE_{p(x, t) \pi_e(a|x, t')} \left[ \left( \Delta_{q}(x, t, t', a) - \Delta_{\hat{q}}(x, t, t', a) \right) - \frac{\ind_{\phi}(t, t')}{p(\phi(t'))} \left( \Delta_{q}(x, t, t', a) - \Delta_{\hat{q}}(x, t, t', a) \right)\right] \notag \\
    ={}& \mE_{p(x, t)} \left[ \sumA \bigg\{ \pi_e(a|x, t) q(x, t, a) - \pi_e(a|x, t') \left( \Delta_{q, \hat{q}}(x, t, a) + \hat{q}(x, t', a) \right) \bigg\} \right] \notag \\
    +& \mE_{p(x, t) \pi_e(a|x, t')} \left[ \left( \Delta_{q}(x, t, t', a) - \Delta_{\hat{q}}(x, t, t', a) \right) - \frac{\ind_{\phi}(t, t')}{p(\phi(t'))} \left( \Delta_{q}(x, t, t', a) - \Delta_{\hat{q}}(x, t, t', a) \right)\right] \notag \\
    ={}& \mE_{p(x, t)} \Bigg[ \sumA \bigg\{ \pi_e(a|x, t) q(x, t, a) - \pi_e(a|x, t') q(x, t, a) \notag \\
    +& \pi_e(a|x, t') q(x, t, a) - \pi_e(a|x, t') \left( \Delta_{q, \hat{q}}(x, t, a) + \hat{q}(x, t', a) \right) \bigg\} \Bigg] \notag \\
    +& \mE_{p(x, t) \pi_e(a|x, t')} \left[ \left( \Delta_{q}(x, t, t', a) - \Delta_{\hat{q}}(x, t, t', a) \right) - \frac{\ind_{\phi}(t, t')}{p(\phi(t'))} \left( \Delta_{q}(x, t, t', a) - \Delta_{\hat{q}}(x, t, t', a) \right)\right] \notag \\
    ={}& \mE_{p(x, t)} \Bigg[ \sumA \bigg\{ (\pi_e(a|x, t) - \pi_e(a|x, t')) q(x, t, a) + \pi_e(a|x, t') \Delta_{\hat{q}}(x, t, t', a) \bigg\} \Bigg] \notag \\
    +& \mE_{p(x, t) \pi_e(a|x, t')} \left[ \left( \Delta_{q}(x, t, t', a) - \Delta_{\hat{q}}(x, t, t', a) \right) - \frac{\ind_{\phi}(t, t')}{p(\phi(t'))} \left( \Delta_{q}(x, t, t', a) - \Delta_{\hat{q}}(x, t, t', a) \right)\right] \notag
\end{align}
\end{proof}

\subsection{Proof of Proposition \ref{prop.unbiasedness_grad_OPFV}}
\label{proof_unbiasedness_grad_OPFV}
\begin{proof}
We prove the unbiasedness of the gradient of the OPFV estimator for F-OPL under Conditions \ref{ass.full-support}, \ref{ass.common_time_feature_support}, and \ref{ass.conditional_piecewise_correctness} below.
\begin{align}
    &\mE_{\calD}\left[ \opfvgrad \right] \notag \\
    ={}& \mE_{\calD}\Bigg[\meanN \Bigg\{ \frac{\ind_{\phi}(t_i, t')}{p(\phi(t'))} \frac{\pi_{\zeta}(a_i|x_i, t')}{\pi_0(a_i|x_i, t_i)} \left(r_i - \hat{f}(x_i, t_i, a_i)\right) s_{\zeta}(x_i, t', a_i) \notag \\
    & + \mE_{\pi_{\zeta}(a|x_i, t')}[\hat{f}(x_i, t', a) s_{\zeta}(x_i, t', a)] \Bigg\}\Bigg] \notag \\
    ={}& \mE_{p(x, t) \pi_0(a|x, t) p(r|x, t, a)} \Bigg[  \frac{\ind_{\phi}(t, t')}{p(\phi(t'))} \frac{\pi_{\zeta}(a|x, t')}{\pi_0(a|x, t)} \left(r - \hat{f}(x, t, a)\right) s_{\zeta}(x, t', a) \notag \\
    & + \mE_{\pi_{\zeta}(a'|x, t')}[\hat{f}(x, t', a') s_{\zeta}(x, t', a')] \Bigg] \notag \quad\because \text{i.i.d} \\
    ={}& \mE_{p(x, t) \pi_0(a|x, t)} \left[  \frac{\ind_{\phi}(t, t')}{p(\phi(t'))} \frac{\pi_{\zeta}(a|x, t')}{\pi_0(a|x, t)} \Delta_{q, \hat{f}}(x, t, a) s_{\zeta}(x, t', a) + \mE_{\pi_{\zeta}(a'|x, t')}[\hat{f}(x, t', a') s_{\zeta}(x, t', a')] \right] \notag  \\
    ={}& \mE_{p(x, t) \pi_0(a|x, t)} \left[  \frac{\ind_{\phi}(t, t')}{p(\phi(t'))} \frac{\pi_{\zeta}(a|x, t')}{\pi_0(a|x, t)} \Delta_{q, \hat{f}}(x, t', a)) s_{\zeta}(x, t', a) + \mE_{\pi_{\zeta}(a'|x, t')}[\hat{f}(x, t', a') s_{\zeta}(x, t', a')] \right] \notag \quad\because \text{Cond. \ref{ass.conditional_piecewise_correctness}} \\
    ={}& \mE_{p(x, t) \pi_{\zeta}(a|x, t')} \left[  \frac{\ind_{\phi}(t, t')}{p(\phi(t'))} \Delta_{q, \hat{f}}(x, t', a) s_{\zeta}(x, t', a) \right] + \mE_{p(x, t) \pi_{\zeta}(a'|x, t')}[\hat{f}(x, t', a') s_{\zeta}(x, t', a')] \notag \quad\because \text{cancel out $\pi_0$} \\
    ={}& \mE_{p(x) p(t) \pi_{\zeta}(a|x, t')} \left[  \frac{\ind_{\phi}(t, t')}{p(\phi(t'))} \Delta_{q, \hat{f}}(x, t', a) s_{\zeta}(x, t', a)\right] + \mE_{p(x) \pi_{\zeta}(a'|x, t')}[\hat{f}(x, t', a') s_{\zeta}(x, t', a')] \notag \quad\because \text{stationary $x$} \\
    ={}& \mE_{p(x) \pi_{\zeta}(a|x, t')} \left[ \int_{t \in [0, T]} p(t) \frac{\ind_{\phi}(t, t')}{p(\phi(t'))} \Delta_{q, \hat{f}}(x, t', a) s_{\zeta}(x, t', a) dt \right] + \mE_{p(x) \pi_{\zeta}(a'|x, t')}[\hat{f}(x, t', a') s_{\zeta}(x, t', a')] \notag \\
    ={}& \mE_{p(x) \pi_{\zeta}(a|x, t')} \left[ \frac{\Delta_{q, \hat{f}}(x, t', a) s_{\zeta}(x, t', a)}{p(\phi(t'))}  \int_{t \in [0, T]} p(t) \ind_{\phi}(t, t') dt \right] + \mE_{p(x) \pi_{\zeta}(a'|x, t')}[\hat{f}(x, t', a') s_{\zeta}(x, t', a')] \notag \\
    ={}& \mE_{p(x) \pi_{\zeta}(a|x, t')} \left[ \frac{\Delta_{q, \hat{f}}(x, t', a) s_{\zeta}(x, t', a)}{p(\phi(t'))} p(\phi(t')) \right] + \mE_{p(x) \pi_{\zeta}(a'|x, t')}[\hat{f}(x, t, a' s_{\zeta}(x, t', a'))] \notag \quad \because \text{def. of $p(\phi(t'))$} \\
    ={}& \mE_{p(x) \pi_{\zeta}(a|x, t')} \left[\Delta_{q, \hat{f}}(x, t', a) s_{\zeta}(x, t', a) \right] + \mE_{p(x) \pi_{\zeta}(a'|x, t')}[\hat{f}(x, t', a') s_{\zeta}(x, t', a')] \notag \quad \because \text{cancel out $p(\phi(t'))$} \\
    ={}& \mE_{p(x) \pi_{\zeta}(a|x, t')} \left[ \left(\Delta_{q, \hat{f}}(x, t', a) + \hat{f}(x, t', a) \right) s_{\zeta}(x, t', a) \right] \notag \\
    ={}& \mE_{p(x) \pi_{\zeta}(a|x, t')} \left[q(x, t', a) s_{\zeta}(x, t', a) \right] \notag \quad \because \text{Condition \ref{ass.conditional_piecewise_correctness}} \\
    ={}& \mE_{p(x|t') \pi_{\zeta}(a|x, t')} \left[q(x, t', a) s_{\zeta}(x, t', a) \right] \notag \quad \because \text{stationary context i.e., $p(x) = p(x|t')$} \\
    ={}& \nabla_{\zeta} V_{t'}(\pi_{\zeta}) \notag \quad \because \text{definition of the value function} 
\end{align}
\end{proof}

\subsection{Proof of Theorem \ref{thm.bias_grad_OPFV}}
\label{proof_bias_grad_OPFV}
\begin{proof}
If the Condition \ref{ass.conditional_piecewise_correctness} is violated, the bias of the gradient of the OPFV estimator under Conditions  \ref{ass.full-support} and \ref{ass.common_time_feature_support} is derived as follows.
\begin{align*}
    & \bias \left( \opfvgrad \right) \\
    ={}& \mE_{\calD} \left[ \opfvgrad \right] - \nabla_{\zeta} V_{t'}(\pi_{\zeta}) \quad \because \text{definition of the bias} \\
    ={}& \mE_{p(x, t) \pi_0(a|x, t) p(r|x, t, a)} \left[  \frac{\ind_{\phi}(t, t')}{p(\phi(t'))} \frac{\pi_{\zeta}(a|x, t')}{\pi_0(a|x, t)} \left(r - \hat{f}(x, t, a)\right) s_{\zeta}(x, t', a) + \mE_{\pi_{\zeta}(a'|x, t')}[\hat{f}(x, t', a') s_{\zeta}(x, t', a')] \right] \\
    & - \mE_{p(x|t') \pi_{\zeta}(a|x, t')}[q(x, t', a) s_{\zeta}(x, t', a)] \quad \because \text{definistion of the gradient of the OPFV and i.i.d.}  \\
    ={}& \mE_{p(x, t) \pi_0(a|x, t)} \left[  \frac{\ind_{\phi}(t, t')}{p(\phi(t'))} \frac{\pi_{\zeta}(a|x, t')}{\pi_0(a|x, t)} \Delta_{q, \hat{f}}(x, t, a) s_{\zeta}(x, t', a) + \mE_{\pi_{\zeta}(a'|x, t')}[\hat{f}(x, t', a') s_{\zeta}(x, t', a')] \right] \\
    & - \mE_{p(x|t') \pi_{\zeta}(a|x, t')}[q(x, t', a) s_{\zeta}(x, t', a)] \\
    ={}& \mE_{p(x) p(t) \pi_0(a|x, t)} \left[  \frac{\ind_{\phi}(t, t')}{p(\phi(t'))} \frac{\pi_{\zeta}(a|x, t')}{\pi_0(a|x, t)} \Delta_{q, \hat{f}}(x, t, a) s_{\zeta}(x, t', a) + \mE_{\pi_{\zeta}(a'|x, t')}[\hat{f}(x, t', a') s_{\zeta}(x, t', a')] \right] \\
    & - \mE_{p(x) \pi_{\zeta}(a|x, t')}[q(x, t', a) s_{\zeta}(x, t', a)] \quad \because \text{stationary context i.e., $p(x|t) = p(x)$} \\
    ={}& \mE_{p(x) p(t) \pi_0(a|x, t)} \left[  \frac{\ind_{\phi}(t, t')}{p(\phi(t'))} \frac{\pi_{\zeta}(a|x, t')}{\pi_0(a|x, t)} \Delta_{q, \hat{f}}(x, t, a) s_{\zeta}(x, t', a) \right] - \mE_{p(x) \pi_{\zeta}(a|x, t')}[\Delta_{q, \hat{f}}(x, t', a) s_{\zeta}(x, t', a)] \\
    ={}& \mE_{p(x) p(t) \pi_{\zeta}(a|x, t')} \left[  \frac{\ind_{\phi}(t, t')}{p(\phi(t'))} \Delta_{q, \hat{f}}(x, t, a) s_{\zeta}(x, t', a) \right] - \mE_{p(x) \pi_{\zeta}(a|x, t')}[\Delta_{q, \hat{f}}(x, t', a) s_{\zeta}(x, t', a)] \quad \because \text{cancel out $\pi_0$} \\
    ={}& \mE_{p(x) \pi_{\zeta}(a|x, t')} \left[  \frac{1}{p(\phi(t'))} \mE_{p(t)} [\ind_{\phi}(t, t') \Delta_{q, \hat{f}}(x, t, a) s_{\zeta}(x, t', a)] - \Delta_{q, \hat{f}}(x, t', a) s_{\zeta}(x, t', a) \right]  \\
    ={}& \mE_{p(x) \pi_{\zeta}(a|x, t')} \left[  \frac{1}{p(\phi(t'))} \left( \mE_{p(t)} [\ind_{\phi}(t, t') \Delta_{q, \hat{f}}(x, t, a) s_{\zeta}(x, t', a)] - p(\phi(t')) \Delta_{q, \hat{f}}(x, t', a) s_{\zeta}(x, t', a) \right) \right] \\
    ={}& \mE_{p(x) \pi_{\zeta}(a|x, t')} \left[  \frac{1}{p(\phi(t'))} \left( \mE_{p(t)} [\ind_{\phi}(t, t') \Delta_{q, \hat{f}}(x, t, a) s_{\zeta}(x, t', a)] - \mE_{p(t)}[\ind_{\phi}(t, t')] \Delta_{q, \hat{f}}(x, t', a) s_{\zeta}(x, t', a) \right) \right] \\
    ={}& \mE_{p(x, t) \pi_{\zeta}(a|x, t')} \left[  \frac{\ind_{\phi}(t, t')}{p(\phi(t'))} \left( \Delta_{q, \hat{f}}(x, t, a)] - \Delta_{q, \hat{f}}(x, t', a) \right) s_{\zeta}(x, t', a) \right]\\
    ={}& \mE_{p(x, t) \pi_{\zeta}(a|x, t')} \left[  \frac{\ind_{\phi}(t, t')}{p(\phi(t'))} \left( \Delta_{q}(x, t, t', a) - \Delta_{\hat{f}}(x, t, t', a)  \right) s_{\zeta}(x, t', a) \right]
\end{align*}
\end{proof}

\subsection{Proof of Proposition \ref{prop.variance_grad_OPFV_true_p_phi}}
\label{proof_variance_grad_OPFV_true_p_phi}
\begin{proof}
We derive the variance of the gradient of the OPFV estimator under Conditions \ref{ass.full-support}, \ref{ass.common_time_feature_support}, and \ref{ass.conditional_piecewise_correctness} below.
\begin{align*}
    % & n \mV_{\calD} \left[ \opfvgrad \right] \\
    & n \var \left[ \opfvgrad \right] \\
    ={}& n \mV_{\calD}\left[\meanN \left\{ \frac{\ind_{\phi}(t_i, t')}{p(\phi(t'))} \frac{\pi_{\zeta}(a_i|x_i, t')}{\pi_0(a_i|x_i, t_i)} \left(r_i - \hat{f}(x_i, t_i, a_i)\right) s_{\zeta}(x_i, t', a_i) + \mE_{\pi_{\zeta}(a|x_i, t')}[\hat{f}(x_i, t', a) s_{\zeta}(x_i, t', a)] \right\}\right]  \\
    ={}& \mV_{p(x, t) \pi_0(a|x, t) p(r|x, t, a)} \left[  \frac{\ind_{\phi}(t, t')}{p(\phi(t'))} \frac{\pi_{\zeta}(a|x, t')}{\pi_0(a|x, t)} \left(r - \hat{f}(x, t, a)\right) s_{\zeta}(x, t', a) + \mE_{\pi_{\zeta}(a'|x, t')}[\hat{f}(x, t', a') s_{\zeta}(x, t', a')] \right] \\
    ={}& \mE_{p(x, t) \pi_0(a|x, t)} \left[ \mV_{p(r|x, t, a)} \left[ \frac{\ind_{\phi}(t, t')}{p(\phi(t'))} \frac{\pi_{\zeta}(a|x, t')}{\pi_0(a|x, t)} \left(r - \hat{f}(x, t, a)\right) s_{\zeta}(x, t', a) + \mE_{\pi_{\zeta}(a'|x, t')}[\hat{f}(x, t', a') s_{\zeta}(x, t', a')] \right] \right] \\
    &+ \mV_{p(x, t) \pi_0(a|x, t)} \left[ \mE_{p(r|x, t, a)} \left[ \frac{\ind_{\phi}(t, t')}{p(\phi(t'))} \frac{\pi_{\zeta}(a|x, t')}{\pi_0(a|x, t)} \left(r - \hat{f}(x, t, a)\right) s_{\zeta}(x, t', a) + \mE_{\pi_{\zeta}(a'|x, t')}[\hat{f}(x, t', a') s_{\zeta}(x, t', a')] \right] \right] \\ 
    & \quad \quad \because \text{law of total variance} \\
    ={}& \mE_{p(x, t) \pi_0(a|x, t)} \left[ \left( \frac{\ind_{\phi}(t, t')}{p(\phi(t'))} \frac{\pi_{\zeta}(a|x, t')}{\pi_0(a|x, t)} s_{\zeta}(x, t', a) \right)^2 \sigma^2(x, t, a) \right] \\
    &+ \mV_{p(x, t) \pi_0(a|x, t)} \left[\frac{\ind_{\phi}(t, t')}{p(\phi(t'))} \frac{\pi_{\zeta}(a|x, t')}{\pi_0(a|x, t)} \Delta_{q, \hat{f}}(x, t, a) s_{\zeta}(x, t', a) + \mE_{\pi_{\zeta}(a'|x, t')}[\hat{f}(x, t', a') s_{\zeta}(x, t', a')]\right] \\
    ={}& \mE_{p(x, t) \pi_0(a|x, t)} \left[ \left( \frac{\ind_{\phi}(t, t')}{p(\phi(t'))} \frac{\pi_{\zeta}(a|x, t')}{\pi_0(a|x, t)} s_{\zeta}(x, t', a) \right)^2 \sigma^2(x, t, a) \right] \\
    &+ \mE_{p(x, t)} \left[ \mV_{\pi_0(a|x, t)} \left[ \frac{\ind_{\phi}(t, t')}{p(\phi(t'))} \frac{\pi_{\zeta}(a|x, t')}{\pi_0(a|x, t)} \Delta_{q, \hat{f}}(x, t, a) s_{\zeta}(x, t', a) + \mE_{\pi_{\zeta}(a'|x, t')}[\hat{f}(x, t', a') s_{\zeta}(x, t', a')] \right] \right] \\
    &+ \mV_{p(x, t)} \left[ \mE_{\pi_0(a|x, t)} \left[ \frac{\ind_{\phi}(t, t')}{p(\phi(t'))} \frac{\pi_{\zeta}(a|x, t')}{\pi_0(a|x, t)} \Delta_{q, \hat{f}}(x, t, a) s_{\zeta}(x, t', a) + \mE_{\pi_{\zeta}(a'|x, t')}[\hat{f}(x, t', a') s_{\zeta}(x, t', a')] \right] \right] \\
    & \quad \quad \because \text{total var.} \\
    ={}& \mE_{p(x, t) \pi_0(a|x, t)} \left[ \left( \frac{\ind_{\phi}(t, t')}{p(\phi(t'))} \frac{\pi_{\zeta}(a|x, t')}{\pi_0(a|x, t)} s_{\zeta}(x, t', a) \right)^2 \sigma^2(x, t, a) \right] \\
    &+ \mE_{p(x, t)} \left[ \left( \frac{\ind_{\phi}(t, t')}{p(\phi(t'))} \right)^2 \mV_{\pi_0(a|x, t)} \left[ \frac{\pi_{\zeta}(a|x, t')}{\pi_0(a|x, t)} \Delta_{q, \hat{f}}(x, t, a) s_{\zeta}(x, t', a) \right] \right] \\
    &+ \mV_{p(x, t)} \left[ \mE_{\pi_{\zeta}(a|x, t')} \left[ \frac{\ind_{\phi}(t, t')}{p(\phi(t'))}  \Delta_{q, \hat{f}}(x, t, a) s_{\zeta}(x, t', a) \right] + \mE_{\pi_{\zeta}(a'|x, t')}[\hat{f}(x, t', a') s_{\zeta}(x, t', a')] \right] \\
    ={}& \mE_{p(x, t) \pi_0(a|x, t)} \left[ \left( \frac{\ind_{\phi}(t, t')}{p(\phi(t'))} \frac{\pi_{\zeta}(a|x, t')}{\pi_0(a|x, t)} s_{\zeta}(x, t', a) \right)^2 \sigma^2(x, t, a) \right] \\
    &+ \mE_{p(x, t)} \left[ \left( \frac{\ind_{\phi}(t, t')}{p(\phi(t'))} \right)^2 \mV_{\pi_0(a|x, t)} \left[ \frac{\pi_{\zeta}(a|x, t')}{\pi_0(a|x, t)} \Delta_{q, \hat{f}}(x, t, a) s_{\zeta}(x, t', a) \right] \right] \\
    &+ \mV_{p(x) p(t)} \left[ \mE_{\pi_{\zeta}(a|x, t')} \left[ \frac{\ind_{\phi}(t, t')}{p(\phi(t'))}  \Delta_{q, \hat{f}}(x, t, a) s_{\zeta}(x, t', a) \right] + \mE_{\pi_{\zeta}(a'|x, t')}[\hat{f}(x, t', a') s_{\zeta}(x, t', a')] \right] \\
    & \quad \quad \because \text{stationary context} \\
\end{align*}
\begin{align*}
    % & n \mV_{\calD} \left[ \opfvgrad \right] \\
    ={}& \mE_{p(x, t) \pi_0(a|x, t)} \left[ \left( \frac{\ind_{\phi}(t, t')}{p(\phi(t'))} \frac{\pi_{\zeta}(a|x, t')}{\pi_0(a|x, t)} s_{\zeta}(x, t', a) \right)^2 \sigma^2(x, t, a) \right] \\
    &+ \mE_{p(x, t)} \left[ \left( \frac{\ind_{\phi}(t, t')}{p(\phi(t'))} \right)^2 \mV_{\pi_0(a|x, t)} \left[ \frac{\pi_{\zeta}(a|x, t')}{\pi_0(a|x, t)} \Delta_{q, \hat{f}}(x, t, a) s_{\zeta}(x, t', a) \right] \right] \\
    &+ \mE_{p(x)} \left[ \mV_{p(t)} \left[ \mE_{\pi_{\zeta}(a|x, t')} \left[ \frac{\ind_{\phi}(t, t')}{p(\phi(t'))}  \Delta_{q, \hat{f}}(x, t, a) s_{\zeta}(x, t', a) \right] + \mE_{\pi_{\zeta}(a'|x, t')}[\hat{f}(x, t', a') s_{\zeta}(x, t', a')] \right] \right] \\
    &+ \mV_{p(x)} \left[ \mE_{p(t)} \left[ \mE_{\pi_{\zeta}(a|x, t')} \left[ \frac{\ind_{\phi}(t, t')}{p(\phi(t'))}  \Delta_{q, \hat{f}}(x, t, a) s_{\zeta}(x, t', a) \right] + \mE_{\pi_{\zeta}(a'|x, t')}[\hat{f}(x, t', a') s_{\zeta}(x, t', a')] \right] \right] \quad \because \text{total var.} \\
    ={}& \mE_{p(x, t) \pi_0(a|x, t)} \left[ \left( \frac{\ind_{\phi}(t, t')}{p(\phi(t'))} \frac{\pi_{\zeta}(a|x, t')}{\pi_0(a|x, t)} s_{\zeta}(x, t', a) \right)^2 \sigma^2(x, t, a) \right] \\
    &+ \mE_{p(x, t)} \left[ \left( \frac{\ind_{\phi}(t, t')}{p(\phi(t'))} \right)^2 \mV_{\pi_0(a|x, t)} \left[ \frac{\pi_{\zeta}(a|x, t')}{\pi_0(a|x, t)} \Delta_{q, \hat{f}}(x, t', a) s_{\zeta}(x, t', a) \right] \right] \\
    &+ \mE_{p(x)} \left[ \mV_{p(t)} \left[ \mE_{\pi_{\zeta}(a|x, t')} \left[ \frac{\ind_{\phi}(t, t')}{p(\phi(t'))}  \Delta_{q, \hat{f}}(x, t', a) s_{\zeta}(x, t', a) \right] \right] \right] \\
    &+ \mV_{p(x)} \left[ \mE_{p(t) \pi_{\zeta}(a|x, t')} \left[ \frac{\ind_{\phi}(t, t')}{p(\phi(t'))}  \Delta_{q, \hat{f}}(x, t', a) s_{\zeta}(x, t', a) \right] + \mE_{\pi_{\zeta}(a'|x, t')}[\hat{f}(x, t', a') s_{\zeta}(x, t', a')] \right] \quad \because \text{Cond. \ref{ass.conditional_piecewise_correctness}} \\
    ={}& \mE_{p(x, t) \pi_0(a|x, t)} \left[ \left( \frac{\ind_{\phi}(t, t')}{p(\phi(t'))} \frac{\pi_{\zeta}(a|x, t')}{\pi_0(a|x, t)} s_{\zeta}(x, t', a) \right)^2 \sigma^2(x, t, a) \right] \\
    &+ \mE_{p(x, t)} \left[ \left( \frac{\ind_{\phi}(t, t')}{p(\phi(t'))} \right)^2 \mV_{\pi_0(a|x, t)} \left[ \frac{\pi_{\zeta}(a|x, t')}{\pi_0(a|x, t)} \Delta_{q, \hat{f}}(x, t', a) s_{\zeta}(x, t', a) \right] \right] \\
    &+ \mE_{p(x)} \left[ \mE_{\pi_{\zeta}(a|x, t')} \left[ \Delta_{q, \hat{f}}(x, t', a) s_{\zeta}(x, t', a) \right]^2 \right] \cdot \mV_{p(t)} \left[ \frac{\ind_{\phi}(t, t')}{p(\phi(t'))} \right] \\
    &+ \mV_{p(x)} \left[ \mE_{\pi_{\zeta}(a|x, t')} \left[ \int_{t \in [0, T]} p(t) \frac{\ind_{\phi}(t, t')}{p(\phi(t'))}  \Delta_{q, \hat{f}}(x, t', a) s_{\zeta}(x, t', a) dt \right] + \mE_{\pi_{\zeta}(a'|x, t')}[\hat{f}(x, t', a') s_{\zeta}(x, t', a')] \right] \\
    ={}& \mE_{p(x, t) \pi_0(a|x, t)} \left[ \left( \frac{\ind_{\phi}(t, t')}{p(\phi(t'))} \frac{\pi_{\zeta}(a|x, t')}{\pi_0(a|x, t)} s_{\zeta}(x, t', a) \right)^2 \sigma^2(x, t, a) \right] \\
    &+ \mE_{p(x, t)} \left[ \left( \frac{\ind_{\phi}(t, t')}{p(\phi(t'))} \right)^2 \mV_{\pi_0(a|x, t)} \left[ \frac{\pi_{\zeta}(a|x, t')}{\pi_0(a|x, t)} \Delta_{q, \hat{f}}(x, t', a) s_{\zeta}(x, t', a) \right] \right] \\
    &+ \mE_{p(x)} \left[ \mE_{\pi_{\zeta}(a|x, t')} \left[ \Delta_{q, \hat{f}}(x, t', a) s_{\zeta}(x, t', a) \right]^2 \right] \cdot \mV_{p(t)} \left[ \frac{\ind_{\phi}(t, t')}{p(\phi(t'))} \right] \\
    &+ \mV_{p(x)} \left[ \mE_{\pi_{\zeta}(a|x, t')} \left[ \Delta_{q, \hat{f}}(x, t', a) s_{\zeta}(x, t', a) \right] + \mE_{\pi_{\zeta}(a'|x, t')}[\hat{f}(x, t', a') s_{\zeta}(x, t', a')] \right] \quad \because \text{cancel out $p(\phi(t'))$} \\
    ={}& \mE_{p(x, t) \pi_0(a|x, t)} \left[ \left( \frac{\ind_{\phi}(t, t')}{p(\phi(t'))} \frac{\pi_{\zeta}(a|x, t')}{\pi_0(a|x, t)} s_{\zeta}(x, t', a) \right)^2 \sigma^2(x, t, a) \right] \\
    &+ \mE_{p(x, t)} \left[ \left( \frac{\ind_{\phi}(t, t')}{p(\phi(t'))} \right)^2 \mV_{\pi_0(a|x, t)} \left[ \frac{\pi_{\zeta}(a|x, t')}{\pi_0(a|x, t)} \Delta_{q, \hat{f}}(x, t', a) s_{\zeta}(x, t', a) \right] \right] \\
    &+ \mE_{p(x)} \left[ \mE_{\pi_{\zeta}(a|x, t')} \left[ \Delta_{q, \hat{f}}(x, t', a) s_{\zeta}(x, t', a) \right]^2 \right] \cdot \mV_{p(t)} \left[ \frac{\ind_{\phi}(t, t')}{p(\phi(t'))} \right] + \mV_{p(x)} \left[ \mE_{\pi_{\zeta}(a|x, t')} \left[ q(x, t', a) s_{\zeta}(x, t', a) \right] \right] \\
    & \quad \because \text{cancel out $\hat{f}(x, t', a)$} \\
\end{align*}
\end{proof}

\subsection{Proof of Theorem \ref{thm.bias_extended_OPFV}}
\label{proof_bias_extended_OPFV}
\begin{proof}
    We derive the bias of the extended OPFV estimator for F-OPE under non-stationary context and reward distributions under Conditions \ref{ass.common_support}, \ref{ass.common_time_feature_support}, and \ref{ass.conditional_stationarity_wrt_context} below.
\begin{align*}
    & \bias \left( \opfv \right) \\
    ={}& \mE_{\calD} \left[ \meanN \left\{ \frac{\ind_{\phi_{x, r}}(t_i, t')}{p(\phi_{x, r}(t'))} \frac{\pi_e(a_i|x_i, t')}{\pi_0(a_i|x_i, t_i)} \left( r_i - \hat{f}(x_i, t_i, a_i) \right) +  \frac{\ind_{\phi_x}(t_i, t')}{p(\phi_x(t'))} \mE_{\pi_e(a|x_i, t')} [\hat{f}(x_i, t', a)] \right\} \right] - V_{t'}(\pi_e) \\
    ={}& \mE_{p(t) p(x|t) \pi_0(a|x, t) p(r|x, t, a)}  \left[ \frac{\ind_{\phi_{x, r}}(t, t')}{p(\phi_{x, r}(t'))} \frac{\pi_e(a|x, t')}{\pi_0(a|x, t)} \left( r - \hat{f}(x, t, a) \right) +  \frac{\ind_{\phi_x}(t, t')}{p(\phi_x(t'))} \mE_{\pi_e(a'|x, t')} [\hat{f}(x, t', a')]  \right] - V_{t'}(\pi_e) \\
    ={}& \mE_{p(t) p(x|t) \pi_0(a|x, t)}  \left[ \frac{\ind_{\phi_{x, r}}(t, t')}{p(\phi_{x, r}(t'))} \frac{\pi_e(a|x, t')}{\pi_0(a|x, t)} \Delta_{q, \hat{f}}(x, t, a) +  \frac{\ind_{\phi_x}(t, t')}{p(\phi_x(t'))} \mE_{\pi_e(a'|x, t')} [\hat{f}(x, t', a')]  \right] \notag \\
    & - \mE_{p(x|t') \pi_e(a|x, t')}[q(x, t', a)]  \\
    ={}& \mE_{p(t) p(x|t) \pi_e(a|x, t')}  \left[ \frac{\ind_{\phi_{x, r}}(t, t')}{p(\phi_{x, r}(t'))} \Delta_{q, \hat{f}}(x, t, a) +  \frac{\ind_{\phi_x}(t, t')}{p(\phi_x(t'))} \mE_{\pi_e(a'|x, t')} [\hat{f}(x, t', a')]  \right] - \mE_{p(x|t') \pi_e(a|x, t')}[q(x, t', a)]  \\
    ={}& \mE_{p(t) p(x|t)}  \left[ \frac{\ind_{\phi_{x, r}}(t, t')}{p(\phi_{x, r}(t'))} \mE_{\pi_e(a|x, t')} [ \Delta_{q, \hat{f}}(x, t, a) ]\right] + \mE_{p(t) p(x|t)}  \left[ \frac{\ind_{\phi_x}(t, t')}{p(\phi_x(t'))} \mE_{\pi_e(a'|x, t')} [\hat{f}(x, t', a')] \right] \notag \\
    & - \mE_{p(x|t') \pi_e(a|x, t')}[q(x, t', a)]  \\
    ={}& \mE_{p(t) p(x|t)}  \left[ \frac{\ind_{\phi_{x, r}}(t, t')}{p(\phi_{x, r}(t'))} \mE_{\pi_e(a|x, t')} [ \Delta_{q, \hat{f}}(x, t, a) ]\right] 
    + \int_{t \in [0, T]} p(t) \int_{x \in \calX} p(x|t) \frac{\ind_{\phi_x}(t, t')}{p(\phi_x(t'))} \mE_{\pi_e(a'|x, t')} [\hat{f}(x, t', a')] dx dt \\
    &- \mE_{p(x|t') \pi_e(a|x, t')}[q(x, t', a)] \\
    ={}& \mE_{p(t) p(x|t)}  \left[ \frac{\ind_{\phi_{x, r}}(t, t')}{p(\phi_{x, r}(t'))} \mE_{\pi_e(a|x, t')} [ \Delta_{q, \hat{f}}(x, t, a) ]\right] \notag \\
    & + \alpha \int_{t \in [0, T]} p(t) \int_{x \in \calX} p_1(x|\phi_x(t))  \frac{\ind_{\phi_x}(t, t')}{p(\phi_x(t'))} \mE_{\pi_e(a'|x, t')} [\hat{f}(x, t', a')] dx dt \\
    &+ (1 - \alpha) \int_{t \in [0, T]} p(t) \int_{x \in \calX}  p_2(x|t) \frac{\ind_{\phi_x}(t, t')}{p(\phi_x(t'))} \mE_{\pi_e(a'|x, t')} [\hat{f}(x, t', a')] dx dt - \mE_{p(x|t') \pi_e(a|x, t')}[q(x, t', a)] \quad \because \text{Cond. \ref{ass.conditional_stationarity_wrt_context}} \\
    ={}& \mE_{p(t) p(x|t)}  \left[ \frac{\ind_{\phi_{x, r}}(t, t')}{p(\phi_{x, r}(t'))} \mE_{\pi_e(a|x, t')} [ \Delta_{q, \hat{f}}(x, t, a) ]\right] 
    + \alpha \int_{x \in \calX} p_1(x|\phi_x(t)) \mE_{\pi_e(a'|x, t')} [\hat{f}(x, t', a')] dx \\
    &+ (1 - \alpha) \int_{x \in \calX}  p_2(x|t)  \mE_{\pi_e(a'|x, t')} [\hat{f}(x, t', a')] dx 
    - \mE_{p(x|t') \pi_e(a|x, t')}[q(x, t', a)] \quad \because \text{def. of  $p(\phi_x(t')$} \\
    ={}& \mE_{p(t) p(x|t)}  \left[ \frac{\ind_{\phi_{x, r}}(t, t')}{p(\phi_{x, r}(t'))} \mE_{\pi_e(a|x, t')} [ \Delta_{q, \hat{f}}(x, t, a) ]\right] 
    + \mE_{p(x|t') \pi_e(a|x, t')}[\hat{f}(x, t', a)]
    - \mE_{p(x|t') \pi_e(a|x, t')}[q(x, t', a)]  \\
    ={}& \mE_{p(t) p(x|t')}  \left[ \frac{\ind_{\phi_{x, r}}(t, t')}{p(\phi_{x, r}(t'))} \mE_{\pi_e(a|x, t')} [ \Delta_{q, \hat{f}}(x, t, a) ]\right] 
    - \mE_{p(x|t') \pi_e(a|x, t')}[\Delta_{q, \hat{f}}(x, t', a)] \quad \because \text{Cond. \ref{ass.conditional_stationarity_wrt_context}} \\
    ={}& \mE_{p(x|t') \pi_e(a|x, t')}  \left[ \frac{1}{p(\phi_{x, r}(t')} \left( \mE_{p(t)}[\ind_{\phi_{x, r}}(t, t') \Delta_{q, \hat{f}}(x, t, a)] - p(\phi_{x, r}(t')) \Delta_{q, \hat{f}}(x, t, a)] \right) \right]\\
    ={}& \mE_{p(x|t') \pi_e(a|x, t')}  \left[ \frac{1}{p(\phi_{x, r}(t')} \left( \mE_{p(t)}[\ind_{\phi_{x, r}}(t, t') \Delta_{q, \hat{f}}(x, t, a)] - \mE_{p(t)}[\ind_{\phi_{x, r}}(t, t')] \Delta_{q, \hat{f}}(x, t, a)] \right) \right] \quad \because \text{def. of $p(\phi_{x, r}(t')$} \\
    ={}& \mE_{p(t) p(x|t') \pi_e(a|x, t')}  \left[ \frac{\ind_{\phi_{x, r}}(t, t')}{p(\phi_{x, r}(t')} \left( \Delta_{q, \hat{f}}(x, t, a) - \Delta_{q, \hat{f}}(x, t', a)\right) \right]  \\
    ={}& \mE_{p(x, t) \pi_e(a|x, t')}  \left[ \frac{\ind_{\phi_{x, r}}(t, t')}{p(\phi_{x, r}(t')} \left( \Delta_q(x, t, t', a) - \Delta_{\hat{f}}(x, t, t', a)\right) \right] \quad \because \text{Condition \ref{ass.conditional_stationarity_wrt_context}} \\
\end{align*}
\end{proof}

\subsection{Proof of Proposition \ref{prop.unbiasedness_extended_OPFV}}
\label{proof_unbiasedness_extended_OPFV}
\begin{proof}
    We prove that the extended OPFV estimator for F-OPE under non-stationary context and reward distributions is unbiased under Conditions \ref{ass.common_support}, \ref{ass.common_time_feature_support}, \ref{ass.conditional_stationarity_wrt_context}, and \ref{ass.conditional_piecewise_correctness} below.
\begin{align*}
    & \mE_{\calD} \left[ \opfv \right] \\
    ={}& \mE_{\calD} \left[ \meanN \left\{ \frac{\ind_{\phi_{x, r}}(t_i, t')}{p(\phi_{x, r}(t'))} \frac{\pi_e(a_i|x_i, t')}{\pi_0(a_i|x_i, t_i)} \left( r_i - \hat{f}(x_i, t_i, a_i) \right) +  \frac{\ind_{\phi_x}(t_i, t')}{p(\phi_x(t'))} \mE_{\pi_e(a|x_i, t')} [\hat{f}(x_i, t', a)] \right\} \right] \\
    ={}& \mE_{p(t) p(x|t) \pi_0(a|x, t) p(r|x, t, a)}  \left[ \frac{\ind_{\phi_{x, r}}(t, t')}{p(\phi_{x, r}(t'))} \frac{\pi_e(a|x, t')}{\pi_0(a|x, t)} \left( r - \hat{f}(x, t, a) \right) +  \frac{\ind_{\phi_x}(t, t')}{p(\phi_x(t'))} \mE_{\pi_e(a'|x, t')} [\hat{f}(x, t', a')]  \right] \quad \because \text{i.i.d.} \\
    ={}& \mE_{p(t) p(x|t) \pi_0(a|x, t)}  \left[ \frac{\ind_{\phi_{x, r}}(t, t')}{p(\phi_{x, r}(t'))} \frac{\pi_e(a|x, t')}{\pi_0(a|x, t)} \Delta_{q, \hat{f}}(x, t, a) +  \frac{\ind_{\phi_x}(t, t')}{p(\phi_x(t'))} \mE_{\pi_e(a'|x, t')} [\hat{f}(x, t', a')]  \right] \\
    ={}& \mE_{p(t) p(x|t) \pi_0(a|x, t)}  \left[ \frac{\ind_{\phi_{x, r}}(t, t')}{p(\phi_{x, r}(t'))} \frac{\pi_e(a|x, t')}{\pi_0(a|x, t)} \Delta_{q, \hat{f}}(x, t', a) +  \frac{\ind_{\phi_x}(t, t')}{p(\phi_x(t'))} \mE_{\pi_e(a'|x, t')} [\hat{f}(x, t', a')]  \right] \quad \because \text{Condition \ref{ass.conditional_piecewise_correctness}} \\
    ={}& \mE_{p(t) p(x|t)}  \left[ \frac{\ind_{\phi_{x, r}}(t, t')}{p(\phi_{x, r}(t'))} \mE_{\pi_e(a|x, t')} [\Delta_{q, \hat{f}}(x, t', a) ] +  \frac{\ind_{\phi_x}(t, t')}{p(\phi_x(t'))} \mE_{\pi_e(a|x, t')} [\hat{f}(x, t', a)]  \right] \quad \because \text{cancel out $\pi_0(a|x, t')$} \\
    ={}& \int_{t \in [0, T]} p(t) \int_{x \in \calX} p(x|t) \left\{ \frac{\ind_{\phi_{x, r}}(t, t')}{p(\phi_{x, r}(t'))} \mE_{\pi_e(a|x, t')} [\Delta_{q, \hat{f}}(x, t', a) ] +  \frac{\ind_{\phi_x}(t, t')}{p(\phi_x(t'))} \mE_{\pi_e(a|x, t')} [\hat{f}(x, t', a)]   \right\} dx dt \\
    ={}& \int_{t \in [0, T]} p(t) \int_{x \in \calX} \left\{ \alpha p_1(x|\phi_x(t)) + (1 - \alpha) p_2(x|t) \right\} \Big\{\frac{\ind_{\phi_{x, r}}(t, t')}{p(\phi_{x, r}(t'))} \mE_{\pi_e(a|x, t')} [\Delta_{q, \hat{f}}(x, t', a) ] \\
    & +  \frac{\ind_{\phi_x}(t, t')}{p(\phi_x(t'))} \mE_{\pi_e(a|x, t')} [\hat{f}(x, t', a)] \Big\} dx dt  \\
    ={}& \alpha \int_{t \in [0, T]} p(t) \int_{x \in \calX} p_1(x|\phi_x(t')) \left\{ \frac{\ind_{\phi_{x, r}}(t, t')}{p(\phi_{x, r}(t'))} \mE_{\pi_e(a|x, t')} [\Delta_{q, \hat{f}}(x, t', a) ] +  \frac{\ind_{\phi_x}(t, t')}{p(\phi_x(t'))} \mE_{\pi_e(a|x, t')} [\hat{f}(x, t', a)] \right\} dx dt \\
    &+ (1 - \alpha) \int_{t \in [0, T]} p(t) \int_{x \in \calX} p_2(x|t) \left\{\frac{\ind_{\phi_{x, r}}(t, t')}{p(\phi_{x, r}(t'))} \mE_{\pi_e(a|x, t')} [\Delta_{q, \hat{f}}(x, t', a) ] +  \frac{\ind_{\phi_x}(t, t')}{p(\phi_x(t'))} \mE_{\pi_e(a|x, t')} [\hat{f}(x, t', a)] \right\} dx dt \\
    ={}& \alpha \mE_{p_1(x|\phi_x(t)) \pi_e(a|x, t')} \left[ \Delta_{q, \hat{f}}(x, t', a) + \hat{f}(x, t', a) \right] \\
    &+ (1 - \alpha) \int_{t \in [0, T]} p(t) \int_{x \in \calX} p_2(x|t) \left\{\frac{\ind_{\phi_{x, r}}(t, t')}{p(\phi_{x, r}(t'))} \mE_{\pi_e(a|x, t')} [\Delta_{q, \hat{f}}(x, t', a) ] +  \frac{\ind_{\phi_x}(t, t')}{p(\phi_x(t'))} \mE_{\pi_e(a|x, t')} [\hat{f}(x, t', a)] \right\} dx dt \\
    &\quad \because \text{cancel out $p(\phi_{x, r}(t'))$ and $p(\phi_r(t'))$} \\
    ={}& \alpha \mE_{p_1(x|\phi_x(t')) \pi_e(a|x, t')} \left[ \Delta_{q, \hat{f}}(x, t', a) + \hat{f}(x, t', a) \right] + (1 - \alpha) \mE_{p_2(x|t) \pi_e(a|x, t')} \left[ \Delta_{q, \hat{f}}(x, t', a) + \hat{f}(x, t', a) \right] \notag \\
    & \quad \quad \because \text{Condition \ref{ass.conditional_stationarity_wrt_context}}\\
    ={}& \alpha \mE_{p_1(x|\phi_x(t')) \pi_e(a|x, t')} \left[q(x, t', a) \right] + (1 - \alpha) \mE_{p_2(x|t) \pi_e(a|x, t')} \left[q(x, t', a) \right] \\
    ={}& G_{\phi_x(t')}(\pi_e) + H_{t'}(\pi_e) \\
    ={}& V_{t'}(\pi_e) \\
\end{align*}
\end{proof}

\section{Additional setup and results for non-stationary synthetic experiment} \label{app:additional}

\subsection{Additional experimental setup for F-OPE and F-OPL under non-stationary reward distribution}\label{app:additional-exp-setup-all}

\subsubsection{Synthetic expected reward function}
\label{app:additional_setup}
In this section, we describe how we synthesize the simplified and controllable reward function $q(x, t, a)$ utilized in Section \ref{sec:experiment} in detail. First, we synthesize the expected reward $q(x, t, a)$ by the weighted sum of the time feature effect $g(x, \phi(t), a)$ and the residual effect $h(x, t, a)$ as follows.

\begin{align*}
    q(x, t, a) = \lambda \cdot g(a, \phi(t), a) + (1 - \lambda) \cdot h(x, t, a)
\end{align*}
Specifically, we construct each effect by the following functions.

\begin{align*}
    g(x, \phi(t), a) &= \nu_x^\top s_{g_1}(x) + \nu_{\phi(t)}^\top \onehottimefeature + \onehottimefeaturetranspose M_{\phi(t), a} \onehotaction \\
    +{}& s_{g_2}(x)^\top M_{x, \phi(t), a} \onehotactiontimefeature \\
    h(x, t, a) &= \xi_x^\top s_{h_1}(x) + \xi_{\phi_f(t)}^\top \onehottimefeaturefiner + \xi_a^\top \onehotaction + \onehottimefeaturefinertranspose M_{\phi_f(t), a} \onehotaction \\
    +{}& s_{h_2}(x)^\top M_{x, a} \onehotaction + s_{h_3}(x)^\top M_{x, \phi_f(t), a} \onehotactiontimefeaturefiner, 
\end{align*}
where $\onehotaction \in \{0, 1\}^{|\calA|}$, $\onehottimefeature \in \{0, 1\}^{|\calC|}$, $\onehottimefeaturefiner  \in \{0, 1\}^{|\calC_f|}$ are the one-hot encoding of action $a$, time feature $\phi$, finer time feature $\phi_f$, respectively and $\calC$ and $\calC_f$ are the sets of unique time features used in the time feature effect and residual effect. We use the time feature function $\phi$, which divides each year into eight seasons with $|\calC| = 8$, and the finer time feature $\phi_f$ that maps the timestamp $t$ to the day of the week with $|\calC_f| = 7$. $\onehotactiontimefeature \in \{0, 1\}^{|\calC \times \calA|}$, and $\onehotactiontimefeaturefiner \in \{0, 1\}^{|\calC_f \times \calA|}$ are the one-hot encoding of time feature $\phi(t)$ and action $a$ and finer time feature $\phi_f(t)$ and action $a$. Furthermore, we sample each element of the parameters used in the time feature effect, $\nu_x, \nu_{\phi(t)}, M_{\phi(t), a}$, and $M_{x, \phi(t), a}$ from a uniform distribution ranging from -3 to 3 and the parameters used in residual effect, $\xi_x, \xi_{\phi_f(t)}, \xi_a, M_{\phi_f(t), a}, M_{x, a}$, and $M_{x, \phi_f(t), a}$ from a uniform distribution ranging from -1 to 1. $s_{g_1}(x), s_{g_2}(x), s_{h_1}(x), s_{h_2}(x)$, and $s_{h_3}(x)$ are defined as follows:
\begin{align*}
    s_{g_1}(x) &= \left( \ind \left\{ \sum_{d=1}^4 x_d < 1.5 \right\}, \ind \left\{ \sum_{d=6}^9 x_d < -0.5 \right\}, \ind \left\{ \sum_{d=4}^5 x_d > 3.0 \right\}, \ind \left\{ \sum_{d=7}^{10} x_d > 3.0 \right\} \right)^\top \\
    % s_{g_5}(x) &= \left( \ind \left\{ \sum_{d=1}^7 x_d > 3.0 \right\}, \ind \left\{ \sum_{d=7}^{10} x_d < -2.5 \right\} \right)^\top \\
    s_{g_2}(x) &= \left( \ind \left\{ \sum_{d=1}^4 x_d < 4.0 \right\}, \ind \left\{ \sum_{d=6}^9 x_d > 3.0 \right\}, \ind \left\{ \sum_{d=3}^{10} x_d < -2.5 \right\} \right)^\top \\
    s_{h_1}(x) &= \left( \ind \left\{ \sum_{d=1}^6 x_d < 2.5 \right\}, \ind \left\{ \sum_{d=8}^9 x_d < -0.5 \right\}, \ind \left\{ \sum_{d=3}^5 x_d > 2.0 \right\} \right)^\top \\
    s_{h_2}(x) &= \left( \ind \left\{ \sum_{d=1}^4 x_d < 3.0 \right\}, \ind \left\{ \sum_{d=3}^9 x_d > 2.5 \right\}, \ind \left\{ \sum_{d=2}^7 x_d  < 1.5 \right\}, \ind \left\{ \sum_{d=7}^{10} x_d > -1.5 \right\} \right)^\top \\
    s_{h_3}(x) &= \left( \ind \left\{ \sum_{d=1}^4 x_d < 4.0 \right\}, \ind \left\{ \sum_{d=3}^9 x_d > 3.5 \right\}, \ind \left\{ \sum_{d=3}^5 x_d > 1.5 \right\} , \ind \left\{ \sum_{d=6}^{10} x_d < 2.5 \right\} \right)^\top, \\
\end{align*}
where $x_d$ is the $d$-th dimension of the context $x$.

\subsubsection{Optimization of time feature function for the OPFV estimator}
\label{app:synthetic-time-feature-opt}
\begin{figure*}[t]
\centering
\vspace{1mm}
\includegraphics[clip, width=14cm]{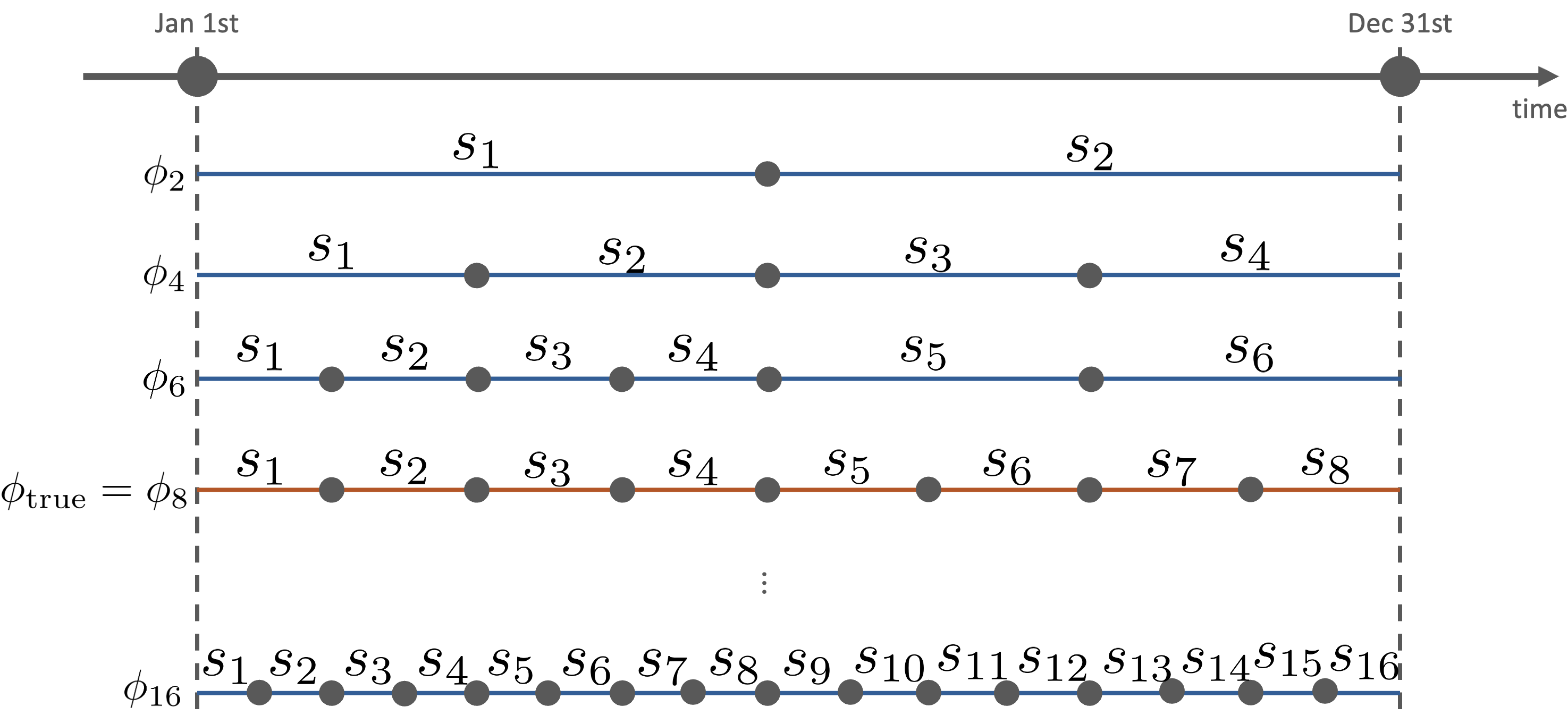}
\vspace{-3mm}
\caption{Comparing the true time feature $\phi_{\text{true}} (= \phi_8)$ and the time feature functions $\phi \in \Phi$ used in the optimization procedure in Section~\ref{sec:tune} for the OPFV estimator where $\Phi := \{ \phi_{2i} : \forall i = 1, 2, \cdots, 8\}$ for any year} \label{fig:synthetic-exp-time-feature}
\raggedright
\end{figure*}

This section describes the true time feature function $\phi_{\text{true}}$ used for the data-generating process as well as the set $\Phi$ of candidate time features used in the OPFV estimator in the synthetic experiment in Section~\ref{sec:synthetic-data}. As briefly described in Section~\ref{sec:synthetic-data}, the true time feature function $\phi_{\text{true}}: [0, t'] \rightarrow \{ s_1, s_2, \cdots, s_8 \}$ evenly divide each year into eight seasons where the $i$-th season is denoted by $s_i$. Figure~\ref{fig:synthetic-exp-time-feature} illustrates the time feature functions used for the optimization procedure of the OPFV estimator described in Section~\ref{sec:tune}. Specifically, we used the set $\Phi := \{ \phi_2, \phi_4, \phi_6,  \cdots, \phi_{16} \}$, where $\phi_i$ is strictly finer than $\phi_j$ for all $i, j \in \{2, 4, 6, \cdots, 16\}$ if $i > j$. For instance, for any $i = 1, 2, 3, 4$, $\phi_{2^i}$ evenly divide each year into $2^i$ seasons. We use this set $\Phi$ to tune the time feature $\hat{\phi}$ for OPFV (tuned $\hat{\phi}$), and we use the finest time feature function $\phi_{\infty} = \phi_{16}$ to estimate the bias described in Section~\ref{sec:tune}. This definition of the set of time feature functions is helpful in observing the empirical bias-variance tradeoff of the OPFV estimator in terms of the crudeness of the time features to support the claim in theoretical analyses in Section~\ref{sec:analysis}.

\subsubsection{Additional baseline estimator for F-OPE}

\textbf{The Prognosticator $\phi$ estimator: }
To show which components in the OPFV estimator in Section~\ref{sec:opfv} improve the MSE compared to the Prognosticator estimator, we introduce the extended Prognosticator estimator named the Prognosticator $\phi$ estimator in F-OPE, which uses the time feature function $\phi$, that is, one of our contributions to the F-OPE/F-OPL, rather than the basis function $\psi$ originally proposed in Prognosticator Chandak et al.~\citep{chandak2020optimizing}. Concretely, we introduce the time feature function for Prognosticator $\phi_p: [K + \delta] \rightarrow \calC$ where $\phi_p$ maps the time period $k \in [K+\delta]$ to a time-series feature $c \in \calC$. Then, the Prognosticator $\phi$ estimator for estimating the true value $V_{K + \delta}(\pi_e)$ in the future period $K + \delta$ is defined as follows.
\begin{align*}
    \hat{V}_{K + \delta}^{\text{Prognosticator $\phi$}}(\pi_e; \calD) := \onehottimefeaturephi(K + \delta)^\top (\Phi_p^T\Phi_p)^{-1} \Phi_p^\top Y
\end{align*}
where $\Phi_p :=(\onehottimefeaturephi(1), \cdots, \onehottimefeaturephi(K))^\top \in \mR^{K \times |\calC|}$ and $Y := (\hat{V}_1(\pi_e; \calD), \cdots, \hat{V}_K(\pi_e; \calD))^\top \in \mR^K$. Note that $\onehottimefeaturephi(k) \in \mR^{|\calC|}$ is the one-hot encoding of the time features $\phi_p(k)$ of the period $k \in [K+\delta]$. This estimator will show how much improvement in the accuracy of the OPFV estimator compared to the Prognosticator estimator can be attributed to the use of the time feature function $\phi$. 

As one of the limitations of the Prognosticator estimator $\hat{V}_{K + \delta} (\pi_e; \calD)$ in Section~\ref{sec:prognosticator} was the unreliable extrapolation (regression) phase, which often incurs significant bias because capturing the periodic trends by using time step index $k$ and basis function $\psi$ is challenging, the Prognosticator $\phi$ is expected to reduce its bias compared to the original Prognosticator estimator in Section~\ref{sec:prognosticator}.

\subsubsection{Baselines methods for F-OPL}\label{app:baselines-methods-for-F-OPL}
In our synthetic data experiment for F-OPL, we compare OPFV-PG with the Regression-based method and policy gradient (PG) methods, including IPS-PG and DR-PG as the conventional OPL methods for stationary environments and Prognosticator as a baseline method for non-stationary environments. We describe these baseline methods in more detail below.

\textbf{Regression-based Method: }
The regression-based method first estimates the expected reward function $q(x, a)$ by some supervised machine learning, where we use a random forest regressor with ten trees. Then, based on the estimated expected reward function $\hat{q}_{\zeta}(x, a)$ parameterized by $\zeta$, it outputs a stochastic decision $\pi_{\zeta}(a|x)$. For instance, we can use the softmax function to obtain the learned policy $\pi_{\zeta}(a|x) = \frac{\exp(\beta \cdot \hat{q}(x, a))}{\sum_{a' \in \calA} \exp(\beta \cdot \hat{q}(x, a'))}$ where $\beta$ is a parameter that controls the stochasticity of the policy $\pi_{\zeta}$. In our experiments, we set the parameter $\beta=10$. One of the statistical advantages of the regression-based method is that it often has a lower bias than methods with importance weights. However, it suffers from significant bias due to the misspecification of the expected reward function $\hat{q}(x, a)$, particularly when the logged data size $n$ is limited compared to the number of pairs of context $x$ and action $a$ in stationary environments. Such a situation is often the case with practical applications~\citep{udagawa2023policy, gao2022kuairec, saito2021open, farajtabar2018more}. In non-stationary environments, the bias of the estimator of the expected reward $\hat{q}(x, a)$ considerably increases since it does not consider the distributional shift of the context and reward in the $\hat{q}$ function (i.e., $q(x, a) \ne q(x, t', a)$ for target time $t'$), failing to learn a policy for the future from the logged data effectively.

\textbf{Inverse Propensity Scoring Policy Gradient (IPS-PG): }
In contrast to the regression-based method, PG methods use estimators of the gradient of the policy value $\nabla_{\zeta} V(\pi_{\zeta})$ to optimize the policy $\pi_{\zeta}$ parameterized by $\zeta$ via the gradient ascent $\zeta_{\tau + 1} \gets \zeta_{\tau} + \eta \nabla_{\zeta} V(\pi_{\zeta_{\tau}})$ where $\eta$ is a learning rate in stationary environments. One of the typical choices of the estimator is IPS-PG:
\begin{align*}
    \nabla_{\zeta} \hat{V}^{\text{IPS}}(\pi_{\zeta}; \calD) := \meanN \frac{\pi_{\zeta}(a_i|x_i)}{\pi_0(a_i|x_i)} r_i s_{\zeta}(x_i, a_i), 
\end{align*}
where $s_{\zeta}(x, a) := \nabla_{\zeta} \log \pi_{\zeta}(a|x)$ is the policy score function. In stationary environments, the IPS-PG estimator is unbiased if the logging policy $\pi_0$ assigns a non-zero probability to any action $a \in \calA$ for arbitrary context $x \in \calX$. In non-stationary environments, however, it is no longer unbiased as the true gradient of the policy value varys over the time (i.e., $\nabla_{\zeta} V(\pi_{\zeta}) \ne \nabla_{\zeta} V_{t'}(\pi_{\zeta})$) for target time $t'$; thus, IPS-PG fails to improve a policy for the future.

\textbf{Doubly Robust Policy Gradient (DR-PG): }
DR-PG employs a doubly robust structure to construct an estimator of the gradient of the policy value as follows.

\begin{align*}
    \nabla_{\zeta} \hat{V}^{\text{DR}}(\pi_{\zeta}; \calD, \hat{q}) := \meanN \left\{ \frac{\pi_{\zeta}(a_i|x_i)}{\pi_0(a_i|x_i)} (r_i - \hat{q}(x_i, a_i) s_{\zeta}(x_i, a_i) + \mE_{\pi_{\zeta}(a|x_i)} [\hat{q}(x_i, a) s_{\zeta}(x_i, a)] \right\}, 
\end{align*}
The doubly robust structure reduces the estimator's variance by using $q(x, a)$ as a control variate in stationary environments. Nonetheless, the DR-PG estimator incurs significant bias under distributional shifts of the context and reward. The failure to evaluate the gradient of the policy value leads to the unreliable F-OPL.

\begin{figure*}[t]
\centering
    \begin{subfigure}
        \centering
        \vspace{1mm}
        \includegraphics[clip, width=13.5cm]{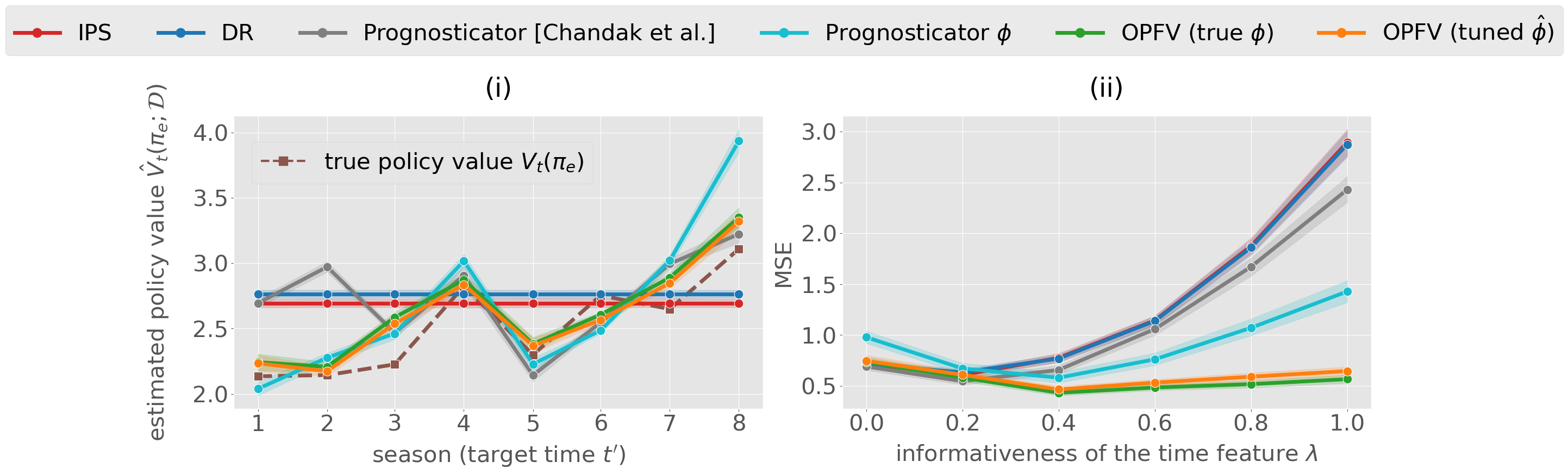}
        \vspace{-3mm}
        \caption{Comparing (i) the true value $V_t(\pi_e)$ and the estimated policy values $\hat{V}_t(\pi_e; \calD)$ with varying target time range $t'$ and (ii) the MSE of estimators with varying $\lambda$ values that controls the informativeness of the time feature} \label{fig:f-ope-target-time-lambda-prog-phi}
        \raggedright
    \end{subfigure}
    \vspace{5mm}
    \begin{subfigure}
        \centering
        \vspace{1mm}
        \includegraphics[clip, width=13.5cm]{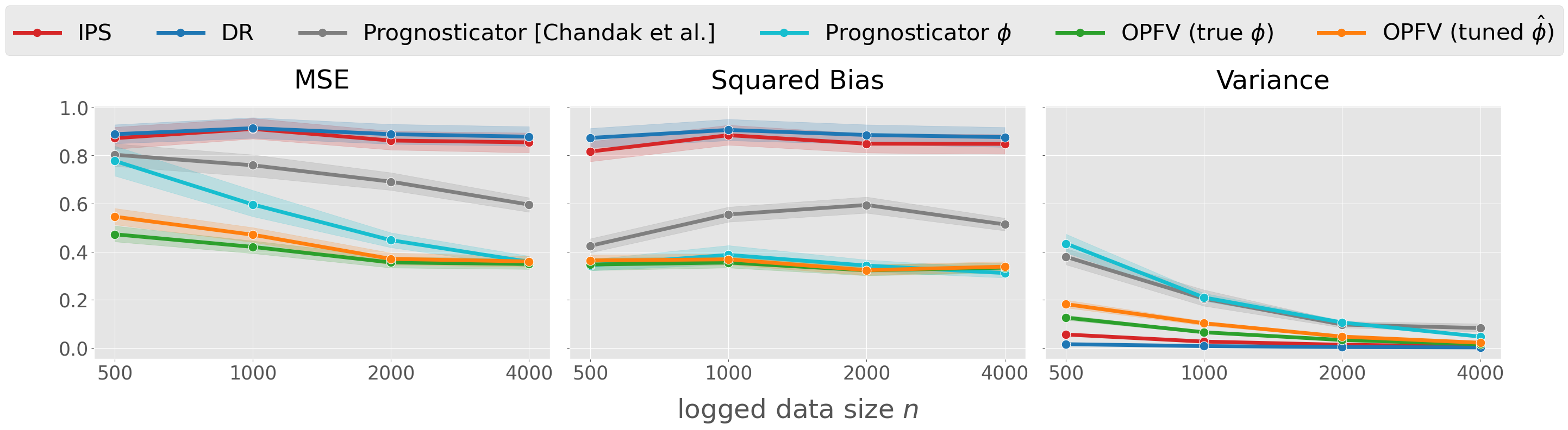}
        \vspace{-3mm}
        \caption{Comparison of the MSE, squared bias, and variance of estimators in an F-OPE problem with varying sample sizes $n$} \label{fig:f-ope-n-trains-prog-phi}
        \raggedright
    \end{subfigure}
\end{figure*}

\textbf{Prognosticator~\citep{chandak2020optimizing}: }
One of the baseline methods in abrupt non-stationary environments is Prognosticator. In contextual bandit protocol, Prognosticator first divides the logged data $\calD$ into $K$ distinct sub-dataset $\calD = \{ \calD_k \}_{k \in [K]}$, where it assumes the stationarity within each subdata $\calD_k$ for any period $k \in [K]$. Then, it uses the following estimator of the gradient of the policy value for future period $K + \delta$: 
\begin{align*}
    \nabla_{\zeta} \hat{V}^{\text{Prognosticator}}_{K+\delta}(\pi_{\zeta}; \calD) 
    &:= \sum_{k=1}^K \underbrace{\frac{\partial \hat{V}_{K+\delta}(\pi_{\zeta}; \calD)}{\partial \hat{V}^{\text{IPS}}_k(\pi_{\zeta}; \calD_k)}}_{\text{(i) extrapolation phase}} \cdot \underbrace{\nabla_{\zeta} \hat{V}^{\text{IPS}}_{k}(\pi_{\zeta}; \calD_k)}_{\text{(ii) estimation phase}} \\
    &= \sum_{k=1}^K \left[ \psi(K + \delta) (\Psi^\top \Psi)^{-1} \Psi^\top \right]_k \nabla_{\zeta} \hat{V}^{\text{IPS}}_{k}(\pi_{\zeta}; \calD_k), 
\end{align*}
where $\left[ \psi(K + \delta) (\Psi^\top \Psi)^{-1} \Psi^\top \right]_k$ denotes the $k$-th dimension of $\psi(K + \delta) (\Psi^\top \Psi)^{-1} \Psi^\top \in \mR^K$ and we set $K = 8$ as we consider eight distinct time features in each year in the synthetic data experiments. For a basis function $\psi$, we use a Fourier basis $\psi(k) := \left( \sin{( 2 \pi \frac{k} {K + \delta})}, \cdots, \sin{( 2 \pi \frac{ d' k} {K + \delta})}, 1, \cos{( 2 \pi \frac{k} {K + \delta})}, \cdots, \cos{( 2 \pi \frac{d' k} {K + \delta})} \right)^\top \in \mR^{2d' + 1}$. We consider $d'$ as the hyperparameter and let Prognosticator tune this hyperparameter from $d' \in [3, 5, 7]$ (same sets as~\citep{chandak2020optimizing}) by using the true policy value. For (ii) estimation phase of the gradient of the policy value in period $k$, Prognosticator uses the IPS-PG estimator to estimate the gradient of the policy value for each period $k \in [K]$ as follows.
\begin{align*}
    \nabla_{\zeta} \hat{V}^{\text{IPS}}_{k}(\pi_{\zeta}; \calD_k) := \frac{1}{n_k} \sum_{i = 1}^{n_k} \frac{\pi_{\zeta}(a_{i, k}|x_{i, k})}{\pi_0(a_{i, k}|x_{i, k})} r_{i, k} s_{\zeta}(x_{i, k}, a_{i, k}), 
\end{align*}
To estimate the policy gradient for the time step index $K+\delta$, Prognosticator takes the weighted sum of the gradient of the policy value in each period $k$ where the weight $\frac{\partial \hat{V}_{K+\delta}(\pi_{\zeta}; \calD)}{\partial \hat{V}^{\text{IPS}}_k(\pi_{\zeta}; \calD_k)}$ represents how much change in the policy value in period $k$ affects the change in the policy value in the future time step index $K + \delta$. Similar to the discussion in Section~\ref{sec:prognosticator}, there are two limitations of Prognosticator in both (i) estimation phase and (ii) extrapolation (regression) phase. First, the variance of the estimator $\nabla_{\zeta} \hat{V}^{\text{IPS}}_{k}(\pi_{\zeta}; \calD_k)$ of the gradient of the policy value considerably increases in (i) estimation phase when the distributional shifts frequently occur, which is often the case in the real-world applications. Moreover, it incurs non-negligible bias as capturing the trends by time step index $k$ in (ii) extrapolation (regression) phase with simple linear regression is challenging.

\subsection{Additional results for F-OPE under non-stationary reward distribution}
\label{app:additional_results_prog_phi}
Figure \ref{fig:f-ope-target-time-lambda-prog-phi} (i) compares the accuracy of the Prognosticator $\phi$ estimator compared to the other estimators in F-OPE with varying target time ($t'$) spanning from target time 1-45 days (season 1) to 320-365 days (season 8) after we collect the logged data $T$. We observe that using the time feature function $\phi$ rather than the mere basis function $\psi$ improves the estimation accuracy, particularly during seasons 1 and 2, where the original Prognosticator estimator evaluates the true values as inaccurately as in season 1 or even worse in season 2 than the conventional OPE estimators.

Figure \ref{fig:f-ope-target-time-lambda-prog-phi}
 (ii) illustrates the MSE of the Prognosticator $\phi$ and the other estimators in the main text with varying levels of the informativeness of the time feature ($\lambda$) from $0.0$ to $1.0$ by $0.2$. As the time feature $\phi$ becomes more informative in the expected reward $q(x, t, a)$, the Prognosticator $\phi$ estimator reduces the MSE of the Prognosticator estimator, demonstrating that the use of the time feature function $\phi$ contributes to the decrease in the MSE by approximately half the difference in the MSE between OPFV and the Prognosticator estimator.

 Figure \ref{fig:f-ope-n-trains-prog-phi} reports the precision of the estimators when we vary the logged data sizes ($n$) from $500$ to $4000$. Overall, OPFV performs more accurately than the baseline methods, irrespective of the data size, because it can at least unbiasedly estimate the time feature effect, leading to the lowest bias, as we can see in the middle figure. Note that OPFV has a higher variance than IPS and DR due to the additional weighting factor to estimate the time feature effect, but its bias reduction advantage is much larger, making OPFV the most effective estimator. IPS and DR produce the largest bias due to their unrealistic stationarity assumption. Prognosticator fails particularly when the data size is limited due to the significant variance produced when performing OPE in each period with a small sub-data size $n_k$. Moreover, the figure clearly illustrates that using the time feature $\phi$ reduces the bias of OPFV since the bias of the Prognosticator $\phi$ is as low as OPFV. In contrast, the Prognosticator $\phi$ estimator still suffers from as high variance as the original Prognosticator estimator, particularly when the logged data size is limited since they conduct the OPE using the logged data with size $n_k$ in each period $k \in [K]$.

\subsection{Additional results for F-OPL under non-stationary reward distribution}
\label{app:additional_results}

We present the additional results on F-OPL under non-stationary reward with (i) varying levels of the informativeness of the time feature ($\lambda$) and (ii) varying cardinality of (tuned) time features ($|\hat{\phi}|$) used in the OPFV-PG. We compared three OPL methods (RegBased, IPS-PG, and DR-PG) for stationary environments, Prognosticator as the method for the non-stationary environments, and our proposed OPFV-PG.

\begin{figure}[t]
\centering
\vspace{1mm}
\includegraphics[clip, width=12cm]{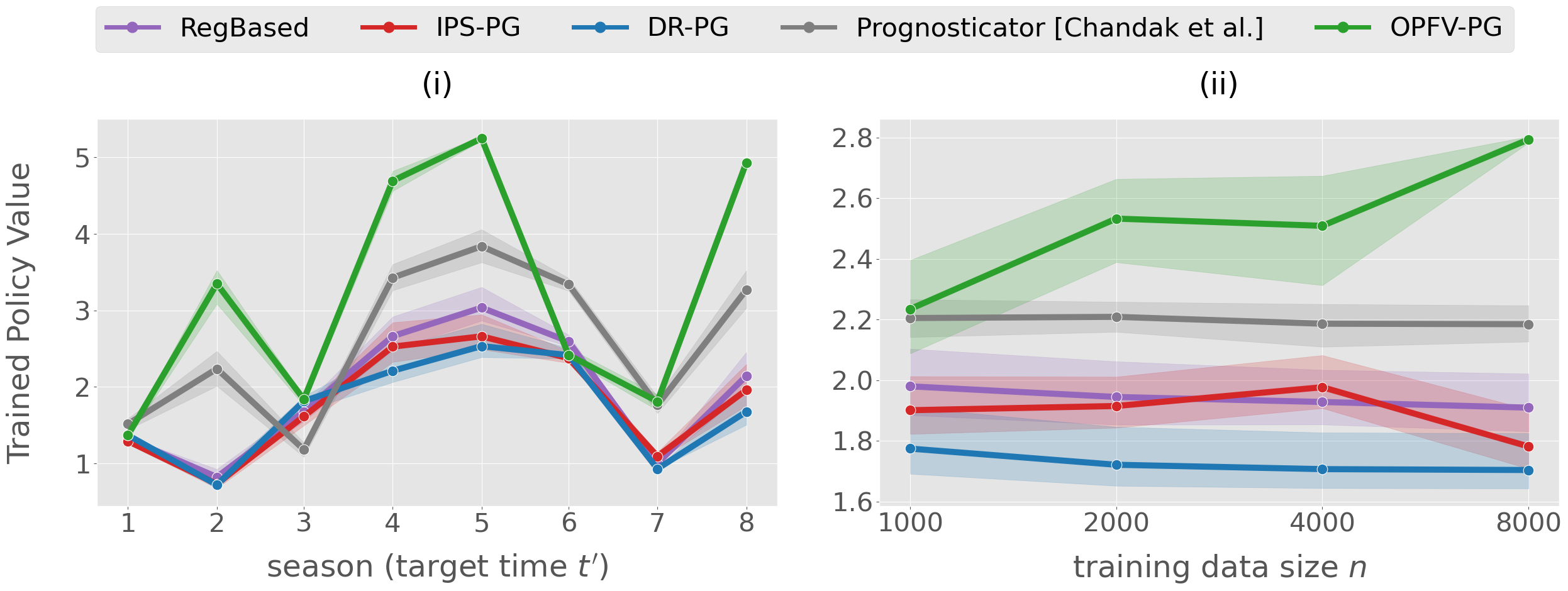}
\vspace{-3mm}
\caption{Comparison of the policy values achieved by F-OPL methods with (left) varying target times $t'$ and (right) the logged data sizes (training data) $n$.} \label{fig:f-opl-targe-time-n-trains}
\vspace{-4mm}
\end{figure}

\begin{figure*}
\centering
\vspace{1mm}
\includegraphics[clip, width=12cm]{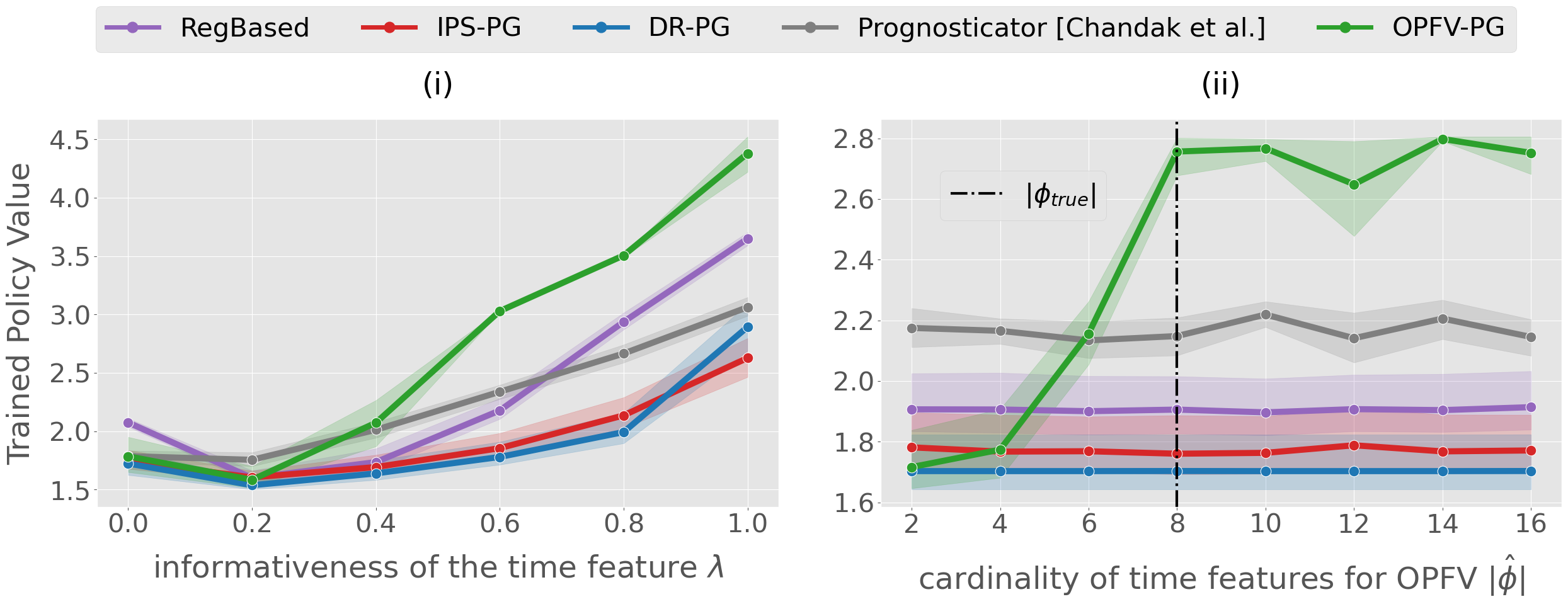}
\vspace{-3mm}
\caption{Comparison of the test policy values of the Regression-based approach, IPS and DR Policy Gradient, Prognosticator, and OPFV-PG with (i) varying levels of informativeness of the time feature $\lambda$, (ii) varying cardinalities of time features for OPFV-PG, and (iii) varying target time $t'$} \label{fig:f-opl-lambda-num-time-features}
\raggedright
\end{figure*}

Figure~\ref{fig:f-opl-targe-time-n-trains} (i) presents the policy values achieved by five OPL methods with varying target times $t'$. We observe that OPFV-PG mostly achieves the highest value adaptive to different target times $t'$ due to its effective usage of the time features in estimating the policy gradient for future values. Prognosticator rarely produces a better policy than OPFV-PG and it ends up with a policy even worse than the conventional OPL methods in season 3. In contrast, OPFV-PG always performs better than the conventional OPL methods. 

Figure~\ref{fig:f-opl-targe-time-n-trains} (ii) reports the earned future policy values with varying logged data sizes. As the logged data size $n$ increases, OPFV-PG is able to learn increasingly better policies because of its low-bias estimation of the policy gradient $\nabla_{\zeta} V_{t'}(\pi_{\zeta})$ for any target time $t'$ in the future. In contrast, the baseline methods are based on biased policy gradients, so they do not necessarily improve even with increased training data.

Figure \ref{fig:f-opl-lambda-num-time-features} (i) illustrates the learned policy values when we vary the levels of how informative the time feature is ($\lambda)$ ranging from $0.0$ to $1.0$ by $0.2$. The results show that OPFV-PG demonstrates a relatively higher learned policy value as the time feature becomes more informative. In instances where the time feature effect is subtle, the policy learned by OPFV-PG is comparable to conventional OPL methods and the Prognosticator.

Figure \ref{fig:f-opl-lambda-num-time-features} (ii) demonstrates the impact of varying the cardinalities of (tuned) time features ($|\hat{\phi}|$) on the performance of OPFV-PG, with the true cardinality of time features set at eight ($|\phi_{\text{true}}|=8$). The results reveal that OPFV-PG enables us to learn a policy with the highest policy value among the baseline OPL methods when the cardinality of the used time feature is from 6 to 16. Even with coarse time features, the policy value learned by OPFV-PG is comparable to DR. As the cardinality of time features increases, the value of OPFV-PG grows until it aligns with the true cardinality of time features, which is eight, which is consistent with our bias analysis in Section~\ref{sec:analysis_opfv_f_opl}. Furthermore, we observe that using finer-grained time features than the true one leads to a decrease in the value of OPFV-PG. This reduction is attributed to an increase in the variance of the estimated policy gradient $\opfvgrad$, while the bias remains unaffected. Consequently, the value of OPFV-PG with overly fine-grained time features slightly decreases compared to the case where the true cardinality of time features is employed. Nevertheless, it maintains the highest value among the five OPL methods, including Prognosticator.

\subsection{Additional experimental setup for F-OPE under non-stationary context and reward distributions}
\label{app:additional_setup_ns_x} To empirically identify the situations when the extended OPFV estimator proposed in Section~\ref{appendix.non_stationary_context} enables us to evaluate a policy in the future under non-stationary context and reward distributions, we conducted a synthetic data experiment. In this section, we describe how we synthesize the non-stationary context. We generate context $x$ from the following non-stationary distribution $p(x|t)$:
\begin{align}
    p(x|t) = \alpha \cdot p_1(x|\phi_x(t)) + (1 - \alpha) \cdot p_2 (x|t) \label{eq:ns_x}, 
\end{align}
where $p_1(x|\phi_x(t)) = \prod_{d \in [d_x]} \calN (\mu_1, \sigma_1^2)$ and $p_2(x|t) = \prod_{d \in [d_x]} \calN (\mu_2, \sigma_2^2)$. For the mean value $\mu_1$ of the component of the non-stationary context depending on the time feature $\phi_x$ for the context, we use $\mu_1 = \gamma^\top \onehottimefeaturecontext$ where $\onehottimefeaturecontext$ is the one hot embedding of the time feature for the context $\phi_x$, and $\gamma$ is a parameter sampled from uniform distribution ranging from -3 to 3. In contrast, for the mean value $\mu_2$, we set it to $\mu_2 = \kappa s(t)$ where $s(t) = \frac{t - \oldesttime}{\futuretime - \oldesttime}$ is the normalized time where we use the UNIX timestamps of January 1st in Year 1 for $\oldesttime$ and January 1st in Year 3 for $\futuretime$, and $\kappa$ is the parameter sampled from the uniform distribution ranging from -1 to 1. We set the variance of the normal distribution to 1 for $\sigma_1^2$ and $\sigma_2^2$.

\begin{figure}
\centering
\vspace{1mm}
\includegraphics[clip, width=8cm]{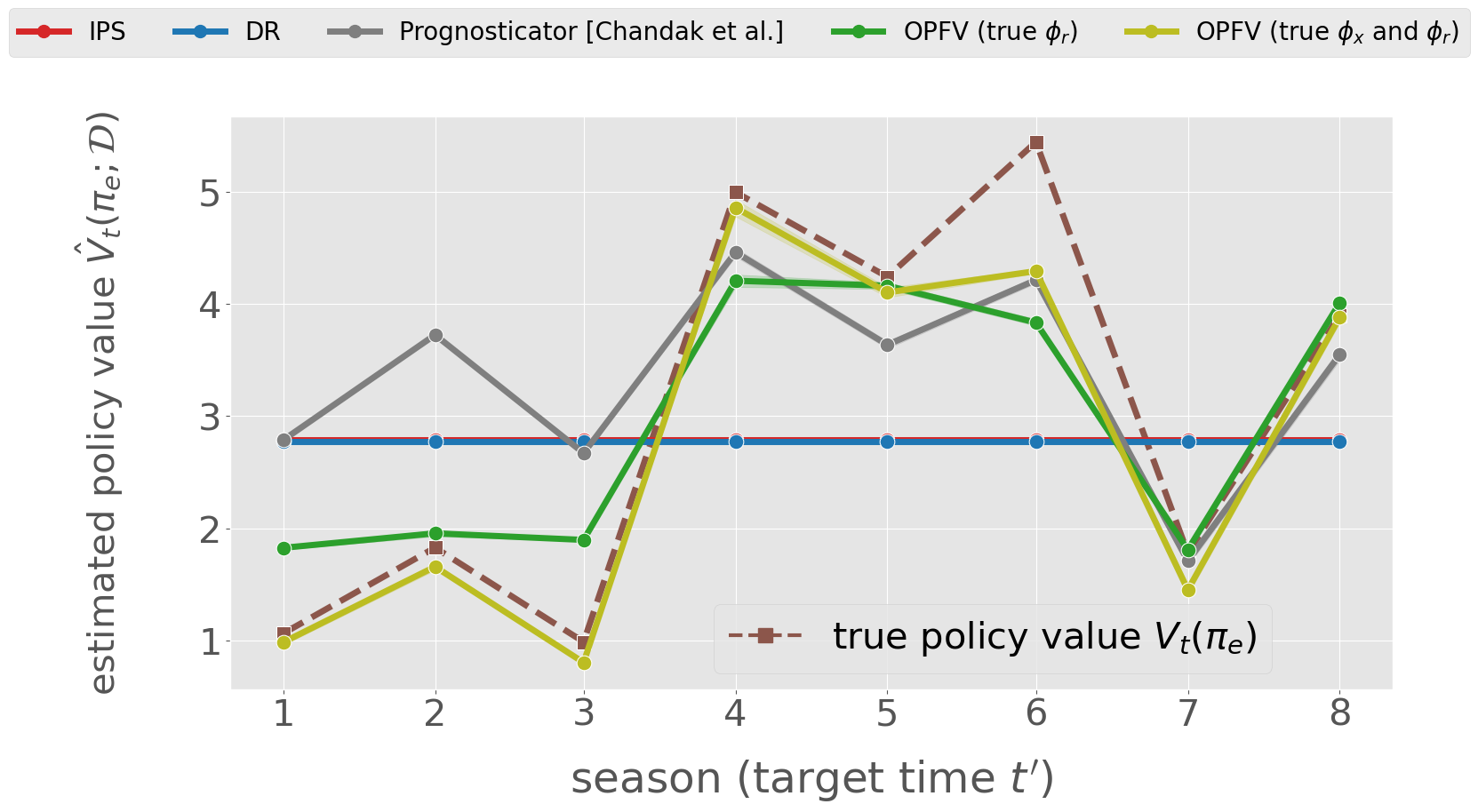}
\vspace{-3mm}
\caption{Comparison of the true value and estimated values of IPS, DR, Prognosticator, OPFV (true $\phi_r$) for non-stationary reward and OPFV (true $\phi_x$ and $\phi_r$) for non-stationary context and reward distributions with varying target time $t'$} \label{fig:f-ope_ns_x_target_time}
\raggedright
\end{figure}

\subsection{Additional results for F-OPE under non-stationary context and reward distributions}
\label{app:additional_result_ns_x}
We compare the precision of OPFV adaptive to the non-stationary reward in the main text denoted by OPFV (true $\phi_r$) and the extended OPFV for both non-stationary context and reward distributions denoted by OPFV (true $\phi_x$ and $\phi_r)$) with conventional OPE estimators (IPS and DR), and the Prognosticator estimator. We use the same experimental setting as the non-stationary reward-only environments except for generating the non-stationary context described in Eq.~\eqref{eq:ns_x}. We set the true cardinality of time feature $|\phi_x|$ for context to eight, the same as that for reward.

\begin{figure*}[t]
\centering
    \begin{subfigure}
        \centering
        \vspace{1mm}
        \includegraphics[clip, width=13.5cm]{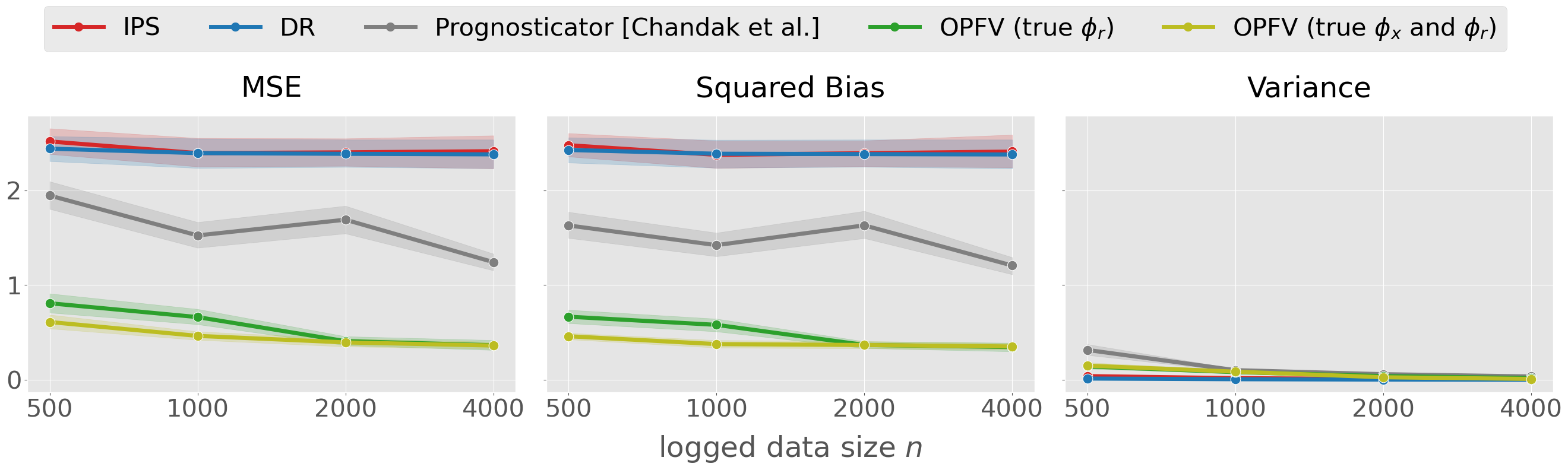}
        \vspace{-3mm}
        \caption{Comparison of the MSE, bias, and variance of IPS, DR, Prognosticator, OPFV (true $\phi_r$) for non-stationary reward distributions and OPFV (true $\phi_x$ and $\phi_r$) for non-stationary context and reward distributions with varying the logged data sizes $n$} \label{fig:f-ope_ns_x_n_trains}
        \raggedright
    \end{subfigure}
    \vspace{5mm}
    \begin{subfigure}
        \centering
        \vspace{1mm}
        \includegraphics[clip, width=13.5cm]{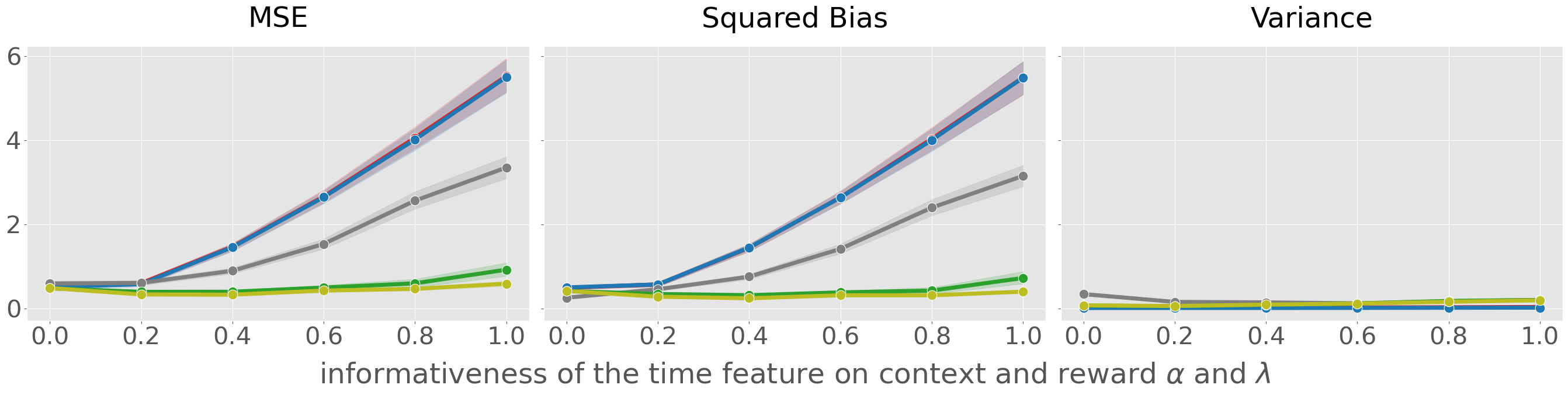}
        \vspace{-3mm}
        \caption{Comparison of the MSE, bias, and variance of IPS, DR, Prognosticator, OPFV (true $\phi_r$) for non-stationary reward distributions and OPFV (true $\phi_x$ and $\phi_r$) for non-stationary context and reward distributions with varying levels of the informativeness of the time features for context and reward $\alpha$ and $\lambda$} \label{fig:f-ope_ns_x_alpha_lambda}
        \raggedright
    \end{subfigure}
\end{figure*}

Figure \ref{fig:f-ope_ns_x_target_time} presents the true value of an evaluation policy and the estimated values by OPE and F-OPE estimators under the distributional shifts of context and reward with varying target time range ($t'$). We observe that OPFV (true $\phi_x$ and $\phi_r$) evaluates an evaluation policy at an arbitrary target time $t'$ more correctly than OPFV (true $\phi_r$) by adapting to not only the non-stationary reward but also non-stationary context. We also observe that the original OPFV estimator enables us to evaluate the policy value better than the Prognosticator and conventional OPE estimators, even under non-stationary context and reward distributions if the time feature for context $\phi_x$ and for reward $\phi_r$ are identical. Prognosticator tries to adapt to the non-stationary environment. Still, it uses the time step index $k$, failing to represent the effect of the time feature correctly, and hence, incurring as significant bias as IPS and DR for target time ranges 1-45 (season 1) and 92-136 (season 3) and more bias than conventional OPE estimators for the target time 46-91 (season 2). IPS and DR evaluate the policy uniformly regardless of the target time because they do not consider the target time $t'$.

Figure \ref{fig:f-ope_ns_x_n_trains} illustrates the accuracy of the estimators under non-stationary context and reward distributions when we vary the logged data sizes ($n$) from $500$ to $4000$. Overall, the extended OPFV for non-stationary context and reward distributions can most precisely estimate the value of an evaluation policy. We also observe that the OPFV adaptive to the non-stationary reward is more precise than the Prognosticator even under the non-stationarity in both the context and reward, provided that the time features for context and reward are identical. This is because the first term of OPFV for non-stationary reward in Eq.~\eqref{eq:opfv} becomes the same as the extended OPFV estimator for nonstationary context and reward in Eq.~\eqref{eq:opfv_ns_x}.

Figure \ref{fig:f-ope_ns_x_alpha_lambda} compares the accuracy of the estimators when we vary the levels of informativeness ($\alpha$) and ($\lambda$) of time features for both contexts and reward. The results illustrate that as the time feature for both context and reward becomes informative, IPS, DR, and the Prognosticator estimator incur significant bias, resulting in an inaccurate future value evaluation, whereas the extended OPFV estimator mitigates this bias by using the time feature. As OPFV (true $\phi_r$) is adaptive only to the non-stationarity in reward, it gradually incurs a slight bias due to the non-stationary context. OPFV (true $\phi_x$ and $\phi_r$), however, overcomes this limitation by using the time feature not only for reward but also for context, constantly yielding the lowest MSE with varying levels of the informativeness of the time features for context and reward. When the time feature is less informative ($\alpha \le 0.2$ and $\lambda \le 0.2$), OPFV (true $\phi_r$) and OPFV (true $\phi_x$ and $\phi_r$) evaluate an evaluation policy as precisely as the conventional OPE estimators and the Prognosticator.

\begin{figure*}[t]
    \centering
    \vspace{1mm}
    \includegraphics[clip, width=13.5cm]{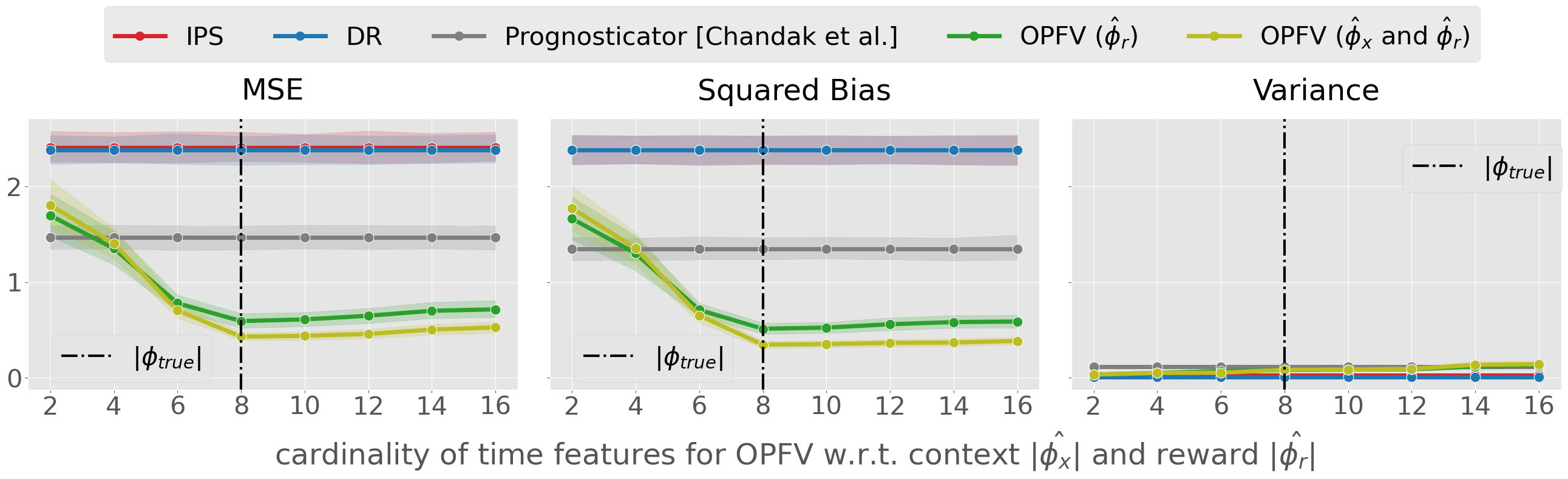}
    \vspace{-3mm}
    \caption{Comparison of the MSE, bias, and variance of IPS, DR, Prognosticator, OPFV ($\hat{\phi}_r$) for non-stationary reward distributions and OPFV ($\hat{\phi}_x$ and $\hat{\phi}_r$) for non-stationary context and reward distributions with varying cardinality of time feature $\hat{\phi}_x$ and $\hat{\phi}_r$ for OPFV} \label{fig:f-ope_ns_x_num_time_feature}
    \raggedright
\end{figure*}

Figure \ref{fig:f-ope_ns_x_num_time_feature} depicts how the number of (tuned) time features for context and reward ($|\hat{\phi_x}|$ and $|\hat{\phi_r}|$) used in OPFV affects the precision of the estimators in F-OPE, where the true numbers of time features for both context and reward are eight. As the cardinality of time features increases, we observe that the precision of the OPFV improves due to the reduction in the bias with a slight increase in the variance. Moreover, the result demonstrates that OPFV with time features for context and reward is more precise than OPFV with a time feature for reward only, after the cardinality of time features for context and reward becomes large enough to capture the true time feature.

\subsection{Investigation on the computational efficiency}\label{app:additional-computational-efficiency}
This section empirically investigates the computational efficiency of our proposed method in comparison to the doubly robust estimator in F-OPE. Our aim is to demonstrate that the computational cost associated with optimizing the time feature functions used in OPFV is manageable, as described in Section~\ref{sec:tune}. Specifically, Table~\ref{tab:synthetic-computatinal-efficiency} demonstrates the average computation times of three F-OPE algorithms with varying logged data sizes from $500$ to $4000$. The experimental setting is consistent with that of Figure~\ref{fig:f-ope-n-trains} in Section~\ref{sec:synthetic-data}. The table shows that the OPFV estimator completes executions in a few seconds even with the largest data size $n=4000$ although OPFV with both true $\phi$ and tuned $\hat{\phi}$ are slower than DR due to the consideration of the additional time feature. Furthermore, in practical applications, it is not always required to complete the executions within minutes. Thus, we can control the computational costs of OPFV by adjusting the cardinality $|\Phi|$ of the set of the candidate time features.

\begin{table}[t]
    \caption{Comparison of the average computation times (sec) for DR, OPFV (true $\phi$), and OPFV (tuned $\hat{\phi}$) under varying logged data sizes ($n$)}
    \vspace{-1mm}
    \centering
    \scalebox{0.85}{
        \begin{tabular}{l|cccc}
        \toprule
        logged data size ($n$) & 500 & 1000 & 2000 & 4000 \\
        \midrule
        \midrule
        DR & 0.03599 & 0.06461 & 0.12937 & 0.27898 \\
        OPFV (true $\phi$) & 0.05026 & 0.08649 & 0.17060 & 0.34300 \\
        OPFV (tuned $\hat{\phi}$) & 0.41265 & 0.70762 & 1.39140 & 2.82739 \\
        \bottomrule
        \end{tabular}
        }
    \label{tab:synthetic-computatinal-efficiency}
\vspace{-3mm}
\end{table}

\section{Additional setup for real-world data experiment}\label{app:additional-setup-real-all}

\subsection{Non-stationarity in KuaiRec dataset}\label{app:non-stationarity-in-kuairec}
\begin{figure*}[t]
\centering
    \begin{subfigure}
    \centering
    \vspace{1mm}
        \includegraphics[clip, width=14cm]{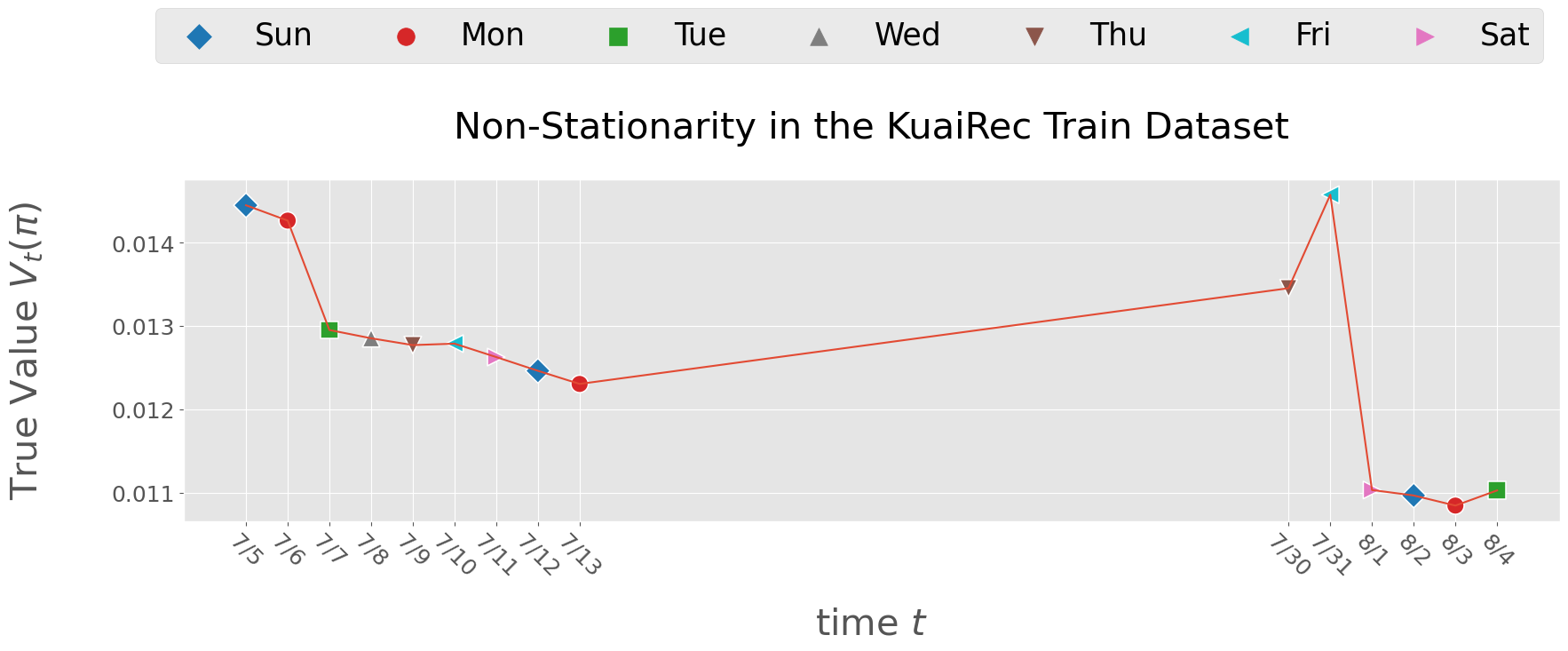}
        \caption{Comparing the true value $V_t(\pi)$ of completely random policy $\pi$ with varying dates in Kuai Rec training dataset}
        \label{fig:f-opl-real-train-non-stationarity}
        \raggedright
    \end{subfigure}
    \vspace{5mm}
    \begin{subfigure}
    \centering
    \vspace{1mm}
        \includegraphics[clip, width=14cm]{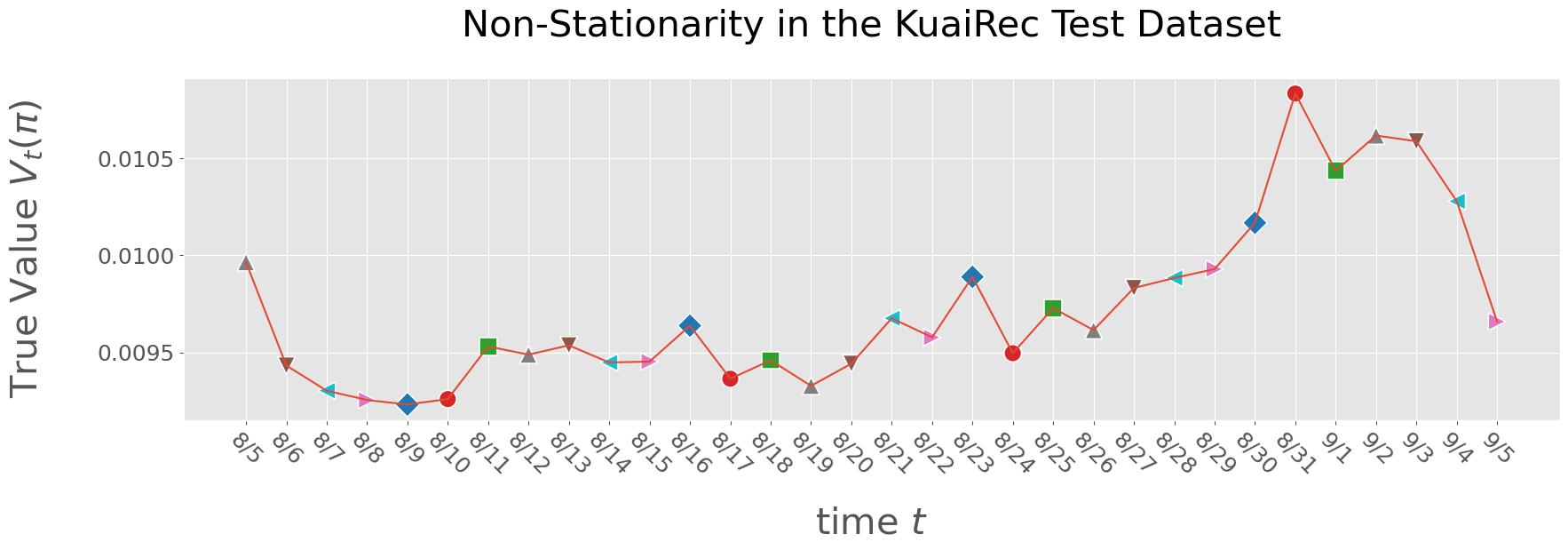}
        \caption{Comparing the true value $V_t(\pi)$ of completely random policy $\pi$ with varying dates in Kuai Rec test dataset}
        \label{fig:f-opl-real-test-non-stationarity}
        \raggedright
    \end{subfigure}
\end{figure*}

This section empirically validates that the KuaiRec dataset we used as the real-world data in Section~\ref{sec:real-experimnet} contains non-stationarity. Figure \ref{fig:f-opl-real-test-non-stationarity} illustrates the value $V_t(\pi)$ of a policy $\pi$ where $\pi(a|x, t) = \frac{1}{|\calA|}$ assigns equal probabilities to all actions throughout the time $t$ in KuaiRec test dataset. Non-stationarity is evident in the dataset, alongside weekly trends. For instance, we observe that the policy's values are relatively high on Sundays, such as on August 16th and 23rd. In contrast, the values on Mondays are relatively low compared to the other weekdays, such as August 17th and 24th. Furthermore, we also observe that the training data also have similar weekly trends, such as high values on Sundays, July 5th, 12th, and August 2nd, and relatively low values on Mondays, July 6th, 13th, and August 3rd. These observations suggest using time features such as the day of the week will enable the capture of periodic fluctuations.

\subsection{Optimization of the time feature function for OPFV}\label{app:real-time-feature-opt}
The previous section addressed the non-stationarity in the KuaiRec dataset, and we observe that there is a day-of-the-week trend in the dataset. To cope with such non-stationarity, this section provides how the OPFV-PG selected the time feature functions in real-world data experiments. For OPFV-PG (w/o tuned $\phi$), we employ the day of the week as the time feature $(i.e., \phi_{\text{day\_of\_week}}: [0, t'] \rightarrow \{ \text{Sun}, \text{Mon}, \text{Tue}, \text{Wed}, \text{Thu},  \text{Fri}, \text{Sat}  \})$ as we briefly described in Section~\ref{sec:real-experimnet}. For OPFV-PG (w/ tuned $\phi$), we employ the following set 
\begin{align*}
    \Phi &:= \Big\{ \phi_{\text{day\_of\_week}}, \phi_{\text{weekday\_weekend}}, \phi_{\text{hour}}, \phi_{\text{four\_per\_day}}, \phi_{\text{am\_pm}}, \\
    & \phi_{\text{day\_of\_week}} \bigotimes \phi_{\text{hour}}, \phi_{\text{day\_of\_week}} \bigotimes \phi_{\text{four\_per\_day}}, \phi_{\text{day\_of\_week}} \bigotimes \phi_{\text{am\_pm}}, \\
    & \phi_{\text{weekday\_weekend}} \bigotimes \phi_{\text{hour}}, \phi_{\text{weekday\_weekend}} \bigotimes \phi_{\text{four\_per\_day}}, \phi_{\text{weekday\_weekend}} \bigotimes \phi_{\text{am\_pm}}  \Big\}, 
\end{align*}
where the first five functions are defined as follows.
\begin{align*}
    \phi_{\text{day\_of\_week}} &: [0, t'] \rightarrow \{ \text{Sun}, \text{Mon}, \text{Tue}, \text{Wed}, \text{Thu},  \text{Fri}, \text{Sat}  \}\\
    \phi_{\text{weekday\_weekend}} &: [0, t'] \rightarrow \{ \text{weekday}, \text{weekend}  \}\\
    \phi_{\text{hour}} &: [0, t'] \rightarrow \{ \text{12:00am--0:59am}, \text{1:00am--1:59am}, \cdots , \text{11:00pm--11:59pm}  \}\\
    \phi_{\text{four\_per\_day}} &: [0, t'] \rightarrow \{ \text{12:00am--5:59am}, \text{6:00am--11:59am}, \text{12:00pm--5:59pm}, \text{6:00pm--11:59pm}  \}\\
    \phi_{\text{am\_pm}} &: [0, t'] \rightarrow \{ \text{12:00am--11:59am}, \text{12:00pm--11:59pm}  \}
\end{align*}
Moreover, for any $\phi_i, \phi_j \in \{ \phi_{\text{day\_of\_week}}, \phi_{\text{weekday\_weekend}}, \phi_{\text{hour}}, \phi_{\text{four\_per\_day}}, \phi_{\text{am\_pm}} \}$, we define the time feature function which combine both $\phi_i$ and $\phi_j$ by $\phi_i \bigotimes \phi_j: [0, t'] \rightarrow C_i \times C_j$ if the codomains of $\phi_i$ and $\phi_j$ are denoted by $C_i$ and $C_j$, respectively, (i.e., $\phi_i: [0, t'] \rightarrow C_i$ and $\phi_j: [0, t'] \rightarrow C_j$). Then, to execute the optimization, we use $\phi_{\infty} := \phi_{\text{day\_of\_week}} \bigotimes \phi_{\text{hour}}$ as the finest time feature since it is strictly finer than any time feature $\phi \in \Phi$.

\subsection{Evaluation of the methods in F-OPL}
In this section, we describe the evaluation of policies learned by the methods in F-OPL. For each learned policy $\pi_{\zeta^*}$, we are interested in its value $V(\pi_{\zeta^*}) := \mE_{p(t')}[V_{t'}(\pi_{\zeta^*})]$ in the future where $V_{t'}(\pi_{\zeta^*}) := \mE_{p(x|t') \pi_{\zeta^*}(a|x, t')}[q(x, t', a)]$ is the value at arbitrary target time $t'$ and $q(x, t', a) := \mE_{p(r|x, t', a)}[r]$ is the expected reward function at given target time $t'$. We estimate the value $V(\pi_{\zeta^*})$ by using the logged data $\calD_{\text{te}} := \{ (x_i, t_i, a_i, r_i) \}_{i \in [n_{\text{te}}]} \sim \prod_{i \in [n_{\text{te}}]} p(t_i) p(x_i|t_i) \pi_0(a_i|x_i, t_i) p(r_i|x_i, t_i, a_i)$. Specifically, we use Direct Method (DM), Self-Normalized Inverse Propensity Scoring (SNIPS) ~\citep{swaminathan2015self}, and Self-Normalized Doubly Robust (SNDR) ~\citep{robins2007comment, thomas2016data}.

\textbf{Direct Method (DM): } DM uses the estimator of the expected reward and is defined as follows.
\begin{align*}
    \hat{V}_{\text{DM}}(\pi_{\zeta^*}; \calD_{\text{te}}, \hat{q}) := \frac{1}{n_{\text{te}}} \sum_{i \in [n_{\text{te}}]} \mE_{\pi_{\zeta^*}(a|x_i, t_i)}[\hat{q}(x_i, t_i, a)] = \frac{1}{n_{\text{te}}} \sum_{i \in [n_{\text{te}}]} \sum_{a \in \calA} \pi_{\zeta^*}(a|x_i, t_i) \hat{q}(x_i, t_i, a), 
\end{align*}
where $\hat{q}(x, t, a)$ is the estimator of the expected reward $q(x, t, a)$. DM often has a lower variance than the estimators, which use importance weights. However, it often incurs bias due to the misspecification of the regression model.

\textbf{Self-Normalized Inverse Propensity Scoring (SNIPS) ~\citep{swaminathan2015self}: } In contrast to DM, SNIPS is one of the model-free approaches defined as follows.
\begin{align*}
    \hat{V}_{\text{SNIPS}}(\pi_{\zeta^*}; \calD_{\text{te}}) := \frac{1}{n_{\text{te}}} \sum_{i \in [n_{\text{te}}]} \frac{w(x_i, t_i, a_i) r_i}{\frac{1}{n_{\text{te}} } \sum_{j \in [n_{\text{te}}]} w(x_j, t_j, a_j)} =   \frac{ \sum_{i \in [n_{\text{te}}]} w(x_i, t_i, a_i) r_i}{ \sum_{j \in [n_{\text{te}}]} w(x_j, t_j, a_j)}, 
\end{align*}
where $w(x, t, a) := \pi_{\zeta^*}(a|x, t) / \pi_0(a|x, t)$ is the importance weight and $\frac{1}{n_{\text{te}}} \sum_{j \in [n_{\text{te}}]} w(x_j, t_j, a_j)$ is the empirical mean of the importance weights as the normalizing factor. SNIPS mitigates the high variance incurred in the IPS estimator due to the use of the normalizing factor. Moreover, SNIPS is a consistent estimator and is known to work empirically well~\citep{saito2021counterfactual}.

\textbf{Self-Normalized Doubly Robust (SNDR) ~\citep{robins2007comment, thomas2016data}: } SNDR combines DM and SNIPS as follows.
\begin{align*}
    \hat{V}_{\text{SNDR}}(\pi_{\zeta^*}; \calD_{\text{te}}, \hat{q}) := \frac{1}{n_{\text{te}}} \sum_{i \in [n_{\text{te}}]}  \left\{ \mE_{\pi_{\zeta^*}(a|x_i, t_i)}[\hat{q}(x_i, t_i, a)] + \frac{w(x_i, t_i, a_i) (r_i - \hat{q}(x_i, t_i, a_i))}{\frac{1}{n_{\text{te}} } \sum_{j \in [n_{\text{te}}]} w(x_j, t_j, a_j)} \right\}, 
\end{align*}
where the first term is the same as DM, and the second term uses $\hat{q}(x, t, a)$ as the control variate to have a lower variance than SNIPS.

\subsection{Additional results by DM, SNIPS, and SNDR evaluations}\label{app:additional-results-by-DM-SNIPS-SNDR}
Table~\ref{tab:real_data_DM_SNIPS_SNDR} compares the mean and standard deviation of the values of the learned policies evaluated by three different OPE estimators across ten experimental runs with different seeds before we take the average of the results of three evaluations shown in Section~\ref{sec:real-experimnet}. No matter what OPE estimator we use to evaluate the policies, two versions of OPFV-PG invariably enable us to achieve the highest policy values under unknown and potentially intricate real-world non-stationary environments. More specifically, we observe that the mean values of OPFV-PG (w/ tuned $\phi$) are consistently higher than OPFV-PG (w/o tuned $\phi$) under all three evaluation methods. These results empirically validate the real-world applicability of the optimization procedure of time feature functions in Section~\ref{sec:tune} for real-world data. Additionally, the table demonstrates that OPFV-PG (w/o tuned $\phi$) achieves higher mean values than any baseline OPL methods for any evaluation method. Thus, even without the optimization of the time feature functions, our proposed method for F-OPL enables effective learning under non-stationarity.

\begin{table}[t]
    \caption{Comparison of estimated values and ranks of policies learned by F-OPL methods in the test set by three different evaluation methods.}
    \vspace{-1mm}
    \centering
    \scalebox{0.63}{
    \begin{tabular}{ll|cccccc}
    \toprule
     Method& Metric & RegBased & IPS-PG & DR-PG & Prognosticator & OPFV-PG (w/o tuned $\phi$) & OPFV-PG (w/ tuned $\phi$) \\
    \midrule
    \midrule
    DM & Mean Value ($\pm$ SD) & 1.52($\pm$ 0.32) & 1.12($\pm$ 0.22) & 1.33($\pm$ 0.23) & 1.36($\pm$ 0.49) & 1.63($\pm$ 0.56) & \textbf{1.68($\pm$ 0.61)} \\
    SNIPS & Mean Value ($\pm$ SD) & 1.43($\pm$ 0.29) & 1.08($\pm$ 0.74) & 1.16($\pm$ 0.56) & 1.38($\pm$ 0.61) & 1.56($\pm$ 0.55) & \textbf{1.64($\pm$ 0.63)} \\
    SNDR & Mean Value ($\pm$ SD) & 1.48($\pm$ 0.33) & 0.85($\pm$ 1.19) & 0.95($\pm$ 0.99) & 1.42($\pm$ 0.71) & 1.6($\pm$ 0.58) & \textbf{1.67($\pm$ 0.65)} \\
    \cline{1-8}
    \bottomrule
    \end{tabular}
        }
    \label{tab:real_data_DM_SNIPS_SNDR}
\vspace{-4mm}
\end{table}

\end{document}